# Reasoning on Interval and Point-based
# Disjunctive Metric Constraints in Temporal Contexts

**Federico Barber**                                    FBARBER@DSIC.UPV.ES
*Dpto. de Sistemas Informáticos y Computación*
*Universidad Politécnica de Valencia*
*Camino de Vera s/n, 46022 Valencia, Spain*

## Abstract

We introduce a temporal model for reasoning on disjunctive metric constraints on intervals and time points in temporal contexts. This temporal model is composed of a labeled temporal algebra and its reasoning algorithms. The labeled temporal algebra defines *labeled* disjunctive metric point-based constraints, where each disjunct in each input disjunctive constraint is univocally associated to a label. Reasoning algorithms manage labeled constraints, associated label lists, and sets of mutually inconsistent disjuncts. These algorithms guarantee consistency and obtain a minimal network. Additionally, constraints can be organized in a hierarchy of alternative temporal contexts. Therefore, we can reason on context-dependent disjunctive metric constraints on intervals and points. Moreover, the model is able to represent non-binary constraints, such that logical dependencies on disjuncts in constraints can be handled. The computational cost of reasoning algorithms is exponential in accordance with the underlying problem complexity, although some improvements are proposed.

## 1. Introduction

Two main lines of research are commonly recognized in the temporal reasoning area. The first approach deals with reasoning about temporal constraints on time-dependent entities. The goal is to determine what consequences ($T$) follow from a set of temporal constraints, *"{Temporal-Constraints}|=T?"*, or to determine whether a set of temporal constraints is consistent, with no assumptions about properties of temporal facts. The second approach deals with reasoning about change, events, actions and causality. Here, the goal is to obtain the consequent state from a set of actions or events which are performed on an initial state: *"[Si, {A₁, A₂, ..., Aₙ}]|= Sⱼ?"*. Both these approaches constitute active fields of research with applications in several artificial intelligence areas such as reasoning about change, scheduling, temporal planning, knowledge-based systems, natural language understanding, etc. In these areas, time plays a crucial role, problems have a dynamic behavior, and it is necessary to represent and reason about the temporal dimension of information.

In this paper, we deal with the first of these approaches. Our goal is reasoning on qualitative and quantitative constraints between intervals or time-points in temporal contexts. Moreover, special cases of non-binary constraints are also managed. These tasks are pending issues in the temporal reasoning area, as well as important features to facilitate modeling of relevant problems in this area (including planning, scheduling, causal or hypothetical reasoning, etc.).

Several temporal reasoning models have been defined in the literature, with a clear trade-off between representation expressiveness and complexity of reasoning algorithms. *Qualitative Point Algebra* (PA) (Vilain, Kautz & Van Beek, 1986) is a limited subset of interval-based models. *Interval*

*Algebra* (IA) introduced by Allen (1983) represents symbolic (*qualitative*) constraints between intervals but metric information, such as *'interval₁ starts 2 seconds before interval₂'*, cannot be included. Metric (*quantitative*) point-based models (Dechter, Meiri & Pearl, 1991) include the 'time line' (metric) in their constraints, but they can only represent a limited subset of disjunctive constraints between intervals. Thus, constraints like *'interval₁ {bef, aft} interval₂'* cannot be represented (Gerevini & Schubert, 1995).

Some efforts have been made to integrate qualitative and quantitative temporal information on points and intervals (Kautz & Ladkin, 1991; Drakengren & Jonsson, 1997; etc.). Particularly, Meiri (1996) introduces *Qualitative Algebra* (QA), where each interval is represented by three nodes (one representing the interval and the other two representing its extreme points) such that QA can represent qualitative and metric constraints on points and intervals. Badaloni and Berati (1996) define the Interval Distance Sub Algebra (IDSA), where nodes are intervals. These intervals are related by disjunctive 4-tuple-metric constraints between their ending time points $\{(\Gamma_i, \Gamma_j), (\Gamma^+_i, \Gamma_j), (\Gamma_i, \Gamma^+_j), (\Gamma^+_i, \Gamma^+_j)\}$. Staab and Hahn (1998) propose a model for reasoning on qualitative and metric boundaries of intervals. However, these models cannot handle constraints on *interval durations*, which were identified earlier by Allen (1983). Constraints such as *'interval₁ lasts 2 seconds more than interval₂'* require a high-order expression (Dechter et al., 1991), or a *duration* primitive which should be integrated with interval and point constraints (Allen, 1983; Barber, 1993). Particularly, Barber (1993) proposes two orthogonal networks to relate constraints on durations and time points. Navarrete (1997) and Wetprasit and Sattar (1998) relate disjunctive constraints on durations and time points, but only a limited subset of interval constraints is managed. More recently, Pujari and Sattar (1999) propose a framework for reasoning on points, intervals and durations (PIDN). Here, variables represent points or intervals, and constraints are an ordered set of three intervals representing (Start, End, Duration) subdomains. However, no specialized algorithms for management of PIDN constraints are proposed.

In relation to the complexity of reasoning algorithms, the consistency problem is polynomial in PA (Vilain, Kautz & Van Beek, 1986) and in non-disjunctive metric networks (Dechter et al., 1991). However, Vilain, Kautz and Van Beek (1986) also showed that determining the consistency of a general-case temporal network (i.e.: disjunctive qualitative and metric constraints between points, intervals or durations) is NP-hard. Thus, in previous qualitative or quantitative models, the consistency problem is tractable only under some properties on constraints, relationships between variable domains and constraints, or by using restricted subsets of constraints (Dechter et al., 1991; Dechter, 1992; van Beek & Detcher, 1995; Wetprasit & Sattar, 1998; Jeavons et al., 1998; etc.). For instance, tractable subclasses of IA have been identified by Vilain, Kautz and Van Beek (1986), Nebel and Burckert (1995), Drakengren and Jonsson (1997), etc. Moreover, some interesting results have been obtained in identification of tractable subclasses of QA. Specifically, Jonsson et al. (1999) identified the five maximal tractable subclasses of the qualitative point-interval algebra. However, to my knowledge the maximal tractable subclass of PIDN model (maximal tractable subclass of qualitative and quantitative point, interval and duration constraints) is still not identified. In any case, these restricted tractable subclasses are not able to obtain expressiveness of full models, and the problem of reasoning on disjunctive constraints on points and intervals remains NP-complete.

On the other hand, these qualitative and metric temporal models do not manage certain types of *non-binary constraints*, which are important for modeling some problems (scheduling, causal reasoning, etc.). For instance, disjunctive assertions like '(*interval₁ {bef, meets} interval₂*) ∨ (*time-*

*point$_3$ is [10 20] from time-point$_4$)'*, or temporal-causal relations like 'If *(interval$_1$ {bef, meets} interval$_2$) then (time-point$_3$ is [10 20] from time-point$_4$)*' should be incorporated in these models (Meiri, 1996). Moreover, the global consistency property introduced by Dechter (1992) is an important property in temporal networks, since it allows us to obtain solutions by backtrack-free search (Dechter, 1992; Freuder, 1982). In particular, a global consistent network would allow us to handle conjunctive queries like *'does '(interval$_1$ {bef, meets} interval$_2$) ∧ (time-point$_3$ is [10 20] from time-point$_4$) hold?'* without propagation of the query, as it is required in (van Beek, 1991).

Stergiou and Koubarakis (1996), Jonsson and Bäckström (1996) dealt with the representation of temporal constraints by means of disjunctions of linear constraints (linear inequalities and inequations) also named Disjunctive Linear Relations (DLRs). These expressions are a unifying approach to manage disjunctive constraints on points, intervals and durations, such that these expressions subsume most of the formalism for temporal constraint reasoning (Jonsson & Bäckström, 1998). Moreover, DLRs are able to represent disjunctions of non-disjunctive metric constraints ($x_1$-$y_1 \leq c_1 \lor x_2$-$y_2 \leq c_2 \lor .... \lor x_n$-$y_n \leq c_n$), where $x_i$ and $y_i$ are time points, $c_i$ real numbers and $n \geq 1$ (Stergiou & Koubarakis, 1998). Obviously, the satisfiability problem for an arbitrary set of disjunctions of linear constraints is NP-complete. Interesting tractable subclasses of DLRs and conditions on tractability are identified in (Cohen et al., 1996; Jonsson & Bäckström, 1996; and Stergiou & Koubarakis, 1996). The two main tractable subclasses are Horn linear and Ord-Horn linear constraints (Stergiou & Koubarakis, 1996; Jonsson & Bäckström, 1998). However, these subclasses subsume temporal algebras whose management is also polynomial.

The management of a set of disjunctions of linear constraints is mainly based on general methods from linear programming, although some specific methods have been defined for tractable subclasses (Stergiou & Koubarakis, 1998; Cohen et al., 1996; etc.). As Pujari and Sattar outline (1999), the linear programming approach, though expressive, does not take advantage of the underlying structures (e.g., domain constraints) of temporal constraints. In addition, usual concepts in temporal reasoning, as composition and intersection operations on constraints, minimal constraints, k-consistency (Freuder, 1982), decomposability (Montanari , 1974), globally consistency (Dechter, 1992), etc., and their consequences should be adapted to reasoning on disjunctive linear constraints, which is not a trivial issue.

In spite of the expressive power of the previous models, some problems (including planning, scheduling, hypothetical reasoning, etc.) also need to reason on *alternative contexts* (situations, intentions or causal projections) and to know what holds in each one of them (Dousson et al., 1993; Gerevini & Schubert, 1995; Garcia & Laborie, 1996; Srivastava & Kambhampati, 1999). This gives rise to the need to reason on context-dependent constraints. This feature is not supported in the usual temporal models in a general way, nor described in the usual expressive power of constraints (Jeavons et al., 1999). Therefore, *ad-hoc* methods should be used when reasoning on temporal contexts is required.

These issues will be addressed in this paper. We describe a temporal model, which integrates qualitative and metric disjunctive constraints on time-points and intervals. The temporal model is based on time-points as primitive, such that intervals are represented by means of their end time-points. However, the representation of interval constraints seems to imply some kind of relation among endpoint constraints (Gerevini & Schubert, 1995). The proposed temporal model introduces *labeled constraints*, where each elemental constraint (disjunct) in a disjunctive point-based metric constraint is associated to *one* unique label. In this way, point-based constraints can be related among them without using hyper-arcs. Therefore, metric and symbolic constraints among intervals and time-points can be fully integrated, represented and managed by means of a *labeled* metric point-based Temporal Constraint Network (TCN). Particularly, the model proposed here handles constraints proposed in QA (Meiri, 1996), IDSA (Badaloni & Berati, 1996), and Distance Constraint Arrays model (Staab & Hahn, 1998). Moreover, several added functionalities are also provided:

- Management of alternative *temporal contexts*. Each input constraint can be associated to a given context. A hierarchy of alternative temporal contexts can be defined, such that constraints between points and intervals are dependent on each context. To my knowledge, these features improve existing temporal models, where contexts are not managed.

- Reasoning algorithms on labeled constraints are based on a closure process. These processes guarantee consistency and obtain a minimal disjunctive context-dependent TCN. Additionally, a special type of *globally labeled-consistent* TCN is obtained. This property allows us to obtain solutions by backtrack-free search (Freuder, 1982).

- Management of a special type of non-binary constraints. Reasoning algorithms are able to manage disjunctions of disjunctive constraints. This supposes an extension of disjunctions of non-disjunctive metric constraints proposed by Stergiou and Koubarakis (1998). Moreover, given a set of disjunctive constraints, the model can handle logical relations among disjunctions of different constraints. Thus, we can express that a set of atomic disjuncts in disjunctive constraints are mutually disjunctive among them. Therefore, a special type of and/or TCN can be managed as a conjunctive (and) TCN. Likewise, the model can also handle special non-binary constraints representing implications among temporal constraints as were identified by Meiri (1996).

With these features, the proposed temporal model is suitable for modeling problems where these requirements appear. The computational cost of reasoning methods is non-polynomial, given the complexity of the underlying problem. However, several improvements are also proposed.

A brief revision of the main temporal reasoning concepts is presented in Section 2. In Section 3, a temporal algebra for labeled point-based disjunctive metric constraints is described. This temporal algebra introduces the concept of labeled constraints and their temporal operations. Reasoning algorithms for guaranteeing a minimal (and consistent) TCN are specified in Section 4. By using this model, the integration of interval and point-based constraints and management of non-binary constraints are respectively described in Sections 5 and 6. Association of constraints to temporal contexts and management of context-dependent constraints are detailed in Section 7. Finally, Section 8 concludes.

## 2. Basic Temporal Concepts

Temporal reasoning deals with reasoning on temporal constraints. The syntax and semantics of constraints are defined by an underlying temporal algebra, which is the basis for performing the reasoning processes. A temporal algebra can be defined according to the following elements:

- *Temporal primitive* (or variable) *'$x_i$'*, usually time-points ($t_i$) or intervals ($I_i$).

- *Interpretation domain D* for primitives $x_i$. The interpretation domain represents the time line.

Time points are instantiated on D ($t_i \in$ D), and temporal intervals can be modelled as pairs of ending time points that can be instantiated on D: $I_i = (I_i^-, I_i^+)$, $I_i \in$ DxD, $I_i^- \leq I_i^+$.

- *Temporal constraints* between primitives, where each constraint relates *n* primitives: $c_{1,2..n}(x_1, x_2, ..., x_n)$. As particular cases, the 'empty constraint' $\{\varnothing\}$ is named the Inconsistent-Constraint and *'U'* is the Universal-Constraint. *Unary-constraints* restrict the interpretation domain D for variables. They are not usually used in symbolic algebras, where an infinite domain is assumed. *Binary-constraints* are temporal constraints between two variables ($x_i$ $c_{ij}$ $x_j$), and n-ary-constraints represent temporal constraints among *n* variables. By default, binary constraints are assumed in this paper. We can also have *qualitative* (relative relation) or *quantitative* (metric relation) constraints, as well as disjunctive ($c_{ij}$ is a set of disjunctive basic constraints, $|c_{ij}| \geq 1$) or non-disjunctive constraints.

- *Operations between constraints.* Mainly, Temporal Composition ($\otimes$), Temporal Intersection ($\oplus$), Temporal Union ($\cup_T$), and Temporal Inclusion ($\subseteq_T$).

A temporal problem is specified by a set of *n* variables X= $\{x_i\}$, an interpretation domain D and a finite set of temporal constraints between variables $\{(x_i$ $c_{ij}$ $x_j)\}$. A temporal problem gives rise to a *Temporal Constraint Network* (TCN) which can be represented as a directed graph where nodes represent temporal primitives ($x_i$) and labeled-directed edges represent the binary constraints between them ($c_{ij}$). The Universal Constraint *U* is not usually represented in the graph, and each direct edge (representing $c_{ij}$) between $x_i$ and $x_j$ implies an inverse one (representing $c_{ji}$) between $x_j$ and $x_i$. According to the underlying Temporal Algebra, we mainly have IA-TCNs based on the Interval Algebra (Allen, 1983), PA-TCNs based on the Point Algebra (Vilain et al., 1986), or Metric-TCNs based on the Metric Point Algebra (Dechter et al., 1991; Dean & McDermott, 1987). In this later case, disjunctive metric point-based constraints give rise to a Temporal Constraint Satisfaction Problem (TCSP) (Dechter et al., 1991).

Reasoning on temporal constraints can be seen as a Constraint Satisfaction Problem (CSP). An *instantiation* of the variables X is a n-tuple ($v_1$, $v_2$, $v_3$, ...,$v_n$) / $v_i \in$ D which represents the assignments of values $\{v_i\}$ to variables $\{x_i\}$: ($x_1$=$v_1$, $x_2$=$v_2$, ...,$x_n$=$v_n$). A (global) *solution* of a TCN is a consistent instantiation of the variables X in their domains such that all TCN constraints are satisfied. A value *v* is a *consistent (or feasible) value* for $x_i$ if there exists a TCN solution in which $x_i$=v. The set of all feasible values of a variable $x_i$ is the *minimal domain* for the variable. A constraint ($x_i$ $c_{ij}$ $x_j$) is *consistent* if there exists a solution in which ($x_i$ $c_{ij}$ $x_j$) holds. A constraint $c_{ij}$ is *minimal* iff it consists only of consistent elements (or feasible values) that is, those which are satisfied by some interpretation of TCN constraints. A TCN is *minimal* iff all its constraints are minimal.

A TCN is *consistent* (or *satisfiable*) iff it has at least one solution. Freuder (1982) generalizes the notion of consistency as: 'a network is *k-consistent* iff (given any instantiation of any k-1 variables satisfying all the direct constraints among those variables) there exists at least one instantiation of any $k_{th}$ variable such that the k values taken together satisfy all the constraints among the k variables'. As consequences: (*i*) all (k-1)-length paths in the network are consistent, (*ii*) for each pair or nodes, there exists an interpretation that satisfies each (k-1)-length path between them, and (*iii*) each sub-TCN of k-nodes is consistent. As particular cases, 1-consistency, 2-consistency and 3-consistency are called node-consistency, arc-consistency and path-consistency, respectively (Mackworth, 1977; Montanari, 1974).

*Path-consistency* is a common concept in constraint networks. From Montanari (1974) and Mackworth (1977), 'a path of k-length through nodes $(x_1, x_2, ..., x_k, x_j)$ is path-consistent iff for any value $v_1 \in d_1$ and $v_j \in d_j$ such that $(x_1 = v_1 \ c_{1j} \ x_j = v_j)$ holds, there exists a sequence of values $v_2 \in d_2$, $v_3 \in d_3$, ..., $v_k \in d_k$ such that $(v_1 \ c_{i2} \ v_2)$, $(v_2 \ c_{23} \ v_3)$,....., and $(v_k \ c_{k,j} \ v_j)$ hold'. A TCN is *path-consistent* iff all its paths are consistent. Moreover, Montanari (1974) proves that to ensure path-consistency it suffices to check every 2-length path. Thus, path-consistency and 3-consistency are equivalent concepts. Alternatively, Meiri (1996) outlines a path of k-length $(x_i, x_1, x_2, ..., x_k, x_j)$ is path-consistent iff $c_{ij} \subseteq_T (c_{i1} \otimes c_{12} \otimes ... \otimes c_{kj})$. However, this definition disregards domain constraints, such that it is equivalent to the former definition if variable domains are infinite or the TCN is also node and arc-consistent, as the usual case in symbolic algebras. In metric algebras, path-consistency usually assumes node and arc-consistency. Therefore, taking into account that it is only necessary to test 2-length paths to assure path-consistency, a TCN is path-consistent iff $\forall c_{ij}, c_{ik}, c_{kj} \subseteq TCN$, $c_{ij} \subseteq_T (c_{ik} \otimes c_{kj})$. This condition gives rise to the more usual path-consistent algorithm: the Transitive Closure Algorithm (TCA) which imposes local 3-consistency in each sub-TCN of 3 nodes, such that all 2-length paths become consistent paths (Mackworth, 1977; Montanari , 1974). The TCA algorithm will obtain an equivalent path-consistent TCN if it exists. Otherwise, it fails.

$$\forall c_{ij}, c_{ik}, c_{kj} \subseteq TCN: c_{ij} \leftarrow c_{ij} \oplus (c_{ik} \otimes c_{kj})$$

A network is *strong k-consistent* iff the network is j-consistent for all $j \leq k$ (Freuder, 1982). An n-consistent TCN is a consistent TCN, and a strong n-consistent TCN is a minimal TCN. Alternatively, Dechter (1992) introduces the concepts of local and global consistency: A partial instantiation of variables $(x_1 = v_1, x_2 = v_2, ..., x_k = v_k) / 1 \leq k < n$ is *locally consistent* if it satisfies all the constraints among these variables. A subTCN is *globally consistent* if any locally consistent instantiation of the variables in the subTCN can be extended to a consistent instantiation of all TCN. A *globally consistent* TCN is one in which all its subTCNs are globally consistent. Thus, a TCN is strong n-consistent iff it is globally consistent (Dechter, 1992).

The first reasoning task on a TCN is to determine whether the TCN is consistent. If the TCN is consistent, we can then obtain the minimal-TCN, all TCN solutions (by assuming a discrete and finite model of time), only one solution, a partial solution (consistent instantiation of a subset of TCN variables, which is a part of a global solution), etc.

Deductive closure, or propagation, is one of the basic reasoning algorithms. The *closure* process is a deductive process on a TCN, where new derived constraints are deduced from the explicitly asserted ones by means of the composition ($\otimes$) and intersection ($\oplus$) operations. Thus, the process of determining the consistency and the minimality of a TCN is related to a *sound* and *complete* closure process (Vilain et al., 1986). Alternatively, CSP-based methods (with several heuristic search criteria) are also used for guaranteeing consistency and obtaining TCN solutions. In this paper, we are mainly interested in TCN closure processes.

Determining the consistency of a general-case TCN is NP-hard, and Minimal TCNs can be obtained by a polynomial number of consistency processes (Vilain et al., 1986). Particularly, Dechter, Meiri and Pearl (1991) showed that determining consistency and obtaining a minimal disjunctive metric TCN can be achieved in $O(n^3 l^e)$, where '*n*' is the number of TCN nodes, '*e*' is the number of explicitly asserted (input) constraints, and '*l*' is the maximum number of intervals in an input constraint. However, specific levels of k-consistency can guarantee consistency and obtain a minimal TCN, depending on the TCN topology or the underlying temporal algebra. For example, path-consistency guarantees consistency and obtains a minimal non-disjunctive metric TCN (Dechter et al., 1991). The path-consistency TCA Algorithm has an $O(n^3)$ cost (Allen, 1983; Vilain, Kautz & Van Beek, 1986). However, assuring path-consistency can become a complex task in disjunctive metric-TCNs if the variable domain D is large or continuous. As was stated by Dechter, Meiri and Pearl (1991), the number of intervals in $c_{ij} \otimes c_{jk}$ is upper bounded by $|c_{ij}|x|c_{jk}|$. Thus, the total number of disjuncts (subintervals) in a path-consistent TCN might be exponential in the number of disjuncts per constraints in the initial (input) TCN. Schwalb and Dechter (1997) call this the *fragmentation problem*, which does not appear in non-disjunctive metric TCNs. Thus, the TCA algorithm is $O(n^3 R^3)$ in disjunctive metric-TCNs if time is not dense (Dechter et al., 1991), where the range *'R'* is the maximum difference between the lowest and highest number specified in any input constraints.

## 3. A Labeled Temporal Algebra

The main elements of the point-based disjunctive metric temporal algebra are (Dechter et al., 1991):

- Time-point $(t_i)$ as primitive variable. A continuous variable domain (like Q or $\Re$) is usually assumed.

- Each temporal constraint $c_{ij} \subseteq U$ is a finite set of *l* mutually exclusive subdomains (or subintervals) of D.

    $c_{ij} \equiv \{[d^-_1\ d^+_1], [d^-_2\ d^+_2], ...., [d^-_k\ d^+_k], ....., [d^-_l\ d^+_l]\}$ ,        where $d^-_k \le d^+_k$ and $d^-_k, d^+_k \in D$,

    and disjunctively restricts the temporal distance between two time-points, $t_i$ and $t_j$:

    $$t_j - t_i \in \{[d^-_1\ d^+_1], [d^-_2\ d^+_2], ....., [d^-_l\ d^+_l]\},$$

    meaning that $(d^-_1 \le t_j - t_i \le d^+_1) \lor .... \lor (d^-_l \le t_j - t_i \le d^+_l)$. Similar conditions can be applied to open $(d^-_k\ d^+_k)$ and semi-open intervals $(d^-_k\ d^+_k], [d^-_k\ d^+_k)$. The Universal-Constraint *U* is $\{(-\infty\ +\infty)\}$. Unary constraints restrict the associated subdomain of a time-point $t_i \in \{[d^-_1\ d^+_1], [d^-_2\ d^+_2], ....., [d^-_l\ d^+_l]\}$. A special time-point $T_0$ is usually included, which represents 'the beginning of the world' (usually, $T_0$=0). Thus, each unary constraint on $t_i$ can be represented as a binary one between $t_i$ and $T_0$:

    $t_i - T_0 \in \{[d^-_1\ d^+_1], [d^-_2\ d^+_2], ..... ,[d^-_l\ d^+_l]\} \equiv t_i \in [d^-_1, d^+_1] \lor t_i \in [d^-_2, d^+_2] \lor, ..., \lor t_i \in [d^-_i, d^+_i]$

    and, by default: $\forall t_i, (T_0 \{[0\ \infty)\}\ t_i)$.

- The algebra operations, mainly $\otimes$, $\oplus$, $\cup_T$ and $\subseteq_T$. From (Meiri, 1996), given two temporal constraints $S=\{[d_{S}^-_i, d_{S}^+_i]\}$ and $T=\{[d_{T}^-_j, d_{T}^+_j]\}$,

    $S \otimes T = \{d_k / \exists d_i \in S \land \exists d_j \in T / d_k = d_i + d_j\}$.

    That is, $\forall [d_{S}^-_i, d_{S}^+_i] \in S, \forall [d_{T}^-_j, d_{T}^+_j] \in T, \cup_T \{[d_{S}^-_i + d_{T}^-_j, d_{S}^+_i + d_{T}^+_j]\}$. Here, resulting subdomains in $S \otimes T$ may not be pairwise disjoint. Therefore, some additional processing may be required to compute a disjoint subdomain set.

    $S \oplus T = \{d_k / d_k \in S \land d_k \in T\}$. That is, the set-intersection of their subdomains.

    $S \cup_T T = \{d_k / d_k \in S \lor d_k \in T\}$, as the set-union of their subdomains.

    $S \subseteq_T T$ = iff $\forall d_k \in S, \exists d_k \in T$.

On the basis of the point-based disjunctive metric temporal algebra and its operations, we introduce a *labeled* point-based disjunctive metric temporal algebra, which gives rise to a labeled-TCN.

## 3.1 Labeled Constraints and Inconsistent Label Sets

An *elemental constraint* (*ec*) is *one* disjunct in a disjunctive constraint. Similar terms are atomic, basic or canonical constraints. However, let's use this term due to the special structure of *labeled elemental constraints* which are introduced further on. Thus, a disjunctive constraint $c_{ij}$ can be considered as a disjunctive set of $l$ mutually exclusive elemental constraints $\{ec_{ij,k}\}$.

$$ec_{ij,k} = [d^-_{ij,k} \ d^+_{ij,k}] \quad / \ \forall i,j,k \ \ d^-_{ij,k} \leq d^+_{ij,k}$$

$$c_{ij} \equiv \{ec_{ij,1}, ec_{ij,2}, ..., ec_{ij,l}\} \subseteq U \quad / \ \forall k,p \in (1,..,l), \ k \neq p, \ (ec_{ij,k} \oplus ec_{ij,p}) = \varnothing$$

**Definition 1 (*Labeled constraints*).** A labeled elemental constraint $lec_{ij,k}$ is an elemental constraint $ec_{ij,k}$ associated to a set of labels *{label$_{ij,k}$}*, where each *label$_{ij,k}$* is a symbol. A labeled constraint *lc$_{ij}$* is a disjunctive set of labeled elemental constraints $\{lec_{ij,k}\}$. That is,

$$lc_{ij} \equiv \{lec_{ij,1}, lec_{ij,2}, ..., lec_{ij,l}\}, \ \text{ where}$$

$$lec_{ij,k} = (ec_{ij,k}\{label_{ij,k}\}), \ \text{and} \ \ \{label_{ij,k}\} \equiv \{label_1, label_2, ..., label_s\} \ \text{is a set of symbols.} \lozenge$$

Each *label* in a labeled-TCN can be considered as a unique symbol. The following cases can occur:

i)   If an *input* (or explicitly asserted) constraint *lc$_{ij}$* has only one elemental constraint, that is, only one disjunct, this elemental constraint has the label 'R$_0$'. The labeled *Universal-Constraint* is $\{U_{\{R0\}}\}$. In a given TCN, the set of all elemental constraints labeled with 'R$_0$' is the *'common context'*. Thus, the label R$_0$ represents the set of elemental constraints which have no other alternatives (disjuncts). All elemental constraints labeled only with R$_0$ should hold since they have no other alternative disjuncts.

ii)  If an *input* constraint *lc$_{ij}$* has more than one elemental constraint, each elemental constraint lec$_{ij,k} \in$ lc$_{ij}$ has a single and exclusive label associated to it ($|\{label_{ij,k}\}|=1$). Thus, each label in the TCN represents bi-univocally an elemental constraint in an explicitly asserted constraint.

iii) Each derived elemental constraint (obtained by combining ($\otimes_{lc}$) or intersecting ($\oplus_{lc}$) two labeled elemental constraints) has a set of labels associated to it. This set of labels is obtained from the label sets associated to the combined (or intersected) labeled elemental constraints. It will be detailed in the later specification of operations ($\otimes_{lc}$, $\oplus_{lc}$) in Section 3.2. In consequence, the label set associated to a derived elemental constraint represents the conjunctive support-set of explicitly asserted elemental constraints that imply this derived elemental constraint.

Let's see a simple example on labeled constraints, which was introduced by Dechter, Meiri and Pearl (1991).

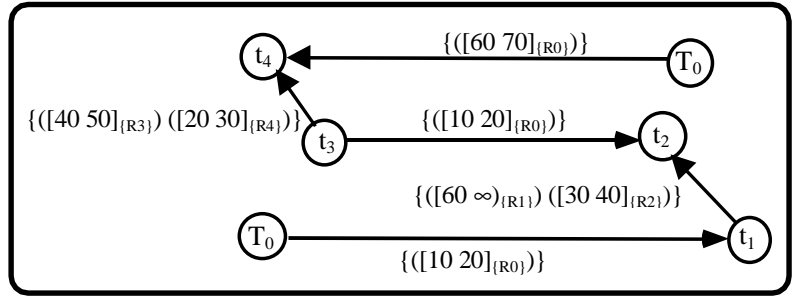

Figure 1: The labeled point-based disjunctive metric TCN of the Example 1

**Example 1:** *"John goes to work either by car [30'-40'], or by bus (at least 60'). Fred goes to work either by car [20'-30'], or in a carpool [40'-50']. Today John left home (t1) between 7:10 and 7:20, and Fred arrived (t4) at work between 8:00 and 8:10. We also know that John arrived (t2) at work about 10'-20' after Fred left home (t3)."*

In this example, we have the disjunctive labeled constraints in Figure 1, where $T_0$ represents the initial time (7:00) and where the granularity is in minutes. A label 'R$_0$' is associated to elemental constraints belonging to constraints with only one disjunct. In constraints with more than one, mutually exclusive disjuncts, each disjunct is labeled with an exclusive label $R_n$ (n>0). Thus,

- The label $R_0$ is associated to "John left home between 7:10 and 7:20", "Fred arrived at work between 8:00 and 8:10", and "John arrived at work about 10'-20' after Fred left home". This is the *common context*.

- The label $R_1$ is associated to "John goes by bus", and $R_2$ to "John goes by car".

- The label $R_3$ is associated to "Fred goes in a carpool", and $R_4$ to "Fred goes by car".

**Definition 2 (*Inconsistent-Label-Sets*).** An *Inconsistent-Label-Set* (I-L-Set) is a set of labels {label$_i$} and represents a set of overall inconsistent elemental constraints. That is, they cannot all simultaneously hold. ◊

**Theorem 1.** Any label set that is a superset of an I-L-Set is also an I-L-Set. The proof is obvious. If a set of elemental constraints is inconsistent, any superset of it is also inconsistent. ◊

**Definition 3.** Elemental constraints {lec$_{ij,k}$} of an input disjunctive constraint lc$_{ij}$ are pairwise disjoint. Thus, each 2-length set of labels from each pair of {lec$_{ij,k}$} is added to the set of I-L-Sets. That is, for each input constraint lc$_{ij} \equiv$ {lec$_{ij,1}$, lec$_{ij,2}$, ..., lec$_{ij,l}$}, where lec$_{ij,k} \equiv$ (ec$_{ij,k}$ {label$_{ij,k}$}) and |{label$_{ij,k}$}|=1:

$$\forall k,p \in (1,..,l) \, / \, k \neq p, \quad \text{I-L-Sets} \leftarrow \text{I-L-Sets} \cup (\{\text{label}_{ij,k}\} \cup \{\text{label}_{ij,p}\}) \, ◊$$

In the example of Figure 1, {R$_1$ R$_2$} and {R$_3$ R$_4$} are I-L-Sets. Other I-L-Sets existing in a labeled TCN will be detected in the reasoning processes later detailed in Section 4.

## 3.2 Operations on Labeled Constraints

The following points define the main operations on labeled constraints.

### 3.2.1 TEMPORAL INCLUSION $\subseteq_{LC}$

The temporal inclusion operation $\subseteq_{lc}$ should take into account the inclusion of temporal intervals and the inclusion of associated label sets:

$lec_{ij.k} \subseteq_{lc} lec_{ij.p} = (ec_{ij.k} \{label_{ij.k}\}) \subseteq_{lc} (ec_{ij.p} \{label_{ij.p}\}) =_{def} ec_{ij.k} \subseteq_{T} ec_{ij.p} \ \wedge \ \{label_{ij.k}\} \subseteq \{label_{ij.p}\}$.

### 3.2.2 TEMPORAL UNION $\cup_{LC}$

Operation $\cup_{lc}$ performs the disjunctive temporal union of labeled constraints as the set-union of their elemental constraints. However, all labeled elemental constraints whose associated labels are I-L-Sets should be rejected.

$lc_{ij} \cup_{lc} lc'_{ij} =_{def} \forall lec_{ij.k} \in lc_{ij}, \ \cup_{lc} [\{lec_{ij.k}\} \ lc'_{ij}]$ , where

$\cup_{lc} [\{lec_{ij.k}\} \ lc'_{ij}] = (ec_{ij.k}\{label_{ij.k}\}) \cup_{lc} lc'_{ij} =_{def}$

> Inconsistent($\{label_{ij.k}\}$) :  $lc'_{ij}$

> $\exists lec_{ij.p} \in lc'_{ij} \ / \ lec_{ij.p} \subseteq_{lc} lec_{ij.k} : lc'_{ij}$           *(s₁)*

> Other :  $(\{lc'_{ij}\} \cup \{lec_{ij.k}\}) - (\{lec_{ij.p}\}, \ \forall lec_{ij.p} \in lc'_{ij} \wedge lec_{ij.k} \subseteq_{lc} lec_{ij.p})$    *(s₂)*.

The function *Inconsistent({label$_{ij.k}$})* returns *true* if the set $\{label_{ij.k}\}$ is a I-L-Set or a superset of any existing I-L-Set (*Theorem 1*). Otherwise, it returns *false*:

> ***Inconsistent({label$_{ij.k}$})*** $=_{def}$

>> If $\exists \{label_s\} \in$ Inconsistent-Label-Sets $/ \ \{label_s\} \subseteq \{label_{ij.k}\}$ Then *True* Else *False*.

The operation $\cup_{lc}$ simplifies the resulting constraint. Equal or less-restricted elemental constraints with equal or bigger associated label sets are removed. For instance:

$$\{([10 \ 30]_{\{R1 \ R3 \ R5 \ R9\}}), ([40 \ 40]_{\{R6 \ R7\}})\} \cup_{lc} \{([10 \ 20]_{\{R1 \ R3\}}), ([40 \ 40]_{\{R6 \ R7 \ R8\}})\} =$$

$$\{([10 \ 20]_{\{R1 \ R3\}}), ([40 \ 40]_{\{R6 \ R7\}})\}.$$

In the resulting constraint, $([10 \ 30]_{\{R1 \ R3 \ R5 \ R9\}})$ and $([40 \ 40]_{\{R6 \ R7 \ R8\}})$ are eliminated, as examples of the cases *s₁* and *s₂*, respectively. That is, $([10 \ 20]_{\{R1 \ R3\}}) \subseteq_{lc} ([10 \ 30]_{\{R1 \ R3 \ R5 \ R9\}})$ and $([40 \ 40]_{\{R6 \ R7\}}) \subseteq_{lc} ([40 \ 40]_{\{R6 \ R7 \ R8\}})$. These simplifications can seem counter-intuitive. However, note that the label set associated to each derived-labeled elemental constraint represents the support set (composed of input elemental constraints) from which the derived-labeled elemental constraint is obtained. Thus, only the minimal associated label set should be represented, for reason of efficiency. Moreover, the more labels are in the associated label set $\{label_{ij.k}\}$, the elemental constraint $(ec_{ij.k})$ should be equal or more restricted.

### 3.2.3 TEMPORAL COMPOSITION $\otimes_{LC}$

Operation $\otimes_{lc}$ performs the temporal composition of labeled constraints. It is based in the operation $\otimes$ of the underlying disjunctive metric point-based algebra.

$$lc_{ij} \otimes_{lc} lc_{jk} =_{def} \forall lec_{ij,p} \in lc_{ij}, \forall lec_{jk,q} \in lc_{jk} \cup_{lc} [ (ec_{ij,p} \otimes ec_{jk,q} \{label_{ij,p}\} \cup \{label_{jk,q}\})].$$

For instance: $\{([0\ 10]_{\{R1\}}), ([20\ 30]_{\{R2\}})\} \otimes_{lc} \{([100\ 200]_{\{R3\}}), ([300\ 400]_{\{R4\}})\} =$

$$\{([320\ 430]_{\{R4\ R2\}}), ([300\ 410]_{\{R4\ R1\}}), ([100\ 210]_{\{R3\ R1\}}), ([120\ 230]_{\{R3\ R2\}})\}.$$

Note that elemental constraints in a labeled derived constraint may not be pairwise disjoint. However, these labeled derived elemental constraints cannot be simplified. This is related to the *fragmentation problem* of the disjunctive metric algebra (Schwalb & Dechter, 1997). We have that each derived-labeled elemental constraint should have its own associated label set. In the example, $(([320\ 430]_{\{R4\ R2\}}), ([300\ 410]_{\{R4\ R1\}}))$ cannot be simplified to $([300\ 430]_{\{R4\ R2\ R1\}})$ since each subinterval depends on a different set of labels (that is, on a different support-set of elemental constraints). If the label set $\{R_4\ R_2\}$ becomes an I-L-Set, only $([320\ 430]_{\{R4\ R2\}})$ should be removed. On the other hand, if $[300\ 410]$ becomes an inconsistent interval between the implied time points, only $\{R_4\ R_1\}$ should be asserted as an I-L-Set.

### 3.2.4 TEMPORAL INTERSECTION $\oplus_{LC}$

Operation $\oplus_{lc}$ performs the temporal intersection of labeled constraints and is based on the operation $\oplus$.

$$lc_{ij} \oplus_{lc} lc'_{ij} =_{def} \forall lec_{ij,k} \in lc_{ij}, \forall lec_{ij,p} \in lc'_{ij}, \cup_{lc} [lec_{ij,k} \oplus_{lc} lec_{ij,p}]$$

where, $lec_{ij,k} \oplus_{lc} lec_{ij,p} =_{def}$

If $ec_{ij,k} \oplus ec_{ij,p} = \varnothing$ Then $\{\varnothing\}$       ;The Inconsistent-Constraint is returned.

Else   $[(ec_{ij,k} \oplus ec_{ij,p}) (\{label_{ij,k}\} \cup \{label_{ij,p}\})]$

As example:

$$\{([0\ 10]_{\{R1\}}), ([20\ 25]_{\{R2\}})\} \oplus_{lc} \{([0\ 30]_{\{R3\}}), ([40\ 50]_{\{R4\}})\} = \{([20\ 25]_{\{R3\ R2\}}), ([0\ 10]_{\{R3\ R1\}})\}$$

In the operations $\otimes_{lc}$ and $\oplus_{lc}$, the label set $\{label_{ij,r}\}$ associated to each derived labeled-elemental constraint $(ec_{ij,r})$ is obtained from the set-union of labels associated to combined $(\otimes_{lc})$ or intersected $(\oplus_{lc})$ labeled-elemental constraints. Therefore, $\{label_{ij,r}\}$ represents the support set (composed of input elemental constraints) that implies the derived elemental constraint $(ec_{ij,r})$.

**Definition 4.** A set of I-L-Sets is complete if it represents all inconsistent sets of TCN elemental constraints. A set of I-L-Sets is sound if each I-L-Set represents an inconsistent set of elemental constraints. ◊

**Theorem 2.** Assuming a complete and sound set of I-L-Sets, a labeled elemental constraint is consistent iff it has an associated label set which is not an I-L-Set. The proof is trivial, since the label set associated to each labeled elemental constraint represents its support-set. ◊

**Theorem 3.** Assuming a complete and sound set of I-L-Sets, no inconsistent labeled elemental constraint is obtained in operations $\otimes_{lc}$ and $\oplus_{lc}$.

***Proof:*** The operations $\otimes_{lc}$ and $\oplus_{lc}$ use the operation $\cup_{lc}$ to obtain their results. This operation $\cup_{lc}$ rejects all labeled elemental constraints whose associated labels are I-L-Sets. Thus, all elemental constraints derived in operations $\otimes_{lc}$ and $\oplus_{lc}$ are consistent (*Theorem 2*). ◊

### 3.3 Distributive Property $\otimes_{lc}$ Over $\oplus_{lc}$ in Disjunctive Labeled Constraints

Operations $\otimes$ and $\oplus$ are distributive (i.e.: $\otimes$ distributes over $\oplus$) in non-disjunctive metric TCN, but this property does not hold in disjunctive metric constraints. Dechter, Meiri and Pearl (1991) show the following example. Given the disjunctive metric constraints:

$$a = \{[0\ 1], [10\ 20]\}, \quad b = \{[25\ 50]\}, \quad c = \{[0\ 30], [40\ 50]\},$$

we have:

$$(a \otimes (b \oplus c) = \{[25\ 31], [35\ 70]\} \qquad (a \otimes b) \oplus (a \otimes c) = \{[25\ 70]\}.$$

Thus, clearly $(a \otimes (b \oplus c) \neq (a \otimes b) \oplus (a \otimes c)$. However, the distributive property holds for operations $\otimes_{lc}$ and $\oplus_{lc}$ in labeled TCN.

**Theorem 4.** By using labeled constraints and I-L-Sets, $\otimes_{lc}$ distributes over $\oplus_{lc}$.

***Proof:*** Let's consider the labeled constraints $lc_i$, $lc_j$ and $lc_k$. Thus,

$$(lc_i \otimes_{lc} lc_j) \oplus_{lc} (lc_i \otimes_{lc} lc_k)$$

can be expressed, according to the definition of operation $\otimes_{lc}$, as:

$$(\forall lec_p \in lc_i, \forall lec_q \in lc_j, \cup_{lc}[(lec_p \otimes_{lc} lec_q)]) \quad \oplus_{lc} \quad (\forall lec_r \in lc_i, \forall lec_s \in lc_k, \cup_{lc}[(lec_r \otimes_{lc} lec_s)]) \quad =$$

$$\forall lec_p \in lc_i, \forall lec_q \in lc_j, \forall lec_r \in lc_i, \forall lec_s \in lc_k \ (\cup_{lc}[(lec_p \otimes_{lc} lec_q)] \oplus_{lc} \cup_{lc}[(lec_r \otimes_{lc} lec_s)])$$

which, according to the definition of $\oplus_{lc}$, can be expressed as:

$$\forall lec_p \in lc_i, \forall lec_q \in lc_j, \forall lec_r \in lc_i, \forall lec_s \in lc_k \ (\cup_{lc}[(lec_p \otimes_{lc} lec_q) \oplus_{lc} (lec_r \otimes_{lc} lec_s)]) \qquad (e1)$$

In this expression, $lec_p$ and $lec_r$ are elemental constraints of the same-labeled constraint $lc_i$. However, the set-union of label sets associated to each pair of elemental constraints in any (input or derived) labeled constraint is an I-L-Set (*Definition 3*). That is, if $lec_p \neq lec_r$, then $\{label_p\} \cup \{label_r\}$ is an I-L-Set. Thus, if $lec_p \neq lec_r$, the label set associated to $(lec_p \otimes_{lc} lec_q) \oplus_{lc} (lec_r \otimes_{lc} lec_s)$ is an I-L-Set. In consequence, $(lec_p \otimes_{lc} lec_q) \oplus_{lc} (lec_r \otimes_{lc} lec_s)$ is rejected in operation $\cup_{lc}$. That is,

$$\forall lec_p \in lc_i, \forall lec_q \in lc_j, \forall lec_r \in lc_i, \forall lec_s \in lc_k \ / \ lec_p \neq lec_r \quad (\cup_{lc}[(lec_p \otimes_{lc} lec_q) \oplus_{lc} (lec_r \otimes_{lc} lec_s)]) = \varnothing.$$

Thus, the above expression (*e1*) results:

$$\forall lec_p \in lc_i, \forall lec_q \in lc_j, \forall lec_s \in lc_k \ (\cup_{lc}[(lec_p \otimes_{lc} lec_q) \oplus_{lc} (lec_p \otimes_{lc} lec_s)]).$$

In this expression, $\otimes_{lc}$ clearly distributes over $\oplus_{lc}$ for elemental constraints (i.e.: non-disjunctive constraints). Therefore:

$$\forall lec_p \in lc_i, \forall lec_q \in lc_j, \forall lec_s \in lc_k \ (\cup_{lc}[(lec_p \otimes_{lc} (lec_q \oplus_{lc} lec_s))]) \ =$$

$$\forall lec_p \in lc_i, \cup_{lc}[lec_p \otimes_{lc} (\forall lec_q \in lc_j, \forall lec_s \in lc_k, \cup_{lc}[lec_q \oplus_{lc} lec_s])] \ = \ lc_i \otimes_{lc} (lc_j \oplus_{lc} lc_k).$$

That is, $\otimes_{lc}$ distributes over $\oplus_{lc}$ for labeled constraints. ◊

For instance, following the previous example:

$$a = \{[0\ 1]_{\{R1\}}, [10\ 20]_{\{R2\}}\}, \quad b = \{[25\ 50]_{\{R0\}}\}, \quad c = \{[0\ 30]_{\{R3\}}, [40\ 50]_{\{R4\}}\}$$

and $\{R_1\ R_2\}$, $\{R_3\ R_4\}$ are I-L-Sets. Thus, we have:

$$(a \otimes_{lc} (b \oplus_{lc} c = \{[0\ 1]_{\{R1\}}, [10\ 20]_{\{R2\}}\} \otimes_{lc} (\{[25\ 50]_{\{R0\}}\} \oplus_{lc} \{[0\ 30]_{\{R3\}}, [40\ 50]_{\{R4\}}\}) =$$

$$\{[0\ 1]_{\{R1\}}, [10\ 20]_{\{R2\}}\} \otimes_{lc} \{[25\ 30]_{\{R3\ R0\}}, [40\ 50]_{\{R4\ R0\}}\} =$$

$$\{[25\ 31]_{\{R1\ R3\ R0\}}, [40\ 51]_{\{R1\ R4\ R0\}}, [35\ 50]_{\{R3\ R2\ R0\}}, [50\ 70]_{\{R4\ R2\ R0\}}\}.$$

Also,

$(a \otimes_{lc} b) \oplus_{lc} (a \otimes_{lc} c) =$

$(\{[0\ 1]_{\{R1\}}, [10\ 20]_{\{R2\}}\} \otimes_{lc} \{[25\ 50]_{\{R0\}}\}) \oplus_{lc}$

$\qquad\qquad (\{[0\ 1]_{\{R1\}}, [10\ 20]_{\{R2\}}\} \otimes_{lc} \{[0\ 30]_{\{R3\}}, [40\ 50]_{\{R4\}}\}) =$

$\{[25\ 51]_{\{R1\ R0\}}, [35\ 70]_{\{R2\ R0\}}\} \oplus_{lc} \{[0\ 31]_{\{R1\ R3\}}, [40\ 51]_{\{R1\ R4\}} [10\ 50]_{\{R2\ R3\}}, [50\ 70]_{\{R2\ R4\}}\} =$

$\cup_{lc} ([25\ 31]_{\{R1\ R3\ R0\}}, [40\ 51]_{\{R1\ R4\ R0\}}, [25\ 50]_{\{R1\ R2\ R3\ R0\}},$

$\qquad\qquad [50\ 51]_{\{R1\ R2\ R4\ R0\}}, [40\ 51]_{\{R1\ R2\ R4\ R0\}}, [35\ 50]_{\{R3\ R2\ R0\}}, [50\ 70]_{\{R4\ R2\ R0\}}).$

However, $\{R_1\ R_2\}$, $\{R_3\ R_4\}$ are I-L-Sets. Thus, $([25\ 50]_{\{R1\ R2\ R3\ R0\}},$ $[50\ 51]_{\{R1\ R2\ R4\ R0\}}, [40\ 51]_{\{R1\ R2\ R4\ R0\}})$ are removed in operation $\cup_{lc}$. Therefore,

$(a \otimes_{lc} b) \oplus_{lc} (a \otimes_{lc} c) = \{[25\ 31]_{\{R1\ R3\ R0\}}, [40\ 51]_{\{R1\ R4\ R0\}}, [35\ 50]_{\{R3\ R2\ R0\}}, [50\ 70]_{\{R4\ R2\ R0\}}\}.$

That is, $(a \otimes_{lc} (b \oplus_{lc} c) = (a \otimes_{lc} b) \oplus_{lc} (a \otimes_{lc} c).$

## 4. Reasoning Algorithms on Labeled Constraints

Several algorithms for reasoning on disjunctive constraints can be applied for the management of labeled temporal constraints, by using the $\otimes_{lc}, \oplus_{lc}, \cup_{lc}$ and $\subseteq_{lc}$ operations. For instance, the well-known Transitive Closure Algorithm, general closure algorithms as in (Dechter, 1992; Dechter et al., 1991; van Beek & Dechter, 1997), CSP-based approaches, etc. However, Montanari (1974) shows that when composition operation distributes over intersection, any path-consistent TCN is also a minimal TCN. From *Theorem 4*, we have that $\otimes_{lc}$ distributes over $\oplus_{lc}$. Thus, application of a path-consistent algorithm on the proposed-labeled TCN will obtain a minimal TCN. Thus, the TCA algorithm could be used as the closure process on labeled constraints, in a similar way as Allen (1983) uses it. However, an *incremental* reasoning process is proposed on the basis of the incremental path-consistent algorithm for non-disjunctive metric constraints described by Barber (1993). An incremental reasoning process is useful when temporal constraints are not initially known but are successively deduced from an independent process; for instance, in an integrated planning and scheduling system (Garrido et al., 1999). The proposed reasoning algorithm is similar to the TCA algorithm. However, updating and closure processes are performed at each new input constraint. Thus, each new input constraint is updated and closured on a previously minimal TCN (Figure 9). Therefore, no further propagation of modified constraints in the closure process is needed. Moreover, the proposed reasoning algorithms will obtain a complete and sound set of I-L-Sets.

The specification of reasoning processes is described in Section 4.1. The properties of these processes will be described later in Section 4.2.

### 4.1 The Updating Process

Given a previous labeled-TCN, composed by a set of nodes $\{n_i\}$, a set of labeled constraints $\{lc_{ij}\}$ among them, and a set of I-L-Sets, the *updating process* of each new $c'_{ij}$ between nodes $n_i$ and $n_j$ constraint is detailed in Figure 2.

---

**Updating ($n_i$ c'$_{ij}$ $n_j$)**          ;c'$_{ij}$≡{ec'$_{ij}$, ec'$_{ij.2}$, ..., ec'$_{ij.l}$}, a disjunctive metric constraint.

lc'$_{ij}$ ← Put-Labels (c'$_{ij}$),    ;An exclusive label is associated to each elemental constraint ec'$_{ij.k}$ in c'$_{ij}$

**If** Consistency-Test (lc$_{ij}$ , lc'$_{ij}$)    ;Consistency test of lc'$_{ij}$. The previously existing constraint between $n_i$ and $n_j$ is lc$_{ij}$. Moreover, new I-L-Sets are detected.

    **Then**  (*Inconsistent Constraint*)

       Return (false)

    **Else**  (*Consistent Constraint*)

       lc$_{ij}$← lc$_{ij}$ ⊕$_{lc}$ lc'$_{ij}$,          lc$_{ji}$ ← Inverse$_{lc}$ (lc$_{ij}$),

       Closure ($n_i$ lc$_{ij}$ $n_j$),  ;Closure algorithm for the updated constraint.

       Return (true)

  **End-If**

**End-Updating**

---

Figure 2: Updating process on labeled constraints

The function *Put-Labels(c'$_{ij}$)* returns a labeled-constraint lc'$_{ij}$≡{lec'$_{ij.1}$, lec'$_{ij.2}$, ..., lec'$_{ij.1}$}, associating an exclusive label to each elemental constraint in c'$_{ij}$. If there is only one disjunct in c'$_{ij}$, the label in the unique elemental constraint is {R$_0$}. Otherwise, each pair of labels in lc'$_{ij}$ is added to the set of I-L-Sets, since elemental constraints in c'$_{ij}$ are pairwise disjoint (*Definition 3*). By using the *Inverse* function on non-labeled constraints, the *Inverse$_{lc}$* function is:

$$\text{Inverse}_{lc} (\{(ec_{ij.k}\{label_{ij.k}\})\}) =_{def} \{(\text{Inverse } (ec_{ij.k}) \{label_{ij.k}\})\}$$

The described updating process is performed *each time* that *one* new input constraint c'$_{ij}$ is asserted on a previous TCN. Thus, an initial TCN with no nodes, no constraints, and no I-L-Sets is assumed (Figure 9). At each new input constraint (c'$_{ij}$), the TCN is incrementally updated and closed. That is, if c'$_{ij}$ is consistent (*Consistency-Test* function), the constraint c'$_{ij}$ is added to the TCN, the closure process (*Closure* function) propagates its effects to all TCN, and the new TCN is obtained. A new updating process can be performed on this new TCN, and so on successively.

### 4.1.1. THE CONSISTENCY-TEST FUNCTION

The Consistency-Test function (Figure 3) is based on the operation ⊕$_{lc}$. A new input constraint lc'$_{ij}$ between nodes $n_i$ and $n_j$ is consistent if it temporally intersects with the previously existing constraint lc$_{ij}$ between these nodes. Moreover, the Consistency-Test function can detect new I-L-Sets:

i)   If the new constraint lc'$_{ij}$ is consistent with the existing constraint lc$_{ij}$, and two elemental constraints ec$_{ij.p}$∈lc'$_{ij}$, ec$_{ij.k}$∈lc$_{ij}$ do not intersect (ec$_{ij.k}$ ⊕ ec$_{ij.p}$=∅), then the label set {label$_{ij.k}$}∪{label$_{ij.p}$} is an I-L-Set and should be added to the current set of I-L-Sets.

ii)   If an existing elemental constraint between nodes $n_i$ and $n_j$ (lec$_{ij.k}$∈lc$_{ij}$) does not intersect with the new constraint lc'$_{ij}$, then {label$_{ij.k}$} is an I-L-Set and should be added to the current set of I-L-Sets.

---

**Consistency-Test** ($lc_{ij}$, $lc'_{ij}$) =

If ($lc_{ij} \oplus_{lc} lc'_{ij}$) = {$\varnothing$}

    Then Return (False)

    Else

        If $\exists lec_{ij,k} \in lc_{ij}$, $\exists lec_{ij,p} \in lc'_{ij}$ / $lec_{ij,k} \oplus_{lc} lec_{ij,p}$={$\varnothing$}

                Then   I-L-Sets ← I-L-Sets ∪ ({$label_{ij,k}$}∪{$label_{ij,p}$}),

    If $\exists lec_{ij,k} \in lc_{ij}$ / $lec_{ij,k} \oplus_{lc} lc'_{ij}$ = {$\varnothing$}

             Then   I-L-Sets ← I-L-Sets ∪ {$label_{ij,k}$},

End-If

Return (True)

**End- Consistency-Test**

Figure 3: Consistency-Test function

For example,

    Consistency-Test ({($[0\ 10]_{\{R1\}}$), ($[20\ 25]_{\{R2\}}$), ($[100\ 110]_{\{Ra\}}$)},

$$\{([0\ 30]_{\{R3\}}), ([40\ 50]_{\{R4\}}), ([-50\ -40]_{\{Rb\}})\}) = \text{True}$$

since

  {{($[0\ 10]_{\{R1\}}$), ($[20\ 25]_{\{R2\}}$), ($[100\ 110]_{\{Ra\}}$)} $\oplus_{lc}$ {($[0\ 30]_{\{R3\}}$), ($[40\ 50]_{\{R4\}}$), ($[-50\ -40]_{\{Rb\}}$)} =

$$\{([20\ 25]_{\{R3\ R2\}}), ([0\ 10]_{\{R3\ R1\}})\} \neq \{\varnothing\}.$$

In this function, the label sets {$R_4\ R_2$}, {$R_4\ R_1$} and {$Ra$} are detected as I-L-Sets and should be added to the current set of I-L-Sets, since:

    {$[20\ 25]_{\{R2\}}$} $\oplus_{lc}$ {$[40\ 50]_{\{R4\}}$}={$\varnothing$},          {$[0\ 10]_{\{R1\}}$}) $\oplus_{lc}$ {$[40\ 50]_{\{R4\}}$})={$\varnothing$},

      {($[100\ 110]_{\{Ra\}}$)} $\oplus_{lc}$ {($[0\ 30]_{\{R3\}}$), ($[40\ 50]_{\{R4\}}$), ($[-50\ -40]_{\{Rb\}}$)}={$\varnothing$}.

Note that {$R_b$} does not need to be detected as an I-L-Set, since the label $R_b$ is not included in the final constraint {($[20\ 25]_{\{R3\ R2\}}$), ($[0\ 10]_{\{R3\ R1\}}$)} to be added to the TCN.

Any superset of an I-L-Set is also an I-L-Set (Theorem 1). Moreover, note that {$R_4\ R_2$}, {$R_4\ R_1$} do not need to be added to the set of I-L-Sets, since the label $R_4$ is not included in the final constraint. Therefore, the following simplifications can also be performed each time a new I-L-Set is added to the current set of I-L-Sets. These simplifications do not modify the results of reasoning processes, but minimize the size of the set of I-L-Sets and improve its management efficiency.

i)    No new I-L-Set that is superset of an existing I-L-Set is added to the set of I-L-Sets.

ii)   If an existing I-L-Set is superset of the new I-L-Set, then the existing I-L-Set is removed.

iii)  No new I-L-Set that contains a label of $lc'_{ij}$, which does not appear in the labeled constraint ($lc_{ij} \oplus_{lc} lc'_{ij}$) to be added to the TCN, should be added to the set of I-L-Sets.

Let's see an example of the updating and consistency-test processes. Let's take the labeled-TCN that results from Example 1 once the following constraints have been updated and closured:

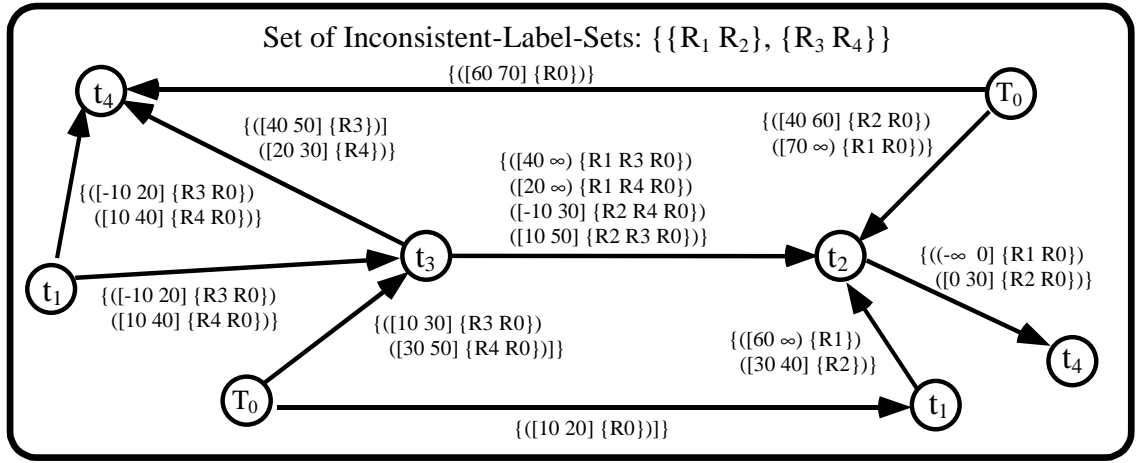

Figure 4: The resulting labeled-TCN of Figure 1 before updating ($t_3$ {[10 20]} $t_2$)

($t_1$ {[60 ∞]}$_{R1}$, [30 40]$_{R2}$} $t_2$), ($t_3$ {[40 50]$_{R3}$, [20 30]$_{R4}$} $t_4$), ($T_0$ {[10 20]$_{R0}$} $t_1$), ($T_0$ {[60 70]$_{R0}$} $t_4$).

The resulting labeled-TCN is shown in Figure 4 and the set of I-L-Set is {{$R_1$ $R_2$}, {$R_3$ $R_4$}}. Now, we update ($t_3$ {[10 20]$_{R0}$} $t_2$). The previously existing constraint between $t_3$ and $t_2$ is (Figure 4):

$$\{([40\ ∞]_{\{R1\ R3\ R0\}})\ \ ([20\ ∞]_{\{R1\ R4\ R0\}}),\ ([-10\ 30]_{\{R2\ R4\ R0\}})\ \ ([10\ 50]_{\{R2\ R3\ R0\}})\}$$

The Consistency-Test function performs:

$$\{[10\ 20]_{\{R0\}}\}\ \oplus_{lc}\ \{([40\ ∞]_{\{R1\ R3\ R0\}})\ \ ([20\ ∞]_{\{R1\ R4\ R0\}}),\ ([-10\ 30]_{\{R2\ R4\ R0\}})\ \ ([10\ 50]_{\{R2\ R3\ R0\}})\}\ =$$

$$\{[20\ 20]_{\{R1\ R4\ R0\}},\ [10\ 20]_{\{R2\ R0\}}\ [\varnothing]_{\{R1\ R3\ R0\}}\}\ \neq\ \{\varnothing\} \qquad (e1)$$

Thus, (t2-t3∈{[10 20]$_{\{R0\}}$}) is consistent. Moreover, {$R_1$ $R_3$ $R_0$} is detected as an I-L-Set. The elemental constraints associated to {$R_1$ $R_3$ $R_0$} are an inconsistent set of disjuncts that cannot hold simultaneously. That is:

*"If today John left home between 7:10 and 7:20 ($R_0$), Fred arrived at work between 8:00 and 8:10 ($R_0$) and John arrived at work about 10'-20' after Fred left home ($R_0$), then it is impossible for John to have gone by bus ($R_1$) and Fred to have gone in a carpool ($R_3$)."*

The set of I-L-Sets obtained in the reasoning process can be considered as special *derived constraints*, which express the inconsistency of a set of *input* elemental constraints. For instance, the I-L-Set {$R_0$ $R_1$ $R_3$} represents (Figure 1):

$$\neg\ (\ (T_0\ [10\ 20]\ T_1)\ \wedge\ (T_3\ [10\ 20]\ T_2)\ \wedge\ (T_0\ [60\ 70]\ T_4)\ \wedge\ (T_3\ [40\ 50]\ T_4)\ \wedge\ (T_1\ [60\ ∞]\ T_2)).$$

This expression is a non-binary constraint. This type of constraints could be represented as a disjunctive linear constraint, as Jonsson and Bäckström (1996), Stergiou and Koubarakis (1996) show. However, input elemental constraints should be represented in derived constraints to be able *to derive* these inconsistent sets of input elemental constraints. In this model, this is done by means of the label sets associated to labeled elemental constraints.

## 4.2 The Closure Process

The closure process (Figure 5) is applied each time a new input constraint (lc'$_{ij}$) is updated, such that the effects of lc'$_{ij}$ are propagated to all TCN.

---

**Closure (n$_i$ lc$_{ij}$ n$_j$)**
(* *First loop: Closure* $n_i \rightarrow n_j \rightarrow n_k$ *)
    $\forall n_k \in$ TCN / lc$_{jk} \neq \{U_{\{R0\}}\}$:
        lc$_{ik} \leftarrow$ lc$_{ik} \oplus_{lc}$ (lc$_{ij} \otimes_{lc}$ lc$_{jk}$),    lc$_{ki} \leftarrow$ Inverse(lc$_{ik}$)
(* *Second loop: Closure* $n_j \rightarrow n_i \rightarrow n_l$ *)
    $\forall n_l \in$ TCN / lc$_{il} \neq \{U_{\{R0\}}\}$:
        lc$_{jl} \leftarrow$ lc$_{jl} \oplus_{lc}$ (Inverse(lc$_{ij}$) $\otimes_{lc}$ lc$_{il}$),    lc$_{lj} \leftarrow$ Inverse(lc$_{jl}$)
(* *Third loop*: Closure $n_l \rightarrow n_i \rightarrow n_j \rightarrow n_k$*)[1]
    $\forall n_l, n_k \in$ TCN / lc$_{li} \neq \{U_{\{R0\}}\}$, lc$_{jk} \neq \{U_{\{R0\}}\}$:
        lc$_{lk} \leftarrow$ lc$_{lk} \oplus_{lc}$ (lc$_{li} \otimes_{lc}$ lc$_{ij} \otimes_{lc}$ lc$_{jk}$),    lc$_{kl} \leftarrow$ Inverse(lc$_{lk}$)
**End-Closure**

---

Figure 5: The closure process on labeled constraints

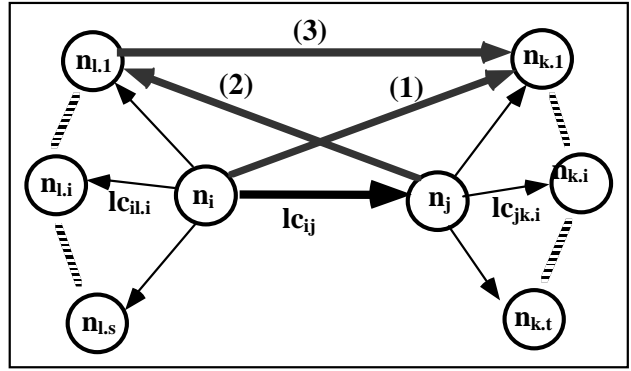

Figure 6: Loops in the Closure Process

The closure process has three loops (Figure 6). In these loops the process obtains:

i)    Derived constraints lc$_{ik}$ between n$_i$ and any node n$_k$, if n$_k$ is previously connected with n$_j$ (edge 1 of Figure 6).

ii)    Derived constraints lc$_{lj}$ between n$_j$ and any node n$_l$, if n$_l$ is previously connected with n$_i$ (edge 2 of Figure 6).

---

[1] This loop could be simplified as:

    (*$n_l \rightarrow n_i \rightarrow n_k$*):  $\forall n_l, n_k \in$ TCN / lc$_{li} \neq \{U_{\{R0\}}\}$, lc$_{jk} \neq \{U_{\{R0\}}\}$:    lc$_{lk} \leftarrow$ lc$_{lk} \oplus_{lc}$ (lc$_{li} \otimes_{lc}$ lc$_{ik}$),  or as

    (*$n_l \rightarrow n_j \rightarrow n_k$*):  $\forall n_l, n_k \in$ TCN / lc$_{li} \neq \{U_{\{R0\}}\}$, lc$_{jk} \neq \{U_{\{R0\}}\}$:    lc$_{lk} \leftarrow$ lc$_{lk} \oplus_{lc}$ (lc$_{lj} \otimes_{lc}$ lc$_{jk}$)

since lc$_{ik}$ (or lc$_{lj}$) has already been closured in the first (or in the second loop). Moreover, the efficiency of the third loop can be improved if only modified constraints in the first (or in the second loop) are considered.

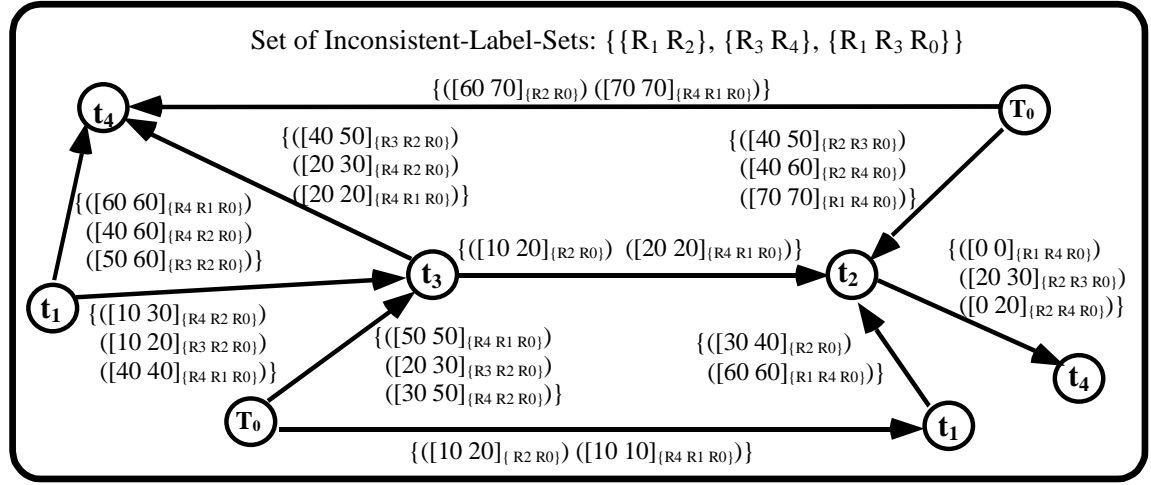

Figure 7: The Labeled-Minimal TCN of the Example 1

iii) Derived constraints $lc_{lk}$ between any pair of nodes $n_l$ and $n_k$, if $n_l$ and $n_k$ are previously connected with $n_i$ and $n_j$ respectively (edge 3 of Figure 6).

Let's see the previous Example 1 represented in Figure 1 and Figure 4, when the consistent constraint (expression *e1*)

$$(t_3 \; \{[20 \; 20]_{\{R1 \; R4 \; R0\}}, \; [10 \; 20]_{\{R2 \; R0\}}\} \; t_2)$$

is closured. In the first loop of the closure process, we have:

$lc_{30} \leftarrow lc_{30} \oplus_{lc} (\{[20 \; 20]_{\{R1 \; R4 \; R0\}}, [10 \; 20]_{\{R2 \; R0\}}\} \otimes_{lc} lc_{20} =$

$\quad \{[-30 \; -10]_{\{R3 \; R0\}} \; [-50 \; -30]_{\{R4 \; R0\}}\} \oplus_{lc}$

$\quad\quad\quad (\{[20 \; 20]_{\{R1 \; R4 \; R0\}}, [10 \; 20]_{\{R2 \; R0\}}\} \otimes_{lc} \{[-60 \; -40]_{\{R2 \; R0\}} \; (-\infty \; -70]_{\{R1 \; R0\}}\}) =$

$\quad \{[-30 \; -10]_{\{R3 \; R0\}}\} \; [-50 \; -30]_{\{R4 \; R0\}}\} \oplus_{lc}$

$\quad\quad\quad \{[-40 \; -20]_{\{R1 \; R2 \; R4 \; R0\}}, (-\infty \; -50]_{\{R1 \; R4 \; R0\}} \; [-50 \; -20]_{\{R2 \; R0\}} \; (-\infty \; -50]_{\{R1 \; R2 \; R0\}}\}.$

However, $\{\{R_1 \; R_2\}, \{R_3 \; R_4\} \; \{R_0 \; R_1 \; R_3\}\}$ are I-L-Sets. No labeled elemental constraints whose associated label set is a superset of these I-L-Sets will be derived (*Theorem 3*). Thus:

$lc_{30} \leftarrow \{[-30 \; -10]_{\{R3 \; R0\}}\} \; [-50 \; -30]_{\{R4 \; R0\}}\} \oplus_{lc} \{(-\infty \; -50]_{\{R1 \; R4 \; R0\}} \; [-50 \; -20]_{\{R2 \; R0\}} \} =$

$\quad\quad\quad \{(-30 \; -20]_{\{R2 \; R3 \; R0\}} \; [-50 \; -50]_{\{R4 \; R1 \; R0\}} \; [-50 \; -30]_{\{R4 \; R2 \; R0\}}\}.$

Similarly,

$lc_{31} \leftarrow lc_{31} \oplus_{lc} (\{[20 \; 20]_{\{R1 \; R4 \; R0\}} \; [10 \; 20]_{\{R2 \; R0\}}\} \otimes_{lc} lc_{21} =$

$\quad\quad\quad \{[-20 \; -10]_{\{R3 \; R2 \; R0\}} \; [-40 \; -40]_{\{R4 \; R1 \; R0\}} \; [-30 \; -10]_{\{R4 \; R2 \; R0\}}\}$

$lc_{34} \leftarrow lc_{34} \oplus_{lc} (\{[20 \; 20]_{\{R1 \; R4 \; R0\}}, [10 \; 20]_{\{R2 \; R0\}}\} \otimes_{lc} lc_{24} =$

$\quad\quad\quad \{[40 \; 50]_{\{R3 \; R2 \; R0\}} \; [20 \; 30]_{\{R4 \; R2 \; R0\}} \; [20 \; 20]_{\{R4 \; R1 \; R0\}}\}.$

After the second and third loops, the final labeled-TCN is obtained (Figure 7). The final set of I-L-Sets is $\{\{R_1 \; R_2\}, \{R_3 \; R_4\} \; \{R_0 \; R_1 \; R_3\}\}$. These sets represent *all* sets of mutually inconsistent input-elemental constraints that exist in the TCN of Figure 1.

### 4.3  Properties of Reasoning Algorithms

In this section, the main properties of the proposed reasoning algorithms are described.

**Theorem 5**. The proposed updating and closure processes (Sections 4.1 and 4.2) guarantee a consistent TCN if they are applied on a previous minimal (and consistent) TCN.

***Proof***: The updating constraint $lc'_{ij}$ is asserted in the TCN if it is consistent with the previous minimal constraint $lc_{ij}$ (Consistency-Test function). ◊

**Theorem 6**. The proposed closure algorithm obtains a path-consistent TCN, if it is applied over a previous minimal TCN.

***Proof:*** This was detailed by Barber (1993) for non-disjunctive TCNs and it is applied here to labeled TCNs. We have:

i)  No derived constraint can exist between a pair of nodes if no path between them combines the asserted constraint $lc_{ij}$.

ii)  The closure process computes a derived constraint between any pair of nodes $(n_l, n_k)$ that become connected by a path across the closured constraint $lc_{ij}$. Let's assume an existing path between the nodes $n_{x1}, n_{y1}$ that includes $lc_{ij}$:

$$n_{x1}, n_{x2}, n_{x3}, ........, n_x, (n_j\ lc_{ij}\ n_j), n_y......, n_{y2}, n_{y1}$$

such that a derived constraint between $n_{x1}\ n_{y1}$ should be computed. However, a minimal constraint between $(n_{x1}, n_i)$ and between $(n_j, n_{y1})$ should already exist in the previous minimal TCN. In consequence, a derived constraint between $(n_{x1}, n_{y1})$ is computed in the third loop of the process.

iii)  If the previous TCN is minimal, all possible derived constraints that can exist between any pair of nodes $(n_l, n_k)$ are already computed in the constraint $lc'_{lk}$ derived between these nodes in the proposed closure process. In the third loop, this process obtains:

$$lc'_{lk} = lc_{lk} \oplus_{lc} (lc_{li} \otimes_{lc} lc_{ij} \otimes_{lc} lc_{jk}).$$

Let's suppose there exists another path between $(n_l, n_k)$ across the updated $lc_{ij}$ constraint: $(n_l, n_p, n_i, n_j, n_q, n_k)$. This path computes another derived constraint between $(n_l, n_k)$:

$$lc''_{lk} = lc_{lk} \oplus_{lc} (lc_{lp} \otimes_{lc} lc_{pi} \otimes_{lc} lc_{ij} \otimes_{lc} lc_{jq} \otimes_{lc} lc_{qk}).$$

However, since the previous TCN is minimal, the previously existing minimal constraints $lc_{li}$ and $lc_{jk}$ imply $(lc_{lp} \otimes_{lc} lc_{pi})$ and $(lc_{jq} \otimes_{lc} lc_{qk})$, respectively. That is, $lc_{li} \subseteq_{lc}(lc_{lp} \otimes_{lc} lc_{pi})$ and $lc_{jk} \subseteq_{lc}(lc_{jq} \otimes_{lc} lc_{qk})$ Thus, $lc''_{lk}$ is also implicitly implied by $lc'_{lk}$ ($lc'_{lk}\subseteq_{lc}lc''_{lk}$). Here, we have assumed the associative property for $\otimes_{lc}$, which is obvious from its definition.

iv)  Derived constraints obtained in the closure process do not need to be closured again if the previous TCN is minimal. That is, no constraint in the TCN would become more restricted if derived constraints were also closured. Let suppose $lc_{lk}$ is modified in the third loop of closure process:

$$lc'_{lk} = lc_{lk} \oplus_{lc} (lc_{li} \otimes_{lc} lc_{ij} \otimes_{lc} lc_{jk})$$

such that it should be propagated to the $(n_l, n_k, n_p)$ subTCN (Figure 8). Thus, the following derived constraints should be obtained:

$$lc'_{lp} = lc_{lp} \oplus_{lc} (lc'_{lk} \otimes_{lc} lc_{kp}) \qquad\qquad lc'_{pq} = lc_{pq} \oplus_{lc} (lc_{pl} \otimes_{lc} lc'_{lk}).$$

For constraint $lc'_{lp}$, we have,

$$lc'_{lp} = lc_{lp} \oplus_{lc} (lc'_{lk} \otimes_{lc} lc_{kp}) = lc_{lp} \oplus_{lc} ((lc_{lk} \oplus_{lc} (lc_{li} \otimes_{lc} lc_{ij} \otimes_{lc} lc_{jk})) \otimes_{lc} lc_{kp}).$$

However, since $\otimes_{lc}$ distributes over $\oplus_{lc}$,

$$lc'_{lp} = lc_{lp} \oplus_{lc} ((lc_{lk} \otimes_{lc} lc_{kp}) \oplus_{lc} (lc_{li} \otimes_{lc} lc_{ij} \otimes_{lc} lc_{jk} \otimes_{lc} lc_{kp})).$$

Since the previous TCN is minimal, the minimal constraints $lc_{pi}$ and $lc_{pj}$ should previously exist, such that $lc_{lp} \subseteq_{lc} (lc_{lk} \otimes_{lc} lc_{kp})$ and $lc_{jp} \subseteq_{lc} (lc_{jk} \otimes_{lc} lc_{kp})$. Thus,

$$lc'_{lp} \subseteq_{lc} lc_{lp} \oplus_{lc} (lc_{li} \otimes_{lc} lc_{ij} \otimes_{lc} lc_{jp}).$$

However, in the third loop of the closure process, the following derived constraint is computed:

$$lc''_{lp} = lc_{lp} \oplus_{lc} (lc_{li} \otimes_{lc} lc_{ij} \otimes_{lc} lc_{jp}).$$

Thus, $lc'_{lp}$ is already represented in the obtained constraint $lc''_{lp}$ (that is, $lc''_{lp} \subseteq_{lc} lc'_{lp}$). In a similar way,

$$lc''_{pq} = lc_{pq} \oplus_{lc} (lc_{pi} \otimes_{lc} lc_{ij} \otimes_{lc} lc_{jq})$$

is also obtained in the proposed closure process, such that $lc''_{pq} \subseteq_{lc} lc'_{pq}$.

Therefore, each derived constraint (any combinable path across $lc_{ij}$) between any pair of nodes in the TCN is computed, so that the closure process obtains a path-consistent TCN. ◊

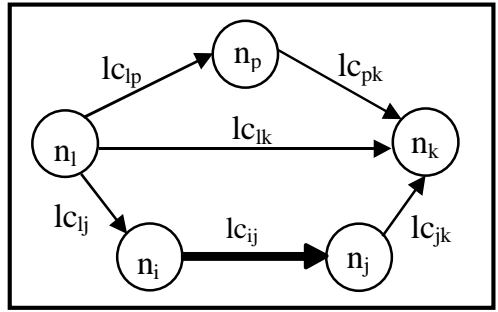

Figure 8: $lc_{lk}$ is also propagated to $lc_{lp}$ and $lc_{pq}$

**Theorem 7.** The proposed reasoning processes obtain a minimal TCN, if the previous TCN is a minimal TCN.

***Proof***: Montanari (1974) shows that when composition distributes over intersection (i.e.: $\otimes$ distributes over $\oplus$), any path-consistent TCN is also a minimal TCN). This is the case in non-disjunctive metric TCNs (Dechter et al., 1991). In our case, $\otimes_{lc}$ distributes over $\oplus_{lc}$ (*Theorem 4*) and the closure process obtains a path consistent TCN (*Theorem 6*). Therefore, the proposed reasoning processes also obtain a minimal TCN. ◊

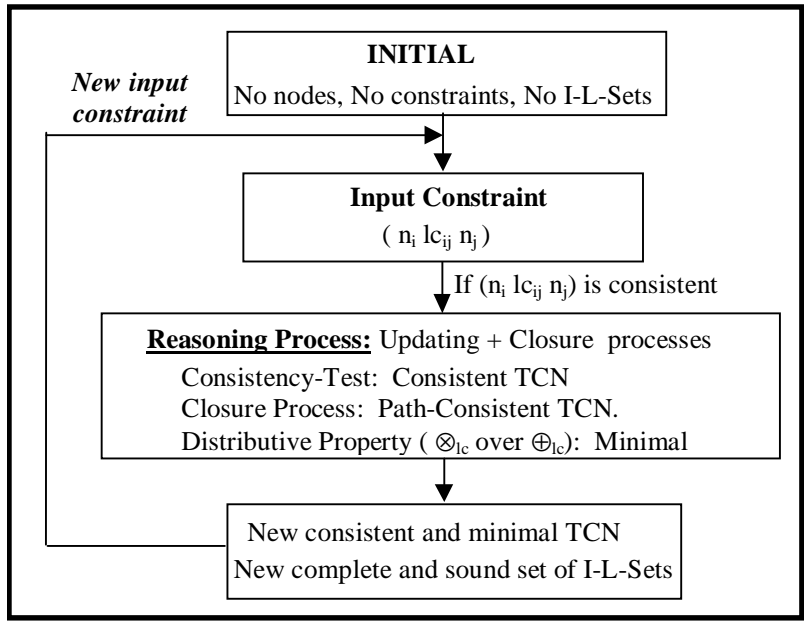

Figure 9: An incremental reasoning process

**Theorem 8.** At each updating process, reasoning algorithms obtain a complete and a sound new set of I-L-Sets (*Definition 4*), if they are applied on a previous minimal TCN and a previous sound and complete set of I-L-Sets.

***Proof:***

*i)* The new set of I-L-Sets is complete. The consistency test of the updated constraint $lc'_{ij}$ obtains all possible new I-L-Sets that can appear when $lc'_{ij}$ is added to the TCN, except those I-L-Sets which are related to the mutual exclusion of the disjuncts in $lc'_{ij}$ (which are determined in the Put-Label function):

  *a)* No new I-L-Sets can appear in which some label of $lc'_{ij}$ does not participate. Otherwise, they would have been detected in a previous updating process, since the previous set of I-L-Sets is assumed complete. Thus, some label of $lc'_{ij}$ should always participate in any new I-L-Set that appears when $lc'_{ij}$ is updated.

  *b)* All new I-L-Sets (in which some label of $lc'_{ij}$ participates) are detected in the consistency test of $lc'_{ij}$. Let's assume that a new and undetected I-L-Set exists $\{R_k, R_1, R_2, \ldots, R_p\}$ in which some new elemental constraint $ec_{k\{Rk\}} \in lc'_{ij}$ takes part. Thus, the elemental constraints associated to $\{R_1, R_2, \ldots, R_p\}$ compute a derived elemental constraint $ec_x$ between the nodes $n_i$ and $n_j$:

$$(ec_{x\ \{R1, R2, \ldots, Rp\}}) \quad / \quad (ec_{x\ \{R1, R2, \ldots, Rp\}}) \oplus_{lc} (ec_{k\{Rk\}}) = \varnothing$$

  This elemental constraint $ec_x$ is already represented in the previously existing constraint $lc_{ij}$

between $n_i$ and $n_j$ since the previous TCN is minimal[2]. Thus, $ec_k \oplus ec_x = \varnothing$, such that the I-L-Set $\{R_k, R_1, R_2, ....., R_p\}$ is detected in the consistency test of $lc'_{ij}$. In conclusion, all new inconsistent sets of elemental constraints in which $lc'_{ij}$ participates are detected and no other new I-L-Sets can exist. Therefore, the new set of I-L-Sets is complete if the previous set of I-L-Sets is complete.

*ii)* The new set of I-L-Sets is sound. All new I-L-Sets obtained represent inconsistent sets of elemental constraints. This is trivial, given the consistency test function. ◊

In conclusion, the proposed reasoning algorithms obtain a minimal (and consistent) TCN if they are applied to a previous minimal-TCN (Figure 9). Therefore, the reasoning algorithms guarantee TCN consistency and obtain a minimal TCN and a complete and sound set of I-L-Sets at each new input assertion.

### 4.4 Global Labeled-Consistency

In a minimal (binary) disjunctive network, every subnetwork of size two is globally consistent (Dechter, 1992). Therefore, any local consistent instantiation of a subset of two variables can be extended to a full consistent instantiation. However, to assure that a local consistent instantiation of a subset of more that two variables is overall consistent, the partial instantiation should be propagated on the whole TCN (van Beek, 1991). Thus, assembling a TCN solution can become a costly propagation process in disjunctive TCNs, even though a minimal TCN was used. The proposed reasoning processes maintain a complete and sound set of I-L-Sets (*Theorem 8*). Thus, we can deduce if a locally consistent set of elemental constraints is overall consistent by means of label sets associated to labeled elemental constraints and the set of I-L-Sets. Specifically, we can deduce whether any locally consistent instantiation of k variables ($1<k<n$) is overall consistent. Let's see the following example, which is based on a previous one proposed by Dechter, Meiri and Pearl (1991):

*Example 2:* *"Dave goes walking to work in [25' 50']. John goes to work either by car [10' 30'], or by bus [45' 60']. Fred goes to work either by car [15' 20'], or in a carpool [35' 40'], or walking [55' 60']. Today, they all left their home between 6:50 and 7:50 (at $t_1$, $t_2$ and $t_3$ time-points), and arrived at work at just the same time ($t_4$) before 8:00."*

Here, we have the following labeled disjunctive constraints where, $T_0$ represents the initial time (6:50) and granularity is in minutes:

$t_1 - T_0 \in \{[0\ 60]_{R0}\}$,     $t_2 - T_0 \in \{[0\ 60]_{R0}\}$,     $t_3 - T_0 \in \{[0\ 60]_{R0}\}$,     $t_4 - T_0 \in \{[0\ 70]_{R0}\}$,

$t_4 - t_1 \in \{[25\ 50]_{R0}\}$,     $t_4 - t_2 \in \{[10\ 30]_{R1}, [45\ 60]_{R2}\}$,     $t_4 - t_3 \in \{[15\ 20]_{R3}, [35\ 40]_{R4}, [55\ 60]_{R5}\}$.

The minimal TCN of Example 2 is represented in Figure 10. Here, the binary constraints between each time-point and $T_0$ represent unary constraints and restrict interpretation domains for variables ($t_1$, $t_2$, $t_3$, $t_4$). Obviously, this minimal TCN is not a globally consistent TCN. For instance, instantiations {(t₁=0), (t₂=0), (t₃=0)} are consistent with the existing constraints involved among (T₀, t₁, t₂, t₃), but this partial solution cannot be extended to the overall TCN.

---

[2] The elemental constraint $ec_x$ is already represented in an explicit way, or by means of another elemental constraint $ec_y$ ($ec_y \subseteq_T ec_x \{label_y\} \subseteq \{R_1, R_2, ....., R_p\}$) due to the simplification process performed in the operation $\cup_k$. In both cases, $ec_k \oplus ec_x = \varnothing$, $ec_k \oplus ec_y = \varnothing$.

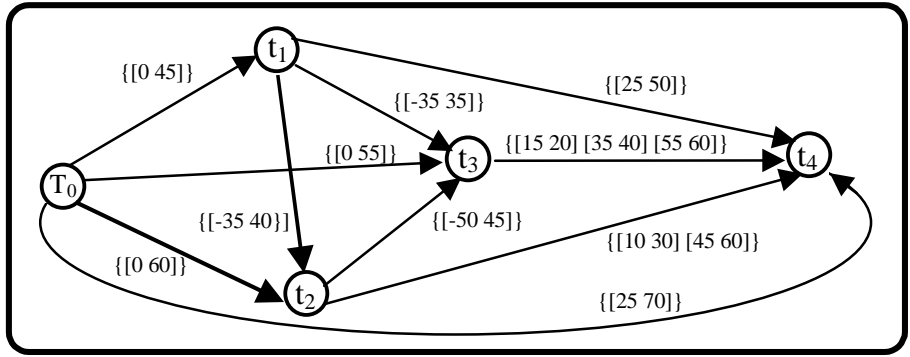

Figure 10: Minimal TCN of Example 2

Let's consider the TCN with labeled constraints. For reasons of simplicity, we only denote the labeled constraints among (T₀, t₁, t₂, t₃):

$$(T_0 \{[5\ 45]_{\{R0\ R5\}}, [0\ 45]_{\{R0\ R4\}}, [0\ 45]_{\{R0\ R3\}}\}\ t_1),$$

$$(T_0 \{[0\ 25]_{\{R2\ R0\}}, [5\ 60]_{\{R1\ R0\ R4\}}, [25\ 60]_{\{R1\ R0\ R5\}}, [0\ 60]_{\{R1\ R0\ R3\}}\}\ t_2),$$

$$(T_0 \{[25\ 55]_{\{R0\ R2\ R3\}}, [0\ 15]_{\{R0\ R5\}}, [0\ 35]_{\{R0\ R1\ R4\}}, [5\ 55]_{\{R0\ R1\ R3\}}, [5\ 35]_{\{R0\ R2\ R4\}}\}\ t_3),$$

$$(t_1 \{([-5\ 35]_{\{R0\ R2\}}, [-40\ 5]_{\{R0\ R1\}}\}\ t_2),$$

$$(t_1 \{[-15\ 15]_{\{R0\ R4\}}, [-35\ -5]_{\{R0\ R3\}}, [5\ 35]_{\{R0\ R5\}}\}\ t_3),$$

$$(t_2 \{[5\ 30]_{\{R1\ R0\ R4\}}, [-45\ -25]_{\{R2\ R0\ R3\}}, [25\ 50]_{\{R1\ R0\ R5\}}, [-15\ 10]_{\{R1\ R0\ R3\}}, [-25\ -5]_{\{R2\ R0\ R4\}}, [-5\ 15]_{\{R2\ R0\ R5\}}\}\ t_3).$$

The set of I-L-Sets is {{R₁ R₂} {R₃ R₄} {R₃ R₅} {R₄ R₅}}. From this labeled TCN and the set of I-L-Sets, we can deduce that instantiations {(t₁=0), (t₂=0), (t₃=0)} are not overall consistent. These instantiations are not *locally* consistent with the *labeled constraints* in the subTCN (T₀, t₁, t₂, t₃): All label sets associated to possible simultaneous fulfillment of

$$(T_0 \{[0\ 0]\}\ t_1),\ (T_0 \{[0\ 0]\}\ t_2)\text{ and }(T_0 \{[0\ 0]\}\ t_3)$$

are I-L-Sets. That is, all label sets in the Cartesian product

$$\{\{R_0\ R_4\}\ \{R_0\ R_3\}\}\quad X\quad \{\{R_2\ R_0\}\ \{R_1\ R_0\ R_3\}\}\quad X\quad \{\{R_0\ R_5\}\ \{R_0\ R_1\ R_4\}\}$$

are I-L-Sets. Thus, the set of I-L-Sets can be used to deduce consistency of a set of labeled elemental constraints and to obtain a *globally consistent* labeled-TCN.

**Theorem 9**. Let's assume a labeled-TCN of *n* nodes (and the corresponding complete and sound set of I-L-Sets) and a local set of *k* $(1{\le}k{\le}\binom{n}{2})$ labeled elemental constraints in the TCN, each one of which is between any pair of nodes:

$$\{lec_1, lec_2, ....., lec_k\} \equiv \{(ec_1\ \{label_1\}), (ec_2\ \{label_2\}), ..., (ec_k\ \{label_k\})\}.$$

The local set of labeled elemental constraints $\{lec_1, lec_2, \ldots, lec_k\}$ is overall consistent iff the set-union of their associated label sets ($\cup_{i=1,k}\{label_i\}$) is not an I-L-Set.

***Proof:*** The label set ($\cup_{i=1,k}\{label_i\}$) is the support-set of the simultaneous fulfillment of $\{lec_1, lec_2, \ldots, lec_k\}$. Moreover, the set of I-L-Sets is complete and sound with respect to overall TCN (*Theorem 8*), such that any label set not in the set of I-L-Set is overall consistent. Therefore (*Theorem 2*), ($\cup_{i=1,k}\{label_i\}$) and $\{lec_1, lec_2, \ldots, lec_k\}$ are overall consistent iff $\cup_{i=1,k}\{label_i\}$ is not an I-L-Set. ◊

**Definition 5 (*Labeled-consistency[3]*):** Let's assume a labeled-TCN of n nodes (and the corresponding complete set of I-L-Sets) and a set of k ($1 \leq k \leq \binom{n}{2}$) constraints, each one of which is between any pair of nodes in the TCN:

$$\{c_{ij}\} \; / \; 1 \leq i \leq n, \; 1 \leq j \leq n, \; i \neq j.$$

The set of constraints $\{c_{ij}\}$ is *labeled-consistent* with respect to the nodes involved in these constraints, iff:

i) For each constraint $c_{ij}$, there exists an elemental labeled constraint $elc_{ij,x}$ between $(n_i, n_j)$ in the TCN such that $elc_{ij,x}$ satisfies $c_{ij}$. That is: $\forall c_{ij}, \exists elc_{ij,x} \in lc_{ij} \; / \; c_{ij} \oplus ec_{ij,x} \neq \varnothing$.

ii) The resulting set of the union of label sets associated to these elemental labeled constraints (which satisfy $\{c_{ij}\}$) is not an I-L-Set: $\bigcup_{\forall c_{ij}}\{label_{ij,x}\}$ is not an I-L-Set. Note that this is the condition of Theorem 9. ◊

**Theorem 10.** Let's assume a labeled-TCN of n nodes (and the corresponding complete set of I-L-Sets) and a set of k ($1 \leq k \leq \binom{n}{2}$) constraints, each one of which is between any pair of nodes in the TCN:

$$\{c_{ij}\} \; / \; 1 \leq i \leq n, \; 1 \leq j \leq n, \; i \neq j.$$

The set of constraints $\{c_{ij}\}$ is overall consistent iff $\{c_{ij}\}$ is labeled-consistent with respect to the nodes involved in constraints $\{c_{ij}\}$.

***Proof:*** The proof is trivial according to *Definition 5* and *Theorem 9*. We have that the set of constraints $\{c_{ij}\}$ is consistent iff there exists a local set of elemental constraints in the TCN $\{elc_{ij,x}\}$ that makes $\{c_{ij}\}$ labeled-consistent (*Definition 5*). Thus, the local set $\{elc_{ij,x}\}$ is consistent (*Theorem 9*), such that $\{c_{ij}\}$ is also consistent. ◊

For instance, we can determine whether any pair of constraints $c'_{ij}$ and $c'_{kl}$ can hold simultaneously (that is, they are overall consistent) if:

$$\exists elc_{ij,x} \in lc_{ij} \, / \, c'_{ij} \oplus ec_{ij,x} \neq \varnothing \; \wedge \; \exists elc_{kl,y} \in lc_{kl} \, / \, c'_{kl} \oplus ec_{kl,y} \neq \varnothing \; \wedge \; \{label_{ij,x}\} \cup \{label_{kl,y}\}$$

is not an I-L-Set.

Moreover, any local instantiation of any k-1 ($1 < k \leq n$) variables $\{t_1 = v_1, t_2 = v_2, \ldots, t_{(k-1)} = v_{(k-1)}\}$ can be extended to a global solution if:

$$\exists elc_{10,x} \in lc_{10} \, / \, v_1 \in ec_{10,x}, \ldots \ldots , \exists elc_{(k-1)0,y} \in lc_{(k-1)0} \, / \, v_{(k-1)} \in ec_{10,x},$$

where $lc_{i0}$ is the constraint between $n_i$ and $T_0$, and $\{label_{10,x}\} \cup \{label_{20,y}\} \cup \ldots \cup \{label_{(k-1)0,y}\}$ is not and I-L-Set.

---

[3] We need to introduce the concept of *labeled-consistency* since it is a different concept from the *consistency* concept.

For instance, in Example 2 of Figure 10, the partial instantiation $\{(t_1=0), (t_2=5), (t_3=5)\}$ is consistent. We have:

$([0\ 45]_{\{R0\ R3\}}) \in lc_{10}\ /\ 0 \in [0\ 45],$ $([0\ 60]_{\{R1\ R0\ R3\}}) \in lc_{20}\ /\ 5 \in [0\ 60],$ $([5\ 55]_{\{R0\ R1\ R3\}}) \in lc_{30}\ /\ 5 \in [5\ 55],$

and $\{R_0\ R_3\} \cup \{R_0\ R_1\ R_3\} \cup \{R_0\ R_1\ R_3\} = \{R_0\ R_1\ R_3\}$ is not an IL-Set. Thus, this partial solution can be extended to a global solution. For instance, $\{(t_1=0), (t_2=5), (t_3=5), (t_4=25)\}$.

Therefore, a labeled-TCN can be considered as a *globally labeled-consistent* TCN. That is, on the basis of the concepts introduced by Dechter (1992):

**Definition 6. (*Local Labeled-consistency*):** A partial instantiation of variables $(1 \leq k < n)$ $\{t_1=v_1, t_2=v_2, ..., t_k=v_k\}$ is local labeled-consistent if it is *labeled-consistent* with respect to $(T_0, t_1, t_2, ..., t_k)$ nodes. This also holds for k=n. ◊

**Definition 7. (*Global Labeled-consistency*):** A labeled sub-TCN (with the global set of I-L-Sets) is *global labeled-consistent* if any partial instantiation of variables in the sub-TCN, which is local labeled-consistent, can be extended to the overall TCN. A globally labeled-consistent TCN is one in which all its sub-TCNs are globally labeled-consistent. ◊

**Theorem 11.** At each new assertion, the proposed reasoning processes obtain a *globally labeled-consistent* TCN, if they are applied on a previous minimal TCN and a previous sound and complete set of I-L-Sets.

***Proof:*** The proof is trivial according to the previous definitions (Definition 6 and Definition 7) and to the properties of the reasoning processes (Theorem 7 and Theorem 8). Any partial instantiation in any subTCN, which is labeled-consistent with respect to the nodes involved in the partial instantiation, is overall consistent (*Theorem 10*). ◊

Similar expressions can be made for *k-labeled-consistency* and *strong k-labeled-consistency* on the basis of the concepts provided by Freuder (1982). Therefore, the set of I-L-Sets in a labeled-TCN provides a useful way to assure whether a local instantiation of variables can be part of a global solution. Moreover, Freuder (1982) shows that in a strong k-consistent TCN, consistent instantiations of variables of any subnetwork of size k can be found in a backtrack-free manner and in any variable ordering. This is also a consequence of the decomposability (Montanari, 1974; Dechter et al., 1991) or globally consistency (Dechter, 1992) properties. Obviously, this feature also holds for labeled TCNs.

## 4.5 Analysis of Temporal Complexity

Let's analyze the computational cost of the proposed reasoning processes. These processes are, basically, an *incremental* path-consistent algorithm (Barber, 1993). At each updating process of a new input constraint on a TCN with *n* nodes, the computational cost of updating and closure processes is bounded by $'n^2\ (O(\otimes_{lc}) + O(\oplus_{lc}))'$. In the proposed reasoning process, the path-consistent algorithm obtains a minimal disjunctive metric TCN. This is possible due to the management of labeled constraints, associated label sets, and I-L-Sets. Thus, the complexity of reasoning processes is mainly due (instead of a complex closure process) to the management of complex data structures (labeled constraints, associated label sets, and I-L-Sets). That is, the complexity of the proposed reasoning processes is mainly due to the complexity of operations $\otimes_{lc}$ and $\oplus_{lc}$.

The computational cost of $\otimes_{lc}$ and $\oplus_{lc}$ depends on the number of elemental constraints in labeled constraints, the size of associated label sets, and the size of I-L-Sets in the previous minimal labeled TCN. Let *'n'* be the number of nodes, *'l'* the maximum number of disjuncts (or labels) in input constraints, and *'e'* the number of updated input constraints in the previous TCN. The maximum number of labels in the TCN is $l*e$, since each disjunct in each updated input labeled constraint has its own, unequivocal label. Moreover, any I-L-Set can have as maximum *one* label from each input labeled constraint $lc_{ij}$, since: *(i)* elemental constraints in $lc_{ij}$ are pairwise disjoint, such that each pair of labels in $lc_{ij}$ is added to the set of I-L-Sets, and *(ii)* any superset of an existing I-L-Set is also an I-L-Set. Thus, the maximum number of labels in any I-L-Set is *e*. Furthermore, each label in an I-L-Set should be from a different input labeled constraint. There are *e* input labeled constraints, and each input labeled constraint has as maximum *l* labels. Thus, the maximum number of I-L-Sets of q-length $(1 \leq q \leq e)$ is $(\binom{e}{q}\ l^q)$.

Therefore, the number of i-length $(1 \leq i \leq e)$ I-L-Sets is $\Sigma_{i=1,e}\ (\binom{e}{i}\ l^i) = O(2^e\ l^e)$. However, any superset of an I-L-Set is already known as inconsistent, such that supersets are not stored in the set of I-L-Sets. Thus, the number of I-L-Set is bounded by $O(l^e)$. Additionally, we also have $e*\binom{e}{2})$ I-L-Sets of 2-length, since the *l* disjuncts in each updated constraint are mutually exclusive among them. Similarly, the maximum number of associated label sets is also bounded by $O(l^e)$, each one with a maximum of *e* labels. Thus, the number of elemental constraints (or labeled subintervals) in any labeled constraint is bound by $O(l^e)$, since each elemental constraint in a labeled constraint has its own associated label set.

According to these parameters, the computational cost of each updating process is bounded by $O(n^2\ l^{3e})$. The recovery process of constraints has a constant cost, since a minimal-TCN is always maintained. The computational cost of the proposed algorithms agreed with the computational cost inherent to the problem of the management of disjunctive metric constraints (Dechter, 1991). In fact, the closure process could be considered as an integrated management of the $l^e$ alternative non-disjunctive TCNs in which a disjunctive TCN can be split, as it is shown by Dechter, Meiri and Pearl (1991). It should be noted that *l* can be bounded in some typical problems like scheduling, where usually l≤2 (Garrido et al., 1999), or by restricting domain size (range or granularity) in metric algebras. On the other hand, several improvements can be made on the described processes. For example, an efficient management of label sets has a direct influence on the efficiency of the reasoning processes. Thus, each label set (for instance, $\{R_3\ R_5\ R_8\}$) can be considered as a unidimensional array of bits, which is the binary representation of an integer number (for instance $(2^3+2^5+2^8)$). Therefore, each associated label set is represented by a number and the set of I-L-Sets becomes a set of numbers. Matching and set-union processes on label sets in operations $\otimes_{lc}$ and $\oplus_{lc}$ can be efficiently performed by means of operations on integer numbers with a *constant* cost. Therefore, the computational cost can be bounded by $O(n^2\ l^{2e})$.

Other alternative implementations are under study. Two different approaches exist for temporal constraint management (Brusoni et al., 1997; Yampratoom, Allen, 1993; Barber, 1993). The first approach is to maintain a closured TCN by recomputing the TCN at each new input constraint and making the derived constraints explicit. Here, queries are answered in constant time, although this implies a high spatial cost. The second approach is to explicitly represent only input constraints, such that the spatial requirements are minimum. However, further computation is needed at query time and when consistency of each new input constraint is tested. The proposed reasoning methods hold in the first approach, which seems more appropriate for problems where queries on the TCN are more usual tasks than updating processes.

In addition, the proposed reasoning algorithms obtain a sound and complete set of I-L-Sets and a *globally labeled-consistent* TCN. Regrettably, assembling a solution in a labeled TCN, although backtrack free, is also costly due to the exponential number of I-L-Sets. However, these features offer the capability of representing and managing special types of non-binary disjunctive constraints (later detailed in Section 6).

Other reasoning algorithms for query processes on a non-closured TCN, as well as CSP approaches can be defined on the basis of the labeled temporal algebra described. Less expensive algorithms can be applied on labeled constraints by using the specified operations $\otimes_{lc}$, $\oplus_{lc}$, $\cup_{Tlc}$ and $\subseteq_{Tlc}$. For instance, the TCA algorithm as is applied by Allen (1983), and the k-consistency algorithms like those described in (Cooper, 1990; Freuder, 1978). Moreover, a minimal TCN of labeled constraints can be obtained without enforcing global consistency; for example, by applying the naive backtracking algorithm described by Dechter, Meiri and Pearl (1991), which is $O(n^3 l^e)$.

## 5. Interval-Based Constraints Through Labeled Point-Based Constraints

The integration of quantitative and qualitative information has been the goal of several temporal models, as was described in Section 1. When intervals are represented by means of their ending points $I_i^+$ $I_i^-$, integration of constraints on intervals and points seems to require some kind of non-binary constraints between time-points (Gerevini & Schubert, 1995; Schwalb & Dechter, 1997; Drakengren & Jonsson, 1997). In this section, the proposed temporal model is applied in order to integrate interval and point-based constraints. Constraints on intervals are managed by means of constraints on ending points of intervals and I-L-Sets. Likewise, metric information can also be added to interval constraints such that an expressive way of integrating qualitative and quantitative constraints is obtained.

### 5.1 Symbolic Interval-Based Constraints

Symbolic constraints on intervals express the qualitative temporal relation between two intervals. Each symbolic constraint is a disjunctive subset of 13 elemental constraints, which are mutually exclusive among them (Allen, 1983). For example, the following constraint

$$I_1 \{ec_1, ec_2\} I_2, \qquad ec_1, ec_2 \in \{b, m, o, d, s, f, e, mi, oi, di, si, fi\},$$

really means '$I_1$ [ $(ec_1 \vee ec_2) \wedge \neg(ec_1 \wedge ec_2)$ ] $I_2$', since $ec_1$ and $ec_2$ are mutually exclusive, and *one and only one* elemental constraint should hold. For reasons of simplicity, we only consider two disjuncts in the symbolic constraint. However, these expressions can be easily extended for managing from 2 to 13 disjuncts. The above expression can be expressed as:

$$I_1 [ (ec_1 \wedge \neg ec_2) \vee (\neg ec_1 \wedge ec_2) ] I_2 \quad \equiv$$
$$I_1 [ (ec_1 \vee \neg ec_1) \wedge (ec_2 \vee \neg ec_2) \wedge \neg(ec_1 \wedge ec_2) \wedge \neg (\neg ec_1 \wedge \neg ec_2) ] I_2 \qquad \textbf{\textit{(e2)}}.$$

In this way, we have:

i)   The constraints $[I_1 \ (ec_1 \lor \neg ec_1) \ I_2]$ and $[I1 \ (ec_2 \lor \neg ec_2) \ I_2]$ can be expressed as disjunctive metric constraints on the same pairs of time-points,

ii)  The constraints $[I_1 \ \neg(ec_1 \land ec_2) \ I_2]$ and $[I_1 \ \neg(\neg ec_1 \land \neg ec_2) \ I_2]$ can be expressed as a mutual exclusion among the associated labels of the above point-based constraints. That is, as a set of I-L-Sets.

We present a simple example to illustrate these conclusions. For instance, $(I_1 \ \{before \ after\} \ I_2)$ can be expressed by means of constraints among the time points $I_1^-$, $I_1^+$, $I_2^-$ and $I_2^+$, as:

$$[I_1 \ \{b \ a\} \ I_2] \equiv (I_1^+ \ \{(0 \ \infty)_{\{Rb1\}}\} \ I_2^-) \lor (I_1^- \ \{(-\infty \ 0)_{\{Ra1\}}\} \ I_2^+).$$

Thus, when intervals are represented by means of their ending points $I_i^+ \ I_i^-$, an interval-based constraint gives rise to disjunctive constraints between different pairs of time points (i.e.: non-binary constraints). These non-binary constraints can be represented as I-L-Sets. Thus, according to the above expression (*e2*),

$$[I_1 \ \{b \ a\} \ I_2] \equiv [I_1 \ (b \lor \neg b) \ I_2] \land [I_1 \ (a \lor \neg a) \ I_2] \land [I_1 \ \neg(b \land a) \ I_2] \land [I_1 \ \neg(\neg b \land \neg a) \ I_2],$$

we have:

$I_1$ before $I_2 \quad \Leftrightarrow \quad I_1^+ \ \{(0 \ \infty)_{\{Rb1\}}\} \ I_2^-, \qquad\quad I_1$ after $I_2 \quad \Leftrightarrow \quad I_1^- \ \{(-\infty \ 0)_{\{Ra1\}}\} \ I_2^+,$

$I_1 \ \neg$before $I_2 \quad \Leftrightarrow \quad I_1^+ \ \{(-\infty \ 0]_{\{Rb2\}}\} \ I_2^-, \qquad I_1 \ \neg$after $I_2 \Leftrightarrow \quad I_1^- \ \{[0 \ \infty)_{\{Ra2\}}\} \ I_2^+.$

Therefore, $[I_1 \ \{b \ a\} \ I_2]$ can be expressed as:

$$[I_1^+ \ \{(0 \ \infty)_{\{Rb1\}} \ (-\infty \ 0]_{\{Rb2\}}\} \ I_2^-] \land [I_1^- \ \{(-\infty \ 0)_{\{Ra1\}} \ [0 \ \infty)_{\{Ra2\}}\} \ I_2^+] \land$$

$$\neg [ \ (I_1^+ \ \{(0 \ \infty)_{\{Rb1\}}\} \ I_2^-) \land (I_1^- \ \{(-\infty \ 0)_{\{Ra1\}}\} \ I_2^+) \ ] \land$$

$$\neg [ \ (I_1^+ \ \{(-\infty \ 0]_{\{Rb2\}}\} \ I_2^-) \land (I_1^- \ \{[0 \ \infty)_{\{Ra2\}}\} \ I_2^+) \ ],$$

which is equivalent to (by using the labels associated to each elemental constraint):

$$[I_1^+ \ \{(0 \ \infty)_{\{Rb1\}} \ (-\infty \ 0]_{\{Rb2\}}\} \ I_2^-] \quad \land \quad [I_1^- \ \{(-\infty \ 0)_{\{Ra1\}} \ [0 \ \infty)_{\{Ra2\}}\} \ I_2^+]$$

and $\{R_{b1} \ R_{a1}\}, \{R_{b2} \ R_{a2}\}$ are I-L-Sets, such that one and only one disjunctive symbolic constraint holds.

Thus, symbolic constraints between intervals can be represented by means of: *(i)* a set of disjunctive metric constraints between time-points, and *(ii)* a set of I-L-Sets. In Table 1, the equivalent metric constraints between interval ending time points for each elemental interval-based constraint are detailed. According to this table, the following steps allow us to represent disjunctive symbolic constraints between intervals by means of disjunctive metric constraints between interval ending points and I-L-Sets:

i)   Each interval $I_i$ is represented by means of its ending points $I_i^+$, $I_i^-$. By default, $(I_i^- \ \{(0, \infty)_{\{R0\}}\} \ I_i^+)$ holds.

ii)  A symbolic constraint between two intervals $(I_i \ c_{ij} \ I_j)$ is composed of a disjunctive set of (from 1 to 13) elemental symbolic constraints $c_{ij} = \{ec_{ij,k}\} \subseteq \{b, m, o, d, s, f, e, bi, mi, oi, di, si, fi\}$.

iii) Each elemental symbolic constraint $ec \in \{b, m, o, d, s, f, e, bi, mi, oi, di, si, fi\}$ is represented

by a conjunctive set of disjunctive point-based metric constraints (fourth column of Table 1). This conjunctive set of point-based constraints expresses the 'fulfillment or non-fulfillment' (ec $\lor \neg$ec) of the elemental symbolic constraint ec.

iv)  A disjunctive set $c_{ij} = \{ec_{ij,k}\}$ of elemental symbolic constraints between $I_i$ and $I_j$ is represented by:

- A conjunctive set of disjunctive point-based metric constraints between the time-points $I_i^+$, $I_i^-$, $I_j^+$ and $I_j^-$. This conjunctive set is composed by the constraints in the fourth column of Table 1 for each elemental constraint in $\{ec_{ij,k}\}$.

- A set of I-L-Sets that expresses the logical relation among elemental symbolic constraints in $\{ec_{ij,k}\}$. That is, *'one and only one elemental symbolic constraint in $\{ec_{ij,k}\}$ should hold'*:

   iv.a)  Only one elemental constraint in $\{ec_{ij,k}\}$ should hold. This condition does not need to be represented since the different sets of point-based constraints that correspond to fulfillment of different elemental symbolic constraints (second column of Table 1) are already mutually exclusive.

   iv.b)  One of the elemental symbolic constraints in $\{ec_{ij,k}\}$ should hold. Let S be the label sets, where each label set corresponds to the point-based constraints which are related to the non-fulfillment of each elemental symbolic constraint in $\{ec_{ij,k}\}$ (third column of Table 1). Thus, the Cartesian product among the label sets in S is a set of I-L-Sets.

For instance, $I_1$ {b m s di} $I_2$ can be represented as:

$$(I_1^- \{ (0 \infty)_{\{R0\}} \} I_1^+), \ (I_2^- \{ (0 \infty)_{\{R0\}} \} I_2^+),$$

$$I_1 \{b \ \neg b\} I_2 \ \Rightarrow \ (I_1^+ \{(0 \infty)_{\{Rb1\}} (-\infty 0]_{\{Rb2\}}\} I_2^-),$$

$$I_1 \{m \ \neg m\} I_2 \ \Rightarrow \ (I_1^+ \{[0 \ 0]_{\{Rm1\}} (0 \infty)_{\{Rm2\}} (-\infty 0)_{\{Rm3\}}\} I_2^-),$$

$$I_1 \{s \ \neg s\} I_2 \ \Rightarrow \ (I_1^- \{[0 \ 0]_{\{Rs1\}} (0 \infty)_{\{Rs3\}} (-\infty 0)_{\{Rs4\}}\} I_2^-) \land (I_1^+ \{(0 \infty)_{\{Rs2\}} (-\infty 0]_{\{Rs5\}}\} I_2^+),$$

$$I_1 \{di \ \neg di\} I_2 \equiv I_2 \{d \ \neg d\} I_1 \Rightarrow \ (I_2^- \{(-\infty 0)_{\{Rd1\}} [0 \infty)_{\{Rd3\}}\} I_1^-) \land (I_2^+ \{(0 \infty)_{\{Rd2\}} (-\infty 0]_{\{Rd4\}}\} I_1^+).$$

Moreover, one of the symbolic constraints in {b, m, s, di} should hold. Thus (according to Point iv.b of the method), the Cartesian product of the associated labels related to the non-fulfillment of each elemental symbolic constraints in {b, m, s, di}. That is:

$$\{\{R_{b2}\}X\{R_{m2}, R_{m3}\}X\{R_{s3}, R_{s4}, R_{s5}\}X\{R_{d3}, R_{d4}\}$$

should be explicitly included in the set of I-L-Sets.

By applying this method, qualitative interval-based constraints can be fully integrated in the proposed labeled point-based constraints. In this case, the interpretation domain for time-points $\{I_i^- I_i^+\}$ can be restricted to only three values ($\{D\} = \{(-\infty, 0), [0 \ 0], (0 \infty)\}$), such that, $l=3$. Therefore, the computational cost of reasoning algorithms is bounded by $O(n^2 \ 3^{2e})$.

To illustrate the proposed method, let's show a typical example on symbolic interval-based constraints (Figure 11.a), which was given by Allen (1983). This example shows how interval-based constraints can be represented and managed by means of disjunctive metric point-based constraints and a minimal IA-TCN can be obtained.

| $I_i$ $ec_{ij,k}$ $I_j$ | $I_i$ $ec_{ij,k}$ $I_j$ | $I_i$ $\neg ec_{ij,k}$ $I_j$ | $I_i$ $(ec_{ij,k} \lor \neg ec_{ij,k})$ $I_j$ |
|---|---|---|---|
| $I_i$ before $I_j$ | $I_i^+$ $\{(0\infty)_{\{Rb1\}}\}$ $I_j^-$ | $I_i^+$ $\{(-\infty 0]_{\{Rb2\}}\}$ $I_j^-$ | $I_i^+$ $\{(0\infty)_{\{Rb1\}}(-\infty 0]_{\{Rb2\}}\}$ $I_j^-$ |
| $I_i$ meets $I_j$ | $I_i^+$ $\{[0\,0]_{\{Rm1\}}\}$ $I_j^-$ | $I_i^+$ $\{(0\infty)_{\{Rm2\}}(-\infty 0)_{\{Rm3\}}\}$ $I_j^-$ | $I_i^+$ $\{[0\,0]_{\{Rm1\}}(0\infty)_{\{Rm2\}}(-\infty 0)_{\{Rm3\}}\}$ $I_j^-$ |
| $I_i$ during $I_j$ | $I_i^-$ $\{(-\infty 0)_{\{Rd1\}}\}$ $I_j^-$<br>$I_i^+$ $\{(0\infty)_{\{Rd2\}}\}$ $I_j^+$ | $(I_i^-$ $\{[0\,0]_{\{Rd3\}}\}$ $I_j^-)$<br>$\lor$ $(I_i^+$ $\{(-\infty 0]_{\{Rd4\}}\}$ $I_j^+)$ | $I_i^-$ $\{(-\infty 0)_{\{Rd1\}}[0\,0]_{\{Rd3\}}\}$ $I_j^-$<br>$I_i^+$ $\{(0\infty)_{\{Rd2\}}(-\infty 0]_{\{Rd4\}}\}$ $I_j^+$ |
| $I_i$ starts $I_j$ | $I_i^-$ $\{[0\,0]_{\{Rs1\}}\}$ $I_j^-$<br>$I_i^+$ $\{(0\infty)_{\{Rs2\}}\}$ $I_j^+$ | $(I_i^-$ $\{(0\infty)_{\{Rs3\}}(-\infty 0)_{\{Rs4\}}\}$ $I_j^-)$<br>$\lor$ $(I_i^+$ $\{(-\infty 0]_{\{Rs5\}}\}$ $I_j^+)$ | $I_i^-$ $\{[0\,0]_{\{Rs1\}}(0\infty)_{\{Rs3\}}(-\infty 0)_{\{Rs4\}}\}$ $I_j^-$<br>$I_i^+$ $\{(0\infty)_{\{Rs2\}}(-\infty 0]_{\{Rs5\}}\}$ $I_j^+$ |
| $I_i$ finishes $I_j$ | $I_i^+$ $\{[0\,0]_{\{Rf1\}}\}$ $I_j^+$<br>$I_i^-$ $\{(-\infty 0)_{\{Rf2\}}\}$ $I_j^-$ | $(I_i^+$ $\{(0\infty)_{\{Rf3\}}(-\infty 0)_{\{Rf4\}}\}$ $I_j^+)$<br>$\lor$ $(I_i^-$ $\{[0\,0]_{\{Rf5\}}\}$ $I_j^-)$ | $I_i^+$ $\{[0\,0]_{\{Rf1\}}(0\infty)_{\{Rf3\}}(-\infty 0)_{\{Rf4\}}\}$ $I_j^+$<br>$I_i^-$ $\{(-\infty 0)_{\{Rf2\}}[0\,0]_{\{Rf5\}}\}$ $I_j^-$ |
| $I_i$ overlaps $I_j$ | $I_i^+$ $\{(-\infty 0)_{\{Ro1\}}\}$ $I_j^-$<br>$I_i^+$ $\{(0\infty)_{\{Ro2\}}\}$ $I_j^+$<br>$I_i^-$ $\{(0\infty)_{\{Ro3\}}\}$ $I_j^-$ | $(I_i^+$ $\{[0\,0]_{\{Ro4\}}\}$ $I_j^-)$<br>$\lor$ $(I_i^+$ $\{(-\infty 0]_{\{Ro5\}}\}$ $I_j^+)$<br>$\lor$ $(I_i^-$ $\{(-\infty 0]_{\{Ro6\}}\}$ $I_j^-)$ | $I_i^+$ $\{(-\infty 0)_{\{Ro1\}}[0\,0]_{\{Ro4\}}\}$ $I_j^-$<br>$I_i^+$ $\{(0\infty)_{\{Ro2\}}(-\infty 0]_{\{Ro5\}}\}$ $I_j^+$<br>$I_i^-$ $\{(0\infty)_{\{Ro3\}}(-\infty 0]_{\{Ro6\}}\}$ $I_j^-$ |
| $I_i$ equal $I_j$ | $I_i^+$ $\{[0\,0]_{\{Re1\}}\}$ $I_j^+$<br>$I_i^-$ $\{[0\,0]_{\{Re2\}}\}$ $I_j^-$ | $(I_i^+$ $\{(0\infty)_{\{Re3\}}(-\infty 0)_{\{Re4\}}\}$ $I_j^+)$<br>$\lor$ $(I_i^-$ $\{(0\infty)_{\{Re5\}}(-\infty 0)_{\{Re6\}}\}$ $I_j^-)$ | $I_i^+$ $\{(0\infty)_{\{Re3\}}[0\,0]_{\{Re1\}}(-\infty 0)_{\{Re4\}}\}$ $I_j^+$<br>$I_i^-$ $\{(0\infty)_{\{Re5\}}[0\,0]_{\{Re2\}}(-\infty 0)_{\{Re6\}}\}$ $I_j^-$ |

Table 1: Interval-based constraints and their equivalent disjunctive metric constraints between interval ending points *(Cells in the second and fourth columns are a conjunctive set of constraints)*

| Symbolic Constraint | Disjunctive Metric Constraint between $I^+ I^-$ | Inconsistent-Label-Sets |
|---|---|---|
| (IA {d di} IB) $\Rightarrow$ | $IA^-$ $\{(-\infty 0)_{\{R1\}}[0\infty)_{\{R3\}}\}$ $IB^-$<br>$IA^+$ $\{(0\infty)_{\{R2\}}(-\infty 0]_{\{R4\}}\}$ $IB^+$<br>$IB^-$ $\{(-\infty 0)_{\{R5\}}[0\infty)_{\{R7\}}\}$ $IA^-$<br>$IB^+$ $\{(0\infty)_{\{R6\}}(-\infty 0]_{\{R8\}}\}$ $IA^+$ | $\{R_4 R_8\}$ $\{R_3 R_8\}$<br>$\{R_4 R_7\}$ $\{R_3 R_7\}$ |
| (IB {d di} IC) $\Rightarrow$ | $IB^-$ $\{(-\infty 0)_{\{R9\}}[0\infty)_{\{R11\}}\}$ $IC^-$<br>$IB^+$ $\{(0\infty)_{\{R10\}}(-\infty 0]_{\{R12\}}\}$ $IC^+$<br>$IC^-$ $\{(-\infty 0)_{\{R13\}}[0\infty)_{\{R15\}}\}$ $IB^-$<br>$IC^+$ $\{(0\infty)_{\{R14\}}(-\infty 0]_{\{R16\}}\}$ $IB^+$ | $\{R_{12} R_{16}\}$ $\{R_{11} R_{16}\}$<br>$\{R_{12} R_{15}\}$ $\{R_{11} R_{15}\}$ |
| (ID {m s} IA) $\Rightarrow$ | $ID^+$ $\{[0\,0]_{\{R17\}}(0\infty)_{\{R18\}}(-\infty 0)_{\{R19\}}\}$ $IA^-$<br>$ID^-$ $\{[0\,0]_{\{R20\}}(0\infty)_{\{R22\}}(-\infty 0)_{\{R23\}}\}$ $IA^-$<br>$ID^+$ $\{(0\infty)_{\{R21\}}(-\infty 0]_{\{R24\}}\}$ $IA^+$ | $\{R_{19} R_{24}\}$ $\{R_{18} R_{24}\}$ $\{R_{19} R_{23}\}$<br>$\{R_{18} R_{23}\}$ $\{R_{19} R_{22}\}$ $\{R_{18} R_{22}\}$ |
| (ID {o} IB) $\Rightarrow$ | $ID^+$ $\{(-\infty 0)_{\{R0\}}\}$ $IB^-$<br>$ID^+$ $\{(0\infty)_{\{R0\}}\}$ $IB^+$<br>$ID^-$ $\{(0\infty)_{\{R0\}}\}$ $IB^-$ | |
| (ID {m s} IC) $\Rightarrow$ | $ID^+$ $\{[0\,0]_{\{R25\}}(0\infty)_{\{R26\}}(-\infty 0)_{\{R27\}}\}$ $IC^-$<br>$ID^-$ $\{[0\,0]_{\{R28\}}(0\infty)_{\{R30\}}(-\infty 0)_{\{R31\}}\}$ $IC^-$<br>$ID^+$ $\{(0\infty)_{\{R29\}}(-\infty 0]_{\{R32\}}\}$ $IC^+$ | $\{R_{27} R_{32}\}$ $\{R_{26} R_{32}\}$ $\{R_{27} R_{31}\}$<br>$\{R_{26} R_{31}\}$ $\{R_{27} R_{30}\}$ $\{R_{26} R_{30}\}$ |

Table 2: Symbolic constraints in Figure 11.a by means of disjunctive metric constraints between $I^+$, $I^-$

Figure 11.a represents a path-consistent IA-TCN, which has inconsistent values in constraints (Allen, 1983). In Table 2, we have the interval-based symbolic constraints for this example, the corresponding disjunctive metric constraints between their ending time-points ($I_i^+$, $I_i^-$) and the corresponding set of I-L-Sets (according to Table 1). Moreover, we also have:

$(IA^- \{(0 \infty)_{\{R0\}}\} IA^+)$, $(IB^- \{(0 \infty)_{\{R0\}}\} IB^+)$, $(IC^- \{(0 \infty)_{\{R0\}}\} IC^+)$ and $(ID^- \{(0 \infty)_{\{R0\}}\} ID^+)$.

When all these metric constraints among the ending time-points of intervals are updated according the proposed methods in Section 4, the labeled minimal TCN in Table 3 is obtained. The associated labels to each elemental constraint (disjunct) in constraints are not included for reasons of brevity.

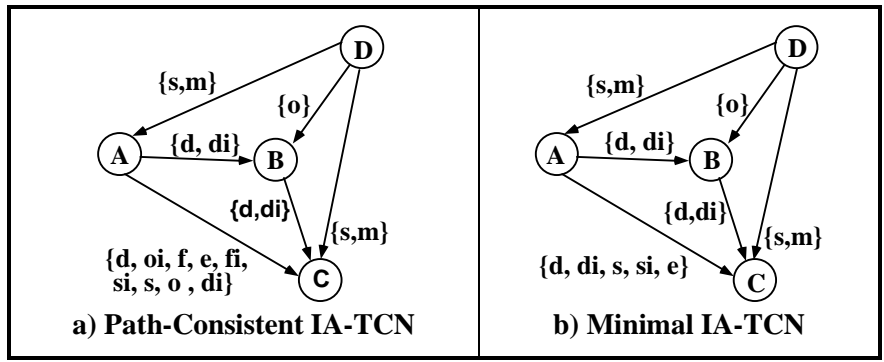

a) Path-Consistent IA-TCN     b) Minimal IA-TCN

Figure 11: Path-Consistent and equivalent Minimal IA-TCN

|  | IA⁺ | IA⁻ | IB⁺ | IB⁻ | IC⁺ | IC⁻ | ID⁺ | ID⁻ |
|---|---|---|---|---|---|---|---|---|
| **IA⁺** |  | {(-∞ 0)} | {(0 ∞), (-∞ 0)} | {(-∞ 0)} | {(-∞ ∞)} | {(-∞ 0)} | {(-∞ 0)} | {(-∞ 0)} |
| **IA⁻** | {(0 ∞)} |  | {(0 ∞)} | {(-∞ 0), (0 ∞)} | {(0 ∞)} | {(-∞ 0), [0 0], (0 ∞)} | {[0 0], (0 ∞)} | {(-∞ 0), [0 0]} |
| **IB⁺** | {(-∞ 0), (0 ∞)} | {(-∞ 0)} |  | {(-∞ 0)} | {(-∞ 0), (0 ∞)} | {(-∞ 0)} | {(-∞ 0)} | {(-∞ 0)} |
| **IB⁻** | {(0 ∞)} | {(0 ∞), (-∞ 0)} | {(0 ∞)} |  | {(0 ∞)} | {(-∞ 0), (0 ∞)} | {(0 ∞)} | {(-∞ 0)} |
| **IC⁺** | {(-∞ ∞)} | {(-∞ 0)} | {(-∞ 0), (0 ∞)} | {(-∞ 0)} |  | {(-∞ 0)} | {(-∞ 0)} | {(-∞ 0)} |
| **IC⁻** | {(0 ∞)} | {(-∞ 0), [0 0], (0 ∞)} | {(0 ∞)} | {(-∞ 0), (0 ∞)} | {(0 ∞)} |  | {(0 ∞), [0 0]} | {(-∞ 0), [0 0]} |
| **ID⁺** | {(0 ∞)} | {(-∞ 0), [0 0]} | {(0 ∞)} | {(-∞ 0)} | {(0 ∞)} | {(-∞ 0), [0 0]} |  | {(-∞ 0)} |
| **ID⁻** | {(0 ∞)} | {[0 0], (0 ∞)} | {(0 ∞)} | {(0 ∞)} | {(0 ∞)} | {[0 0], (0 ∞)} | {(0 ∞)} |  |

Table 3: The minimal metric point-based TCN of the IA-TCN in Figure 11.a Allen (1983) remarks that the symbolic constraint (IA {f fi} IC) cannot hold given the existing constraints between IA, IB, IC and ID. In the labeled point-based TCN, (IA {f fi} IC) is represented by a set of constraints among ending points of IA and IC. Moreover, the labels associated to each labeled elemental constraint allow us to determine whether a set of elemental constraints between different pairs of time-points can be part of a global solution (*Theorem 10*). Thus, we can deduce whether (IA {f fi} IC) can hold in the point-based TCN.

The existing constraints between the ending time-points of IC and IA, with their associated label-sets are:

$IC^+$  {$(-\infty \infty)_{\{R25\ R30\ R29\ R17\ R22\ R21\ R0\}} \lor \{R27\ R28\ R29\ R19\ R20\ R21\ R0\}$},

$(-\infty\ 0)_{\{R27\ R28\ R29\ R17\ R22\ R21\ R9\ R10\ R15\ R16\ R1\ R2\ R7\ R0\ R8\}}$,

$(0\ \infty)_{\{R25\ R30\ R29\ R19\ R20\ R21\ R11\ R12\ R13\ R14\ R3\ R4\ R5\ R0\ R6\}}$}  $IA^+$

$IC^-$  {$(0\ \infty)_{\{R27\ R28\ R29\ R17\ R22\ R21\ R9\ R10\ R15\ R16\ R1\ R2\ R7\ R8\ R0\}}$,

$[0\ 0]_{\{R25\ R30\ R29\ R17\ R22\ R21\ R0\}} \lor \{R27\ R28\ R29\ R19\ R20\ R21\ R0\}$},

$(-\infty\ 0)_{\{R25\ R30\ R29\ R19\ R20\ R21\ R11\ R12\ R13\ R14\ R3\ R4\ R5\ R6\ R0\}}$}  $IA^-$

Let's ask for each disjunct in (IA {f fi} IC):

i) The constraint (IA {f} IC) implies ($IC^+$ {[0 0]} $IA^+$) $\land$ ($IC^-$ {$(-\infty\ 0)$} $IA^-$). According to Theorem 10, these constraints hold iff the set-union of the label sets associated to ($IC^+$ [0 0] $IA^+$) and to ($IC^-$ $(-\infty\ 0)$ $IA^-$) is not an I-L-Set. We have two possibilities:

*i.1)*  {$R_{25}\ R_{30}\ R_{29}\ R_{17}\ R_{22}\ R_{21}\ R_0$} $\cup$ {$R_{25}\ R_{30}\ R_{29}\ R_{19}\ R_{20}\ R_{21}\ R_{11}\ R_{12}\ R_{13}\ R_{14}\ R_3\ R_4\ R_5\ R_6\ R_0$} = {$R_6\ R_5\ R_4\ R_3\ R_{20}\ R_{19}\ R_{25}\ R_{30}\ R_{29}\ R_{17}\ R_{22}\ R_{21}\ R_{11}\ R_{12}\ R_{13}\ R_{14}\ R_0$ }, or

*i.2)*  {$R_{27}\ R_{28}\ R_{29}\ R_{19}\ R_{20}\ R_{21}\ R_0$} $\cup$ {$R_{25}\ R_{30}\ R_{29}\ R_{19}\ R_{20}\ R_{21}\ R_{11}\ R_{12}\ R_{13}\ R_{14}\ R_3\ R_4\ R_5\ R_6\ R_0$} = {$R_{14}\ R_{13}\ R_{12}\ R_{11}\ R_{30}\ R_{25}\ R_{27}\ R_{28}\ R_{29}\ R_{19}\ R_{20}\ R_{21}\ R_3\ R_4\ R_5\ R_0\ R_6$}.

However, both label sets (*i.1, i.2*) are I-L-Sets: For instance, {$R_{19}\ R_{22}$} and {$R_{27}\ R_{30}$} are I-L-Sets (Table 2) and they are subsets of *i.1* and *i.2*, respectively. Thus, (IA {f} IC) does not hold.

ii) The constraint (IA {fi} IC) implies ($IC^+$ {[0 0]} $IA^+$) $\land$ ($IC^-$ {$(0\ \infty)$} $IA^-$). Similarly:

*ii.1)*  {$R_{25}\ R_{30}\ R_{29}\ R_{17}\ R_{22}\ R_{21}\ R_0$} $\cup$ {$R_{27}\ R_{28}\ R_{29}\ R_{17}\ R_{22}\ R_{21}\ R_9\ R_{10}\ R_{15}\ R_{16}\ R_1\ R_2\ R_7\ R_8\ R_0$} = {$R_{16}\ R_{15}\ R_{10}\ R_9\ R_{28}\ R_{27}\ R_{25}\ R_{30}\ R_{29}\ R_{17}\ R_{22}\ R_{21}\ R_1\ R_2\ R_7\ R_0\ R_8$}.

This label set is an I-L-Set. For instance, {$R_{30}\ R_{27}$} is an I-L-Set. Also,

*ii.2)*  {$R_{27}\ R_{28}\ R_{29}\ R_{19}\ R_{20}\ R_{21}\ R_0$} $\cup$ {$R_{27}\ R_{28}\ R_{29}\ R_{17}\ R_{22}\ R_{21}\ R_9\ R_{10}\ R_{15}\ R_{16}\ R_1\ R_2\ R_7\ R_8\ R_0$} = {$R_8\ R_7\ R_2\ R_1\ R_{22}\ R_{17}\ R_{27}\ R_{28}\ R_{29}\ R_{19}\ R_{20}\ R_{21}\ R_9\ R_{10}\ R_{15}\ R_{16}\ R_0$}.

Both these label sets (*ii.1, ii.2*) are also I-L-Sets. For instance, {$R_{30}\ R_{27}$} and {$R_{19}\ R_{22}$} are I-L-Sets. Thus, (IA {fi} IC) does not hold either.

In conclusion, the symbolic constraint (IA {f fi} IC) cannot hold on the globally labeled-consistent point-based TCN. This conclusion could be also obtained from a minimal IA-TCN (Figure 11.b). Additionally, we have that (IA {f fi} IC) implies ($IA^+$ [0 0] $IC^+$). That is, if the constraint (IA$^+$ [0 0] IC$^+$) holds, we have that the associated constraints to the label sets {$R_{25}\ R_{30}\ R_{29}\ R_{17}\ R_{22}\ R_{21}\ R_0$} or {$R_{27}\ R_{28}\ R_{29}\ R_{19}\ R_{20}\ R_{21}\ R_0$} should also hold. Each one of these label sets implies (IC$^-$ {[0 0]} IA$^-$). That is: (IA$^+$ [0 0] IC$^+$) $\rightarrow$ (IC$^-$ {[0 0]} IA$^-$). Thus, the only way that (IA$^+$ [0 0] IC$^+$) can hold is if (IA {e} IC) holds. These relations will be detailed in Section 6.

## 5.2 Metric Constraints on Intervals

Metric constraints between intervals can also be managed in the described temporal model. From a general point of view, metric information can be added to each elemental interval-based constraint in a standard way (Table 4). These metric constraints on interval boundaries (Table 4) are similar to the ones proposed by Staab and Hahn (1998).

| IA Symbolic Elemental Constraints | IA Metric Elemental Constraints $c_{ij} \equiv \{[dm_1\ dM_1], [dm_2\ dM_2], ..... [dm_n\ dM_n]\}$ $c'_{ij} \equiv \{[dm'_1\ dM'_1], [dm'_2\ dM'_2], ..... [dm'_n\ dM'_n]\}$ | |
|---|---|---|
| $I_i$ before $I_j$ | $I_i$ (before $c_{ij}$) $I_j$ | 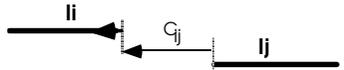 |
| $I_i$ meets $I_j$ | $I_i$ (meets $c_{ij}$) $I_j$ | 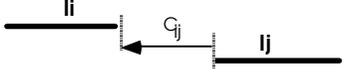 |
| $I_i$ during $I_j$ | $I_i$ ($c_{ij}$ during $c'_{ij}$) $I_j$ | 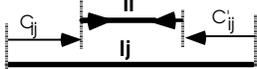 |
| $I_i$ starts $I_j$ | $I_i$ (starts $c_{ij}$) $I_j$ | 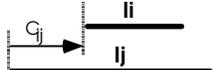 |
| $I_i$ finishes $I_j$ | $I_i$ (finishes $c_{ij}$) $I_j$ | 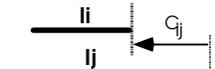 |
| $I_i$ overlaps $I_j$ | $I_i$ (overlaps $c_{ij}$) $I_j$ | 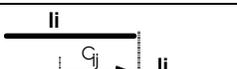 |
| $I_i$ equal $I_j$ | $I_i$ ($c_{ij}$ equal $c'_{ij}$) $I_j$ | 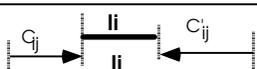 |

Table 4: Metric interval constraints on interval boundaries

Obviously, the metric constraints of Table 4 can be managed in the proposed model, by means of metric constraints on interval ending points. Thus, symbolic constraints of Interval Algebra can be extended in this way to metric domain. However, since each interval is represented by means of its ending time-points, more flexible metric constraints on intervals can be represented by means of metric constraints on their ending time-points. In this way, the described model also subsumes the Interval Distance Sub Algebra model proposed by Badaloni and Berati (1996). Moreover, ending points of intervals can also be related to the initial time-point $T_0$, and unary metric constraints on interval durations can be expressed by means of metric constraints between the two ending points of each interval:

$$dur\ (I_i) = \{[dm_1\ dM_1], [dm_2\ dM_2], ..... [dm_n\ dM_n]\} \Rightarrow$$

$$(I_i^- \{[dm_1\ dM_1], [dm_2\ dM_2], ..... [dm_n\ dM_n]\}\ I_i^+).$$

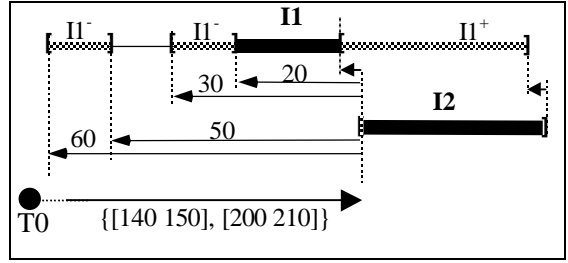

Figure 12: Metric constraints between intervals

Thus, following constraints (Figure 12):

($I1$ {b, o} $I2$) $\land$ ($I1^-$ is [[20 30], [50 60]] before $I2^-$) $\land$ ($I2^-$ is {[140 150], [200 210]} after $T_0$)

can be represented as (Table 1):

By default: ($I1^-$ { $(0\ \infty)_{\{R0\}}$ } $I1^+$), ($I2^-$ { $(0\ \infty)_{\{R0\}}$ } $I2^+$), and

($I1$ {b, o} $I2$) $\Rightarrow$ ($I1^+$ { $(0\ \infty)_{\{Rb1\}}$ $(-\infty\ 0]_{\{Rb2\}}$ } $I2^-$), ($I1^+$ { $(-\infty\ 0)_{\{Ro1\}}$ $(0\ \infty)_{\{Ro4\}}$ } $I2^-$),

($I1^+$ { $(0\ \infty)_{\{Ro2\}}$ $(-\infty\ 0]_{\{Ro5\}}$ } $I2^+$), ($I1^-$ { $(0\ \infty)_{\{Ro3\}}$ $(-\infty\ 0]_{\{Ro6\}}$ } $I2^-$),

($I1^-$ is [[20 30], [50 60]] at the left of $I2^-$) $\Rightarrow$ ($I1^-$ {[50 60]$_{\{R1\}}$ [20 30]$_{\{R2\}}$} $I2^-$),

($I2^-$ is {[140 150], [200 210]} after $T_0$) $\Rightarrow$ ($T_0$ {[140 150]$_{\{R3\}}$ [200 210]$_{\{R4\}}$} $I2^-$),

and $\{R_{b2}\ R_{o4}\}$, $\{R_{b2}\ R_{o5}\}$, $\{R_{b2}\ R_{o6}\}$, $\{R_1\ R_2\}$ and $\{R_3\ R_4\}$ are I-L-Sets.

## 6. Reasoning on Logical Expressions of Constraints

In the described model, each disjunct in an input constraint is univocally associated to a label. Moreover, the label set associated to each derived elemental constraint represents the support-set of input elemental constraints from which the elemental constraint is derived. I-L-Sets represent inconsistent sets of input elemental constraints. By reasoning on labeled disjunctive constraints, associated label lists and I-L-Sets, the temporal model offers the capability of reasoning on logical expressions of elemental constraints belonging to disjunctive constraints between different pairs of time points. Let's assume the following labeled input constraints:

$(n_i\ lc_{ij}\ n_j) \equiv (n_i\ \{(lec_{ij.1})_{\{Rij.1\}}\ (lec_{ij.2})_{\{Rij.2\}}\ .....(lec_{ij.p})_{\{Rij.p\}}\}\ n_j)$,

$(n_k\ lc_{kl}\ n_l) \equiv (n_k\ \{(lec_{kl.1})_{\{Rkl.1\}}\ (lec_{kl.2})_{\{Rkl.2\}}\ .....(lec_{kl.q})_{\{Rkl.q\}}\}\ n_l)$

i) To represent that two elemental constraints[4] ($elc_{ij.x} \in lc_{ij}$, $elc_{kl.y} \in lc_{kl}$) cannot hold simultaneously (that is $\neg(elc_{ij.x} \land elc_{kl.y})$) the label set $\{R_{ij.x}\ R_{kl.y}\}$ should be added to the set of I-L-Sets.

ii) To represent a logical dependency between two elemental constraints, such as *'If $lec_{ij.x}$ then $lec_{kl.y}$'* (where $lec_{ij.x} \in c_{ij}$, $lec_{kl.y} \in c_{kl}$), the Cartesian product $\{R_{ij.x}\}$ X $\{\{R_{kl.1}, R_{kl.2}, ....., R_{kl.q}\}$-$\{R_{kl.y}\}\}$ should be added to the set of I-L-Sets.

iii) To represent that two elemental constraints ($elc_{ij.x} \in lc_{ij}$, $elc_{kl.y} \in lc_{kl}$) should hold simultaneously (bi-directional logical dependency), the Cartesian products $\{R_{ij.x}\}$ X $\{\{R_{kl.1}, R_{kl.2}, ....., R_{kl.q}\}$- $\{R_{kl.y}\}\}$ and $\{R_{kl.y}\}$ X $\{\{R_{ij.1}, R_{ij.2}, \ldots, R_{ij.p}\}$-$\{R_{ij.x}\}\}$ should be added to the set of I-L-Sets.

---

[4] For reasons of simplicity, only two elemental constraints are shown. However, more than two disjunctions can be managed in a similar way. Likewise, these features can be also applied to labeled derived constraints.

For instance, let's see the Example 2 of Section 4.4 (Figure 10):

- To represent that *'John goes to work by car and Fred goes to work walking'* is not possible, $\{R_1 \; R_5\}$ should be asserted as an I-L-Set.

- To represent that *'if John goes to work by car then Fred goes to work walking'*, $\{R_1 \; R_3\}$ and $\{R_1 \; R_4\}$ should be asserted as I-L-Sets.

- To represent that *'if John goes to work by car then Fred goes to work walking, and vice versa'*, $\{R_1 \; R_3\}$, $\{R_1 \; R_4\}$ and $\{R_5 \; R_2\}$ should be asserted as I-L-Sets.

In a similar way, logical relations among point-based and interval-based elemental constraints can also be represented. For instance, the logical dependence *"the duration of $I_1$ is [5 8] if $I_2$ is before $I_3$ and the duration of $I_1$ is [12 15] if $I_2$ is after $I_3$"* can be represented as:

$(I_2 \; \{b, \; bi\} \; I_3) \implies$  $(I_2^+ \; \{(0 \; \infty)_{\{Rb9\}} \; (-\infty \; 0]_{\{Rb10\}}\} \; I_3^-)$,  $(I_3^+ \; \{(0 \; \infty)_{\{Rb11\}} \; (-\infty \; 0]_{\{Rb12\}}\} \; I_2^-)$,

$\{R_{b10} \; R_{b12}\}$ is an I-L-Set,

$(I_1^- \; \{[5 \; 8]_{\{R1\}} \; [12 \; 15]_{\{R2\}}\} \; I_1^+)$,

and $\{R_1 \; R_{b11}\}$, $\{R_2 \; R_{b9}\}$ are I-L-Sets, since $R_{b11}$ is associated to *'$I_2$ is after $I_3$'* and $R_{b9}$ is associated to *'$I_2$ is before $I_3$'*. Likewise, *"$I_1$ starts at the same time as $I_2$ if $t_1$ occurs after $t_2$"* can be represented as *(see Table 1)*:

$I_1 \; \{s, \; \neg s\} \; I_2 \implies (I_1^- \; \{[0 \; 0]_{\{Rs1\}} \; (0 \; \infty)_{\{Rs3\}} \; (-\infty \; 0)_{\{Rs4\}}\} \; I_2^-)$, $(I_1^+ \; \{(0 \; \infty)_{\{Rs2\}} \; (-\infty \; 0]_{\{Rs5\}}\} \; I_2^+)$,

$(t_1 \; \{(-\infty \; -1]_{\{R1\}}, \; [0 \; 0]_{\{R2\}}, \; [1 \; \infty)_{\{R3\}}\} \; t_2)$,

and $\{R_3 \; R_{s3}\}$, $\{R_3 \; R_{s4}\}$, and $\{R_3 \; R_{s5}\}$ are I-L-Sets, since $R_3$ is associated to *'$t_1$ occurs after $t_2$'* and $R_{s3}$, $R_{s4}$ and $R_{s5}$ are associated to *'$I_1$ does not start at the same time as $I_2$'*.

## 6.1 Disjunctions of Point and Interval-Based Constraints

Disjunctions of constraints between different pairs of points and intervals can be represented in the proposed model by means of labeled constraints between points and a set of I-L-Sets. This subsumes the related expressiveness in the subset of disjunctive linear constraints proposed by Stergiou and Koubarakis (1998), where only disjunctions of constraints between different pairs of points are managed.

To represent a *disjunctive set of disjunctive constraints between points*, we have[5]:

$(n_i \; lc_{ij} \; n_j) \vee (n_k \; lc_{kl} \; n_l)$   *can be represented as:*   $(n_i \; \{lc_{ij} \vee \neg lc_{ij}\} \; n_j) \wedge (n_k \; \{lc_{kl} \vee \neg lc_{kl}\} \; n_l)$,

and some logical relation among $lc_{ij}$, $\neg lc_{ij}$, $lc_{kl}$ and $\neg lc_{kl}$. Thus, the *disjunctive set* of constraints:

$\{(n_i \; lc_{ij} \; n_j) \vee (n_k \; lc_{kl} \; n_l)\} \equiv$

$\{(n_i \; \{(lec_{ij.1})_{\{Rij.1\}}, (lec_{ij.2})_{\{Rij.2\}}, \ldots, (lec_{ij.p})_{\{Rij.p\}}\} \; n_j) \vee$

$(n_k \; \{(lec_{kl.1})_{\{Rkl.1\}}, (lec_{kl.2})_{\{Rkl.2\}}, \ldots, (lec_{kj.q})_{\{Rkl.q\}}\} \; n_l)\}$

---

[5] For reasons of simplicity, only two constraints are shown. However, more than two disjunctive constraints can be managed in a similar way.

can be represented as:

i) A *conjunctive set of constraints* between $(n_i, n_j)$ and between $(n_k, n_l)$, where, $\neg(lec_x)$ can be represented by means of the complementary domain of $(lec_x)$:

$(n_i \{(lec_{ij.1})_{\{Rij.1\}}, (lec_{ij.2})_{\{Rij.2\}}, ...., (lec_{ij.p})_{\{Rij.p\}}, \neg\{(lec_{ij.1})_{\{Rij.1\}}, (lec_{ij.2})_{\{Rij.2\}}, ...., (lec_{ij.p})_{\{Rij.p\}}\}\} n_j) \wedge$
$(n_k \{(lec_{kl.1})_{\{Rkl.1\}}, (lec_{kl.2})_{\{Rkl.2\}}, ..., (lec_{kj.q})_{\{Rkl.q\}}, \neg\{(lec_{kl.1})_{\{Rkl.1\}}, (lec_{kl.2})_{\{Rkl.2\}}, ..., (lec_{kl.q})_{\{Rkl.q\}}\}\} n_l)$

$\equiv \{(n_i \{(lec_{ij.1})_{\{Rij.1\}}, (lec_{ij.2})_{\{Rij.2\}}, ..., (lec_{ij.p})_{\{Rij.p\}}, (\neg lec_{ij.1})_{\{R'ij.1\}}, (\neg lec_{ij.2})_{\{R'ij.2\}}, ..., (\neg lec_{ij.p})_{\{R'ij.p\}}\} n_j) \wedge$
$(n_k \{(lec_{kl.1})_{\{Rkl.1\}}, (lec_{kl.2})_{\{Rkl.2\}}, .., (lec_{kj.q})_{\{Rkl.q\}}, (\neg lec_{kl.1})_{\{R'kl.1\}}, (\neg lec_{kl.2})_{\{R'kl.2\}}, ..., (\neg lec_{kl.q})_{\{R'kl.q\}}\} n_l)\}$

ii) A set of I-L-Sets to represent the *mutually exclusive* disjunction of $lc_{ij}$ and $lc_{kl}$ (they cannot simultaneously hold):

    ii.a) One of the constraints $lc_{ij}$ or $lc_{kl}$ should hold: The Cartesian product of label sets from *complementary* domains of $lc_{ij}$ and $lc_{kl}$, $\{R'_{ij.1}, R'_{ij.2}, ...., R'_{ij.p}\}X\{R'_{kl.1}, R'_{kl.2}, ...., R'_{kl.q}\}$, are I-L-Sets.

    ii.b) Only one of the constraints $lc_{ij}$ or $lc_{kl}$ should hold: The Cartesian product of label sets from $lc_{ij}$ and $lc_{kl}$, $\{R_{ij.1}, R_{ij.2}, ...., R_{ij.p}\}X\{R_{kl.1}, R_{kl.2}, ...., R_{kl.q}\}$ are I-L-Sets.

Thus, *disjunctive* and *conjunctive* sets of disjunctive constraints between points can be represented and managed by means of a *conjunctive* set of disjunctive constraints and a set of I-L-Sets. For example:

$(t_i \{[5\ 5]_{\{R1\}} [10\ 10]_{\{R2\}}\} t_j) \vee (t_k \{[0\ 0]_{\{R3\}} [20\ 20]_{\{R4\}}\} t_l) \equiv$

$(t_i \{[5\ 5]_{\{R1\}} [10\ 10]_{\{R2\}} (-\infty\ 5)_{\{R5\}} (5\ 10)_{\{R6\}} (10\ \infty)_{\{R7\}}\} t_j) \wedge$
                     $(t_k \{[0\ 0]_{\{R3\}} [20\ 20]_{\{R4\}} (-\infty\ 0)_{\{R8\}} (0\ 20)_{\{R9\}} (20\ \infty)_{\{R10\}}\} t_l),$

and

    (ii.a) since $(t_i \{[5\ 5]_{\{R1\}}, [10\ 10]_{\{R2\}}\} t_j]$ or $[t_k \{[0\ 0]_{\{R3\}}, [20\ 20]_{\{R4\}}\} t_l]$ should hold:

                         $\{R_5\ R_6\ R_7\}X\{R_8\ R_9\ R_{10}\}$ are I-L-Sets,

    (ii.b) since only one constraint $(t_i \{[5\ 5]_{\{R1\}} [10\ 10]_{\{R2\}}\} t_j)$ or $(t_k \{[0\ 0]_{\{R3\}} [20\ 20]_{\{R4\}}\} t_l)$ should hold:

           $\{R_1\ R_2\}X\{R_3\ R_4\} = \{R_1\ R_3\}, \{R_1\ R_4\}, \{R_2\ R_3\}, \{R_2\ R_4\}$ are I-L-Sets.

| $I_i\ ec_{ij}\ I_j$ | $I_i\ ec_{ij}\ I_j$ | $I_i\ \neg ec_{ij}\ I_j$ | $I_i\ (ec_{ij} \vee \neg ec_{ij})\ I_j$ |
|---|---|---|---|
| $I_1$ before $I_2$ | $I_1^+ \{(0\ \infty)_{\{Rb1\}}\} I_2^-$ | $I_1^+ \{(-\infty\ 0)_{\{Rb2\}}\} I_2^-$ | $I_1^+ \{(0\ \infty)_{\{Rb1\}} (-\infty\ 0)_{\{Rb2\}}\} I_2^-$ |
| $I_3$ before $I_4$ | $I_3^+ \{(0\ \infty)_{\{Rb3\}}\} I_4^-$ | $I_3^+ \{(-\infty\ 0)_{\{Rb4\}}\} I_4^-$ | $I_3^+ \{(0\ \infty)_{\{Rb3\}} (-\infty\ 0)_{\{Rb4\}}\} I_4^-$ |

Table 5: Point-based constraints for $(I_1$ before $I_2)$ and $(I_3$ before $I_4)$

Similarly, *disjunctions of interval-based constraints* between different pairs of intervals can also be represented. For instance, from Table 1 and Table 5, $\{(I_1$ before $I_2) \vee (I_3$ before $I_4)\}$ can be represented as:

    $(I_1^+ \{(0\ \infty)_{\{Rb1\}} (-\infty\ 0]_{\{Rb2\}}\} I_2^-), (I_3^+ \{(0\ \infty)_{\{Rb3\}} (-\infty\ 0]_{\{Rb4\}}\} I_4^-),$

and

a) one of the constraints ($I_1$ before $I_2$) or ($I_3$ before $I_4$) should hold. Thus, the Cartesian product of label sets associated to the disjunctive constraints in ($I_i \neg ec_{ij} I_j$) is a set of I-L-Sets: {$R_{b2}$, $R_{b4}$} is an I-L-Set,

b) only one of the constraints ($I_1$ before $I_2$) or ($I_3$ before $I_4$) should hold. Thus, the label set associated to the mutual fulfillment of constraints in ($I_i ec_{ij} I_j$) is an I-L-Set: {$R_{b1}$, $R_{b3}$} is an I-L-Set.

Thus:

$\{(I_1 \text{ before } I_2) \vee (I_3 \text{ before } I_4)\} \equiv$

$(I_1^+ \{(0 \infty)_{\{Rb1\}} (-\infty 0]_{\{Rb2\}}\} I_2^-), (I_3^+ \{(0 \infty)_{\{Rb3\}} (-\infty 0]_{\{Rb4\}}\} I_4^-),$

and {$R_{b2}$, $R_{b4}$}, {$R_{b1}$, $R_{b3}$} are I-L-Sets.

| $I_i ec_{ij} I_j$ | $I_i ec_{ij} I_j$ | $I_i \neg ec_{ij} I_j$ | $I_i (ec_{ij} \vee \neg ec_{ij}) I_j$ |
|---|---|---|---|
| ($I_1$ during $I_2$) | $I_1^- \{(-\infty 0)_{\{Rd1\}}\} I_2^-$ | ($I_1^- \{[0 \infty)_{\{Rd3\}}\} I_2^-$) | $I_1^- \{(-\infty 0)_{\{Rd1\}} [0 \infty)_{\{Rd3\}}\} I_2^-$ |
| | $I_1^+ \{(0 \infty)_{\{Rd2\}}\} I_2^+$ | $\vee$ ($I_1^+ \{(-\infty 0]_{\{Rd4\}}\} I_2^+$) | $I_1^+ \{(0 \infty)_{\{Rd2\}} (-\infty 0]_{\{Rd4\}}\} I_2^+$ |
| ($I_3$ starts $I_4$) | $I_3^- \{[0 0]_{\{Rs1\}}\} I_4^-$ | ($I_3^- \{(0 \infty)_{\{Rs3\}} (-\infty 0)_{\{Rs4\}}\} I_4^-$) | $I_3^- \{[0 0]_{\{Rs1\}} (0 \infty)_{\{Rs3\}} (-\infty 0)_{\{Rs4\}}\} I_4^-$ |
| | $I_3^+ \{(0 \infty)_{\{Rs2\}}\} I_4^+$ | $\vee$ ($I_3^+ \{(-\infty 0]_{\{Rs5\}}\} I_4^+$) | $I_3^+ \{(0 \infty)_{\{Rs2\}} (-\infty 0]_{\{Rs5\}}\} I_4^+$ |

Table 6: Point-based constraints for ($I_1$ during $I_2$) and ($I_3$ starts $I_4$)

In a similar way (Table 6), ($I_1$ during $I_2$) $\vee$ ($I_3$ starts $I_4$) $\equiv$

$(I_1^- \{(-\infty 0)_{\{Rd1\}} [0 \infty)_{\{Rd3\}}\} I_2^-), (I_1^+ \{(0 \infty)_{\{Rd2\}} (-\infty 0]_{\{Rd4\}}\} I_2^+),$

$(I_3^- \{[0 0]_{\{Rs1\}} (0 \infty)_{\{Rs3\}} (-\infty 0)_{\{Rs4\}}\} I_4^-), (I_3^+ \{(0 \infty)_{\{Rs2\}} (-\infty 0]_{\{Rs5\}}\} I_4^+),$

and {$R_{d1} R_{d2} R_{s1} R_{s2}$} and the Cartesian product {$R_{d3} R_{d4}$} X {$R_{s3} R_{s4} R_{s5}$} are I-L-Sets.

Therefore, logical relations on elemental constraints can be represented by a set of I-L-Sets. Thus, a labeled TCN (and the set of I-L-Sets) can represent a *special type of and/or* TCN. These types of non-binary constraints enrich the expressiveness of language and allow for the modeling of more complex problems (Meiri, 1996). Stergiou and Koubarakis (1996) and Jonsson and Bäckström (1998) show that Disjunctions of Linear Constraints (DLR) are also able to represent these non-binary constraints. However, Pujari and Sattar (1999) remark that general methods from linear programming should then be applied for DLR management, such that specific temporal concepts (like the ones detailed in Section 2) are not considered in these general methods. In the proposed model, management of these non-binary constraints are performed by the proposed reasoning methods without increasing their computational complexity. The added functionality is of interest in several temporal reasoning problems, including planning, scheduling and temporal constraint databases (Barber et al., 1994; Gerevini & Schubert, 1995; Brusoni et al., 1997; Stergiou & Koubarakis, 1998; etc.) where no general solutions are provided in the specific temporal reasoning area.

In addition, the proposed reasoning algorithms obtain a globally *labeled-consistent* TCN (*Theorem 11*). This feature allows us to manage *hypothetical queries*, which is an important requirement in query processes on temporal constraint databases (Brusoni et al., 1997). Thus, queries such as *Does $c'_{ij}$ hold, if $c'_{kl}$?* can be answered without any TCN propagation. The label set associated to each derived elemental constraint represents the set of input elemental constraints that should hold for the fulfillment of this elemental constraint. Therefore,

$$(x_k \ c'_{kl} \ x_l) \rightarrow (x_i \ c'_{ij} \ x_j)$$

holds, if $\forall elc_{kl.y} \in lc_{kl} / ec_{kl.y} \subseteq c'_{kl}$ then $\exists elc_{ij.x} \in lc_{ij} / ec_{ij.x} \subseteq c'_{ij}$ and labels($elc_{ij.x}$)$\subseteq$labels($elc_{kl.y}$) hold.

For example, from the labeled minimal TCN in Figure 7, we have:

$$(T_1 \ \{[40 \ 40]\} \ T_3) \rightarrow (T_2 \ \{ \ [0 \ 0] \ \} \ T_4), \qquad (T_3 \ \{ \ [20 \ 20] \ \} \ T_2) \rightarrow (T_3 \ \{ \ [20 \ 20] \ \} \ T_4).$$

However, $(T_3 \ \{[10 \ 20]\} \ T_2)$ does not imply $(T_1 \ \{[70 \ 70]\} \ T_4)$. Similarly, questions such as *'Can $c'_{ij}$ hold, if $c'_{kl}$?'* can also be easily answered by applying Theorem 9 and Theorem 10.

## 7. Alternative Temporal Contexts

When we reason on temporal facts, we can simultaneously work on different alternative temporal contexts, situations, trends, plans, intentions or possible worlds (Dousson et al., 1993; Garcia & Laborie, 1996). This is usual in a branching (backward or forward) model of time. Here, we can have alternative past contexts (i.e.: different lines about how facts may have occurred) or alternative future contexts (i.e.: different lines about how facts may occur). Thus, temporal context management is also required in hypothetical or causal reasoning. Also, having different contexts permits a partition of the whole TCN in a set of independent chains in order to decrease the complexity problem size (Gerevini & Schubert, 1995). In this section, we do not deal with hypothetical reasoning issues. Our goal is temporal management of *context-dependent constraints*. Thus, in general, a *hierarchy of alternative temporal contexts* can be established, such that constraints can be associated to different temporal contexts. For instance, Figure 13 represents a hierarchy of alternative contexts, where $W_0$ represents the root context and there are different disjunctive constraints between $(n_1, n_2)$ in each context. Temporal reasoning algorithms detailed in this paper are able to manage these context-dependent constraints:

- Input disjunctive constraints are asserted in different temporal contexts. To do this, the labels associated to input elemental constraints can also be used to represent the *context* in which the disjunctive is asserted. For instance (Figure 13), if the constraint:

$$(n_1 \ \{[0 \ 50]_{\{R1\}}, \ [200 \ 210]_{\{R2\}}\} \ n_2)$$

is asserted in context $W_1$, we have the following input *context-dependent* labeled constraint:

$$(n_1 \ \{[0 \ 25]_{\{R1, \ W1\}}, \ [260 \ 280]_{\{R2, \ W1\}}\} \ n_2).$$

Here, each *context-dependent* label set associated to each elemental constraint represents both the alternative temporal disjunct (i.e.: $R_1$ or $R_2$) and the context in which the elemental constraint is asserted ($W_1$).

- Label sets associated to context-dependent derived elemental constraints will represent the temporal contexts in which derived elemental constraints hold.

**Definition 8.** A *context-dependent disjunctive constraint* is a disjunctive constraint where each elemental constraint (i.e.: disjunct) is associated to an alternative temporal context. The universal labeled constraint is $\{(-\infty \ \infty)_{\{W0 \ R0\}}\}$, where $W_0$ is the root context. ◊

The proposed reasoning processes can manage *context-dependent* disjunctive constraints in a way similar to previously defined labeled disjunctive constraints (Section 3). For instance, according to the constraints and contexts in Figure 13, the following input labeled constraints between nodes $n_1$ $n_2$ should be updated:

$(n_1 \{[0\ 100]_{\{R1\ W0\}}, [200\ 300]_{\{R2\ W0\}}\}\ n_2)$,  $(n_1 \{[0\ 50]_{\{R3\ W1\}}, [200\ 210]_{\{R4\ W1\}}\}\ n_2)$,

$(n_1 \{[60\ 100]_{\{R5\ W2\}}, [290\ 300]_{\{R6\ W2\}}\}\ n_2)$,  $(n_1 \{[0\ 25]_{\{R7\ W3\}}, [260\ 280]_{\{R8\ W3\}}\}\ n_2)$,

$(n_1 \{\ [0\ 25]_{\{R0\ W11\}}\}\ n_2)$,  $(n_1 \{\ [30\ 50]_{\{R9\ W12\}}, [200\ 205]_{\{R10\ W12\}}\}\ n_2)$,

$(n_1 \{[0\ 20]_{\{R0\ W31\}}, [210\ 215]_{\{R0\ W32\}}\}\ n_2)$,  $(n_1 \{[260\ 280]_{\{R0\ W33\}}\}\ n_2)$.

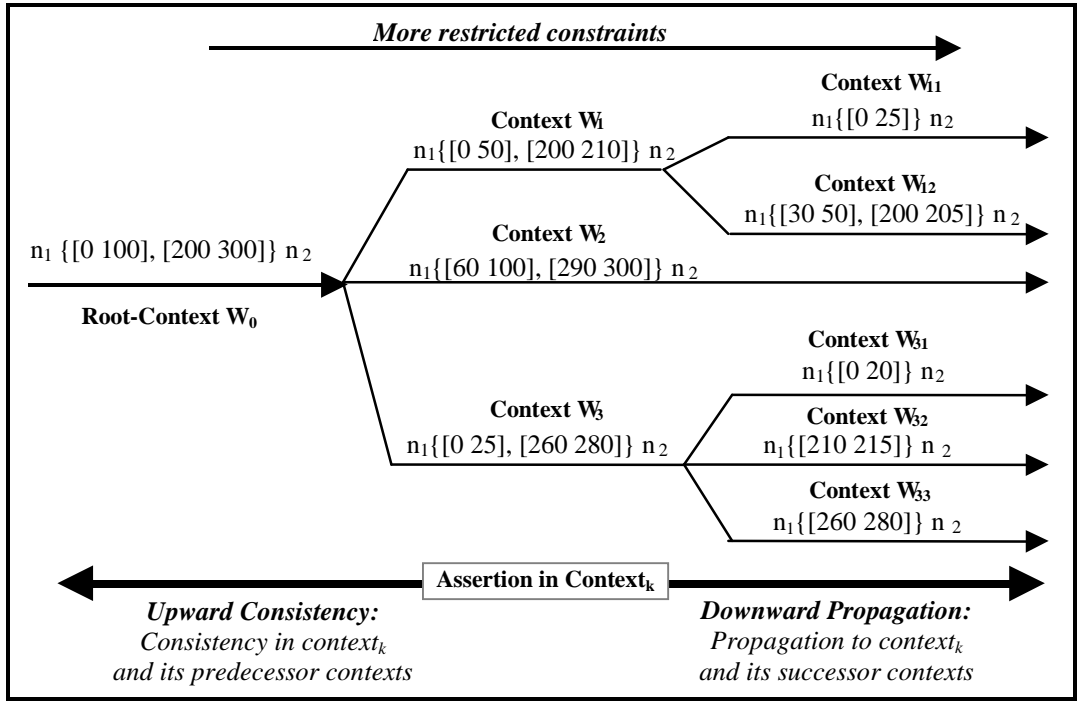

Figure 13: A hierarchy of alternative contexts

The updating process of each new constraint $c_{ij}$ in a given context $W_p$ should assure the consistency of $c_{ij}$ in the context $W_p$, as well as in its predecessor contexts (Figure 13). The consistency of $c_{ij}$ with the successor contexts of $W_p$ will be detailed in Section 7.2, since several options can be identified. However, it is not necessary to assure consistency among constraints belonging to contexts of different hierarchies. Successor contexts of a given context represent different alternatives, which are mutually exclusive. Thus, constraints belonging to contexts of different hierarchies can be mutually inconsistent. However, this does not imply that constraints in these contexts should necessarily be mutually disjoint. For instance (Figure 13), the constraints $(n_1 \{[0\ 50]_{\{R3\ W1\}}, [200\ 210]_{\{R4\ W1\}}\}\ n_2)$ in context $W_1$ and $(n_1 \{[0\ 25]_{\{R7\ W3\}}, [260\ 280]_{\{R8\ W3\}}\}\ n_2)$ in context $W_3$ are not mutually disjoint. However, $W_1$, $W_2$ and $W_3$ are assumed as three mutually exclusive alternatives of $W_0$.

The closure process of each new constraint $c_{ij}$ in context $W_p$ should downward propagate the new constraint $c_{ij}$ to all its successor contexts (Figure 13). Moreover, no propagation should be performed to the predecessor contexts of $context_k$, nor among contexts of different hierarchies. Elemental constraints belonging to contexts of different hierarchies cannot be simultaneously considered, that is, combined or intersected.

## 7.1 Context-Dependent Updating and Closure Processes

The update and closure processes defined in Section 4 should be adapted in order to manage context-dependent disjunctive constraints. The *Context-Update* process (Figure 14) asserts the constraint $c'_{ij} \equiv \{ec'_1, ec'_2, ..., ec'_n\}$ in the context *context$_k$*. In a way similar to the updated process described in Section 4, Context-Update should be performed each time a new context-dependent constraint is asserted.

---

**Context-Update ($n_i$ $c'_{ij}$ $n_j$ context$_k$)**

  $lc'_{ij} \leftarrow$ Put-label-context ($c'_{ij}$, context$_k$)        *;Labelling and mutual inconsistency.*

  **If** Consistency-Test (get-upward ($n_i$, $n_j$, context$_k$), $lc'_{ij}$)     *;Upwards Consistency test*

     **Then** *(*Inconsistent Constraint*)*

       Return (false)

     **Else** *(*Consistent Constraint*)*         *;lc'$_{ij}$ is asserted in the context$_k$ and in all its*

       $lc_{ij} \leftarrow (lc_{ij}$ - get ($n_i$, $n_j$, context$_k$)) $\cup_{lc}$ ($lc_{ij} \oplus_{lc} lc'_{ij}$),        *;successor contexts.*

       $lc_{ji} \leftarrow$ Inverse$_{lc}$ ($lc_{ij}$).

       Context-Closure ($n_i$ $lc_{ij}$ $n_j$ context$_k$)       *;Downwards Closure algorithm in context$_k$.*

.        Return (true)

  **End-If**

**End-Context-Update**

---

Figure 14: Context-Update process for context-dependent labeled constraints

Where:

- ***Put-label-context (c'$_{ij}$, context$_k$)*** associates an exclusive *label set* to each elemental constraint $ec'_{ij,p} \in c'_{ij}$. This label set has two labels $\{R_{ij,p}$, context$_k\}$. In this label set, the first label is the label associated to each temporal disjunct. In a way similar to *Put-labels* function, these labels are mutually exclusive (*Definition 3*). The second label represents the context in which $c'_{ij}$ is updated. Moreover, each pair of labels associated to successor contexts of the parent context of context$_k$ is added to the I-L-Sets, since all the successor contexts of a given context are mutually exclusive:

$$\forall context_p \ / \ context_p \in \text{Succesor-Contexts(Parent-Context(Context}_k)),$$
$$\text{I-L-Sets} \leftarrow \text{I-L-Sets} \cup (\{context_k\} \cup \{context_p\}).$$

Where Parent-Context(context$_k$) and Successor-Contexts(context$_k$) return the parent-context and the set of successor-contexts of context$_k$, respectively. Thus, in Figure 13, $\{\{W_1, W_2\}, \{W_1, W_3\}, \{W_2, W_3\}, \{W_{11}, W_{12}\}, \{W_{31}, W_{32}\}, \{W_{31}, W_{33}\}, \{W_{32}, W_{33}\}\}$ are I-L-Sets.

- **get ($n_i$, $n_j$, context$_k$)** returns the set of labeled elemental constraints between $n_i$ and $n_j$ in the context$_k$ (and in all its successor contexts). That is:

  get ($n_i$, $n_j$, context$_k$)::= {(ec$_{ij,p}$ {label$_{ij,p}$})∈ lc$_{ij}$ / context$_k$∈ {label$_{ij,p}$} }.

  Note that get($n_i$, $n_j$, context$_k$) is a subset of lc$_{ij}$. Thus, (lc$_{ij}$ - get ($n_i$, $n_j$, context$_k$)) means the set-difference between lc$_{ij}$ and get ($n_i$, $n_j$, context$_k$). That is, the set of elemental constraints in the context-dependent constraint lc$_{ij}$, which are not in context$_k$, nor in any of its successor contexts.

- **get-upward ($n_i$, $n_j$, context$_k$)**, similarly to the previous *get* function, it returns the existing constraints between $n_i$ and $n_j$ in the context$_k$ (and in all its successor contexts). However, if there is no constraint between $n_i$ and $n_j$ in the context$_k$, then the function returns the constraints between $n_i$ and $n_j$ that exist in the predecesor context of *context$_k$*:

  **get-upward ($n_i$, $n_j$, context$_k$)** ::=
  If get ($n_i$, $n_j$, context$_k$) ≠ ∅ Then return (get ($n_i$, $n_j$, context$_k$))
  Else
      Context$_k$ ← Parent-Context (Context$_k$)
      Until get ($n_i$, $n_j$, context$_k$) ≠ ∅ ∨ Context$_k$=W$_0$ do
          If get ($n_i$, $n_j$, context$_k$) ≠ ∅    Then return (get ($n_i$, $n_j$, context$_k$))
                           Else  return({(-∞ +∞)}$_{\{W0\ R0\}}$)

      **End-get-upward**

The context-dependent closure (Figure 15) process is similar to the closure process described in Section 4 and it is also performed at each updating process. The closure process of each updated constraint in context$_k$ is downwards performed in context$_k$ and in all its successor contexts.

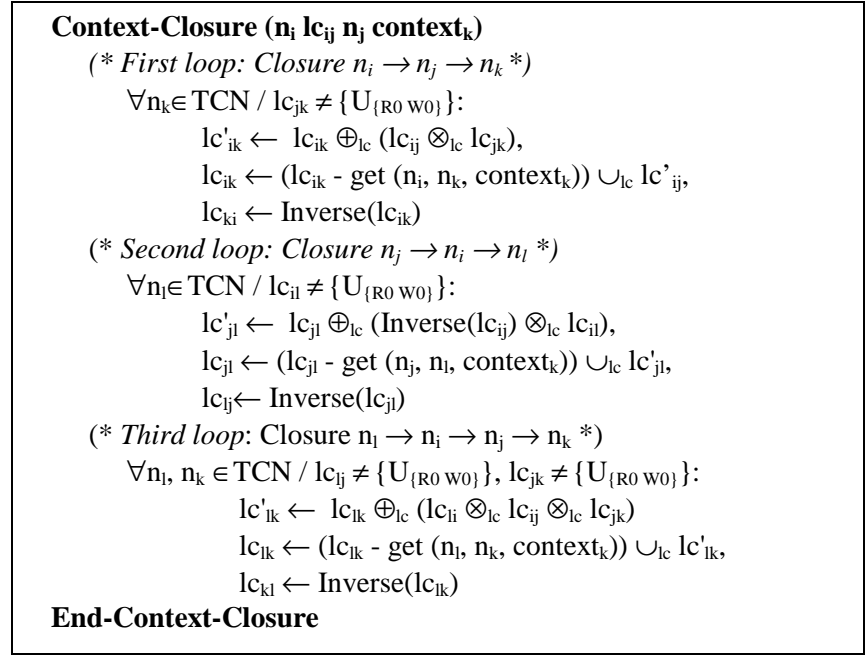

**Context-Closure ($n_i$ lc$_{ij}$ $n_j$ context$_k$)**
    (* *First loop: Closure* $n_i \rightarrow n_j \rightarrow n_k$ *)
        ∀n$_k$∈ TCN / lc$_{jk}$ ≠ {U$_{\{R0\ W0\}}$}:
            lc'$_{ik}$ ← lc$_{ik}$ ⊕$_{lc}$ (lc$_{ij}$ ⊗$_{lc}$ lc$_{jk}$),
            lc$_{ik}$ ← (lc$_{ik}$ - get ($n_i$, $n_k$, context$_k$)) ∪$_{lc}$ lc'$_{ij}$,
            lc$_{ki}$ ← Inverse(lc$_{ik}$)
    (* *Second loop: Closure* $n_j \rightarrow n_i \rightarrow n_l$ *)
        ∀n$_l$∈ TCN / lc$_{il}$ ≠ {U$_{\{R0\ W0\}}$}:
            lc'$_{jl}$ ← lc$_{jl}$ ⊕$_{lc}$ (Inverse(lc$_{ij}$) ⊗$_{lc}$ lc$_{il}$),
            lc$_{jl}$ ← (lc$_{jl}$ - get ($n_j$, $n_l$, context$_k$)) ∪$_{lc}$ lc'$_{jl}$,
            lc$_{lj}$← Inverse(lc$_{jl}$)
    (* *Third loop:* Closure $n_l \rightarrow n_i \rightarrow n_j \rightarrow n_k$ *)
        ∀n$_l$, n$_k$ ∈ TCN / lc$_{lj}$ ≠ {U$_{\{R0\ W0\}}$}, lc$_{jk}$ ≠ {U$_{\{R0\ W0\}}$}:
            lc'$_{lk}$ ← lc$_{lk}$ ⊕$_{lc}$ (lc$_{li}$ ⊗$_{lc}$ lc$_{ij}$ ⊗$_{lc}$ lc$_{jk}$)
            lc$_{lk}$ ← (lc$_{lk}$ - get ($n_l$, $n_k$, context$_k$)) ∪$_{lc}$ lc'$_{lk}$,
            lc$_{kl}$ ← Inverse(lc$_{lk}$)
**End-Context-Closure**

Figure 15: Context-Closure process for context-dependent labeled constraints

The resulting label set associated to each context-dependent derived elemental constraint represents the contexts where the elemental constraint holds, as well as the hierarchy of predecessor contexts of the elemental constraint. For instance, Figure 16 shows the contextual labeling for the example in Figure 13. Moreover, after successively performing the updating and closure processes for all constraints in this example, we have the following constraint between nodes $n_1$ and $n_2$:

$(n_1 \ lc_{12} \ n_2)$:  $(n_1$  $\{[0 \ 100]_{\{R1 \ W0\}}, [200 \ 300]_{\{R2 \ W0\}}, [0 \ 50]_{\{R3 \ R1 \ W1 \ W0\}}, [200 \ 210]_{\{R4 \ R2 \ W1 \ W0\}},$  ***(e3)***
$[60 \ 100]_{\{R5 \ R1 \ W2 \ W0\}}, [290 \ 300]_{\{R6 \ R2 \ W2 \ W0\}}, [0 \ 25]_{\{R7 \ R1 \ W3 \ W0\}}, [260 \ 280]_{\{R8 \ R2 \ W3 \ W0\}},$
$[0 \ 25]_{\{R0 \ R3 \ R1 \ W11 \ W1 \ W0\}}, [30 \ 50]_{\{R9 \ R3 \ R1 \ W12 \ W1 \ W0\}}, [200 \ 205]_{\{R10 \ R2 \ R4 \ W12 \ W1 \ W0\}},$
$[0 \ 20]_{\{R0 \ R7 \ R1 \ W31 \ W3 \ W0\}}, [210 \ 215]_{\{R0 \ R2 \ R8 \ W32 \ W3 \ W0\}}, [260 \ 280]_{\{R0 \ R2 \ R8 \ W33 \ W3 \ W0\}}\}$  $n_2)$

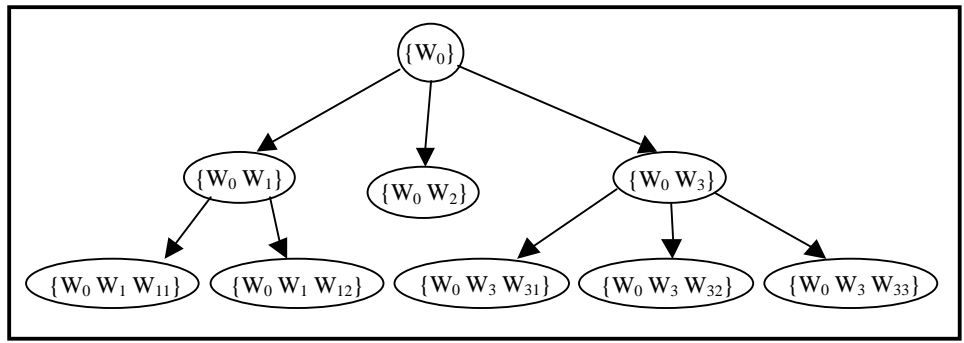

Figure 16: Labels in contexts

No closure process is performed among constraints belonging to contexts of different hierarchies. According to *Put-label-context* function, each pair of labels related to the successor contexts of each context is an I-L-Set. Thus, these I-L-Sets prevent deriving elemental constraints from contexts of different hierarchies. That is, each derived elemental constraint obtained (combining or intersecting) from two elemental constraints in contexts of different hierarchy will have an inconsistent associated label set. Therefore, these derived elemental constraints will be rejected in the operation $\cup_{lc}$. For instance, in the example of Figure 13, $\{\{W_1, W_2\}, \{W_1, W_3\}, \{W_2, W_3\}, \{W_{11}, W_{12}\}, \{W_{31}, W_{32}\}, \{W_{31}, W_{33}\}, \{W_{32}, W_{33}\}\}$ are I-L-Sets. Thus, if a constraint is asserted in context $W_1$:

i)  No propagation is performed using constraints in contexts $W_{11}$ and $W_{12}$ simultaneously, since $\{W_{11}, W_{12}\}$ is an I-L-Set.

ii)  No propagation is performed in context $W_2$, nor in $W_3$, nor in their successors, since $\{W_1, W_2\}$ and $\{W_1 \ W_3\}$ are I-L-Sets.

Let's see an example of the Context-Update and Context-Closure processes. Let's assume that the context-dependent constraints in Figure 13 are already updated and closured, such that the previous constraint $lc_{12}$ *(expression e3)* exists between $n_1$ and $n_2$. Now, we update $(n_1 \ \{[20 \ 40]\} \ n_2)$ in context $W_1$. The call to *Consistency-Test* function in the Context-Update function is:

Consistency-Test (get-upward $(n_1, n_2, W_1)$, $\{[20 \ 40]_{\{R0 \ W1\}}\}$).

Given the previous constraint $lc_{12}$ between $n_1$ and $n_2$ *(expression e3)*, the function performs:

$\{[0 \ 50]_{\{R3 \ R1 \ W1 \ W0\}}, [200 \ 210]_{\{R4 \ R2 \ W1 \ W0\}}, [0 \ 25]_{\{R0 \ R3 \ R1 \ W11 \ W1 \ W0\}},$

$$[30\ 50]_{\{R9\ R3\ R1\ W12\ W1\ W0\}},\ [200\ 205]_{\{R10\ R2\ R4\ W12\ W1\ W0\}}\}\ \oplus_{lc}\ \{[20\ 40]_{\{R0\ W1\}}\}=$$

$$\{[20\ 40]_{\{R3\ R1\ R0\ W1\ W0\}},\ [20\ 25]_{\{R0\ R3\ R1\ W11\ W1\ W0\}},\ [30\ 40]_{\{R9\ R3\ R1\ R0\ W12\ W1\ W0\}}\}\neq\varnothing$$

Thus, the new constraint ($n_1$ $\{[20\ 40]\}$ $n_2$) is consistent in context $W_1$. Therefore, the constraint between $n_1$ $n_2$ results:

$lc_{12}\leftarrow(lc_{12}$ - get ($n_1$, $n_2$, $W_1$)) $\cup_{lc}(lc_{12}\oplus_{lc}\{[20\ 40]_{\{R0\ W1\}}\})=$

$$\{[0\ 100]_{\{R1\ W0\}},\ [200\ 300]_{\{R2\ W0\}},\ [60\ 100]_{\{R5\ R1\ W2\ W0\}},\ [290\ 300]_{\{R6\ R2\ W2\ W0\}},$$

$$[0\ 25]_{\{R7\ R1\ W3\ W0\}},\ [260\ 280]_{\{R8\ R2\ W3\ W0\}},\ [0\ 20]_{\{R0\ R7\ R1\ W31\ W3\ W0\}},$$

$$[210\ 215]_{\{R0\ R2\ R8\ W32\ W3\ W0\}},\ [260\ 280]_{\{R0\ R2\ R8\ W33\ W3\ W0\}}\}\ \cup_{lc}$$

$$\{[20\ 40]_{\{R1\ R0\ W1\ W0\}},\ [20\ 40]_{\{R3\ R1\ R0\ W1\ W0\}},\ [20\ 25]_{\{R0\ R3\ R1\ W11\ W1\ W0\}},\ [30\ 40]_{\{R9\ R3\ R1\ R0\ W12\ W1\ W0\}}\}=$$

$$\{[0\ 100]_{\{R1\ W0\}},\ [200\ 300]_{\{R2\ W0\}},\ [60\ 100]_{\{R5\ R1\ W2\ W0\}},\ [290\ 300]_{\{R6\ R2\ W2\ W0\}},\ [0\ 25]_{\{R7\ R1\ W3\ W0\}},\qquad\textbf{\textit{(e4)}}$$

$$[260\ 280]_{\{R8\ R2\ W3\ W0\}},\ [0\ 20]_{\{R0\ R7\ R1\ W31\ W3\ W0\}},\ [210\ 215]_{\{R0\ R2\ R8\ W32\ W3\ W0\}},\ [260\ 280]_{\{R0\ R2\ R8\ W33\ W3\ W0\}},$$

$$[20\ 40]_{\{R1\ R0\ W1\ W0\}},\ [20\ 25]_{\{R0\ R3\ R1\ W11\ W1\ W0\}},\ [30\ 40]_{\{R9\ R3\ R1\ R0\ W12\ W1\ W0\}}\}.$$

Note that the new updated constraint is asserted in context $W_1$ and propagated to all its successor contexts ($W_{11}$ and $W_{12}$). However, the new constraint in context $W_1$ does not affect the existing constraints in predecessor contexts of $W_1$ ($W_0$) nor the constraints belonging to contexts of different hierarchies ($W_2$, $W_3$ and their successors).

In this update process, no closure process is performed, since no node is related with $n_1$ or $n_2$. Now, let's update ($n_3$ $\{[10\ 20]\}$ $n_1$) in context $W_1$. We have:

Consistency-Test (get-upward ($n_3$, $n_1$, $W_1$), $\{[10\ 20]_{\{R0\ W1\}}\}$),

that performs:

$$\{(-\infty\ +\infty)\}_{\{W0\ R0\}}\oplus_{lc}\{[20\ 40]_{\{R0\ W1\}}\}=\{[20\ 40]_{\{R0\ W0\ W1\}}\}\neq\varnothing,$$

since no previous constraint exists between ($n_3$ $n_1$) in context $W_1$. The constraint ($n_3$ $\{[10\ 20]\}$ $n_1$) is consistent, and asserted in the TCN:

$$lc_{31}\leftarrow\{(-\infty\ +\infty)\}_{\{W0\ R0\}},\ [20\ 40]_{\{R0\ W0\ W1\}}\}.\qquad\textbf{\textit{(e5)}}$$

Afterwards, this constraint is closed. The call to Context-Closure process is:

Context-Closure ($n_3$, $\{(-\infty\ +\infty)\}_{\{W0\ R0\}},\ [20\ 40]_{\{R0\ W0\ W1\}}\}$, $n_1$, $W_1$).

In this closure process, only the first loop is performed since no node is related to $n_3$. Moreover, only the previous constraint $lc_{12}$ (expression *e4*) exists in the current TCN between $n_1$ and $n_2$. Thus, the first loop performs:

$lc'_{32}\leftarrow lc_{32}\oplus_{lc}(\{(-\infty\ +\infty)\}_{\{W0\ R0\}},\ [20\ 40]_{\{R0\ W0\ W1\}}\}\otimes_{lc}lc_{12})=$

$$\{(-\infty\ \infty)_{\{W0\ R0\}}\}\oplus_{lc}(\{(-\infty\ +\infty)\}_{\{R0\ W0\ W1\}},\ [20\ 40]_{\{R0\ W0\ W1\}}\}\otimes_{lc}lc_{12})=$$

$$\{(-\infty\ +\infty)\}_{\{W0\ R0\}},\ [220\ 340]_{\{R2\ R0\ W0\ W1\}},\ [40\ 80]_{\{R1\ R0\ W1\ W0\}},$$

$$[40\ 65]_{\{R0\ R3\ R1\ W11\ W1\ W0\}},\ [50\ 80]_{\{R9\ R3\ R1\ R0\ W12\ W1\ W0\}}\},$$

such that,

$lc_{32}\leftarrow(lc_{32}$ - get ($n_3$, $n_2$, $W_1$)) $\cup_{lc}lc'_{32}=(\{(-\infty\ \infty)\}_{\{W0\ R0\}}\}$ - $\{\})\cup_{lc}lc'_{32}=$

$$\{(-\infty\ \infty)\}_{\{W0\ R0\}},\ [220\ 340]_{\{R2\ R0\ W0\ W1\}},\ [40\ 80]_{\{R1\ R0\ W1\ W0\}},$$

$$[40\ 65]_{\{R0\ R3\ R1\ W11\ W1\ W0\}},\ [50\ 80]_{\{R9\ R3\ R1\ R0\ W12\ W1\ W0\}}\}.\qquad\textbf{\textit{(e6)}}$$

Thus, the asserted constraint between ($n_3$, $n_2$) in context $W_1$ is closured in the context $W_1$ and in all its successor contexts ($W_{11}$ and $W_{12}$). Likewise, the closure process does not perform any propagation simultaneously using constraints of the contexts $W_{11}$ and $W_{22}$, nor any of the context $W_2$, $W_3$, nor any of their successors.

## 7.2 Complete Versus Incomplete Partition of Contexts

In each updating process, the consistency of each new constraint lc'$_{ij}$ in a given context is assured in this context and in all its parent contexts. Let's deal with consistency issues between a context and its successor contexts. Here, we have that constraints in a given context $W_i$ can be either completely covered or only partially covered by the existing constraints in the successor contexts of $W_i$. That is, the successor contexts of $W_i$ can be either a complete partition or only a partial partition of $W_i$.

For instance, let's assert the constraint ($n_1$ {[210 210]$_{\{R0\,W1\}}$} $n_2$) in the context $W_1$ of the example in Figure 13. In the *Consistency-test* function, we have (where the constraint lc$_{12}$ is the previous expression *e2*):

get-upward ($n_1$, $n_2$, $W_1$) $\oplus_{lc}$ {[210 210]$_{\{R0\,W1\}}$} =

{[0 50]$_{\{R3\,R1\,W1\,W0\}}$, [200 210]$_{\{R4\,R2\,W1\,W0\}}$, [0 25]$_{\{R0\,R3\,R1\,W11\,W1\,W0\}}$, [30 50]$_{\{R9\,R3\,R1\,W12\,W1\,W0\}}$,

[200 205]$_{\{R10\,R2\,R4\,W12\,W1\,W0\}}$} $\oplus_{lc}$ {[210 210]$_{\{R0\,W1\}}$} = {[210 210]$_{\{R0\,W1\,R4\,R2\,W0\}}$}.

That is, the asserted constraint is consistent with the existing constraints in context $W_1$. However, no resulting elemental constraint is associated to context $W_{11}$ nor $W_{12}$. This means that the asserted constraint ($n_1$ {[210 210]$_{\{R0\,W1\}}$} $n_2$) is consistent in $W_1$, but is inconsistent in $W_{11}$ and in $W_{12}$. Here, two alternatives appear:

i) To assume that existing successor contexts are a complete partition of their parent context. Therefore, a new constraint $c_{ij}$ in a context $W_i$ should be rejected, if $c_{ij}$ is inconsistent in all successor contexts of $W_i$. For instance, we can assume that $W_{11}$ and $W_{12}$ in Figure 13 are a complete partition of $W_1$. Thus, ($n_1$ {[210 210]$_{\{R0\,W1\}}$} $n_2$) should be rejected.

ii) To assume that successor contexts are not a complete partition of their parent context. Therefore, successor contexts become inconsistent and they should be removed. In the example, we can assume that contexts $W_{11}$ and $W_{12}$ are not a complete partition of the context $W_1$, such that another possible new successor context of $W_1$ would be able to match in the future the asserted constraint ($n_1$ {[210 210]$_{\{R0\,W1\}}$} $n_2$). In this case, the constraint ($n_1$ {[210 210]$_{\{R0\,W1\}}$} $n_2$) is assumed to be correct, such that it can be asserted in the TCN. Therefore, the contexts $W_{11}$ and $W_{12}$ become inconsistent. {$W_{11}$} and {$W_{12}$} should be added to the set of I-L-Sets, such that these contexts (and all their successor contexts and all their constraints) become inconsistent and removed from the TCN. That is, all elemental constraints with an associated label set containing {$W_{11}$} or {$W_{12}$} should be removed.

In both cases, each context will always be consistent with all its successor contexts. The option to be adopted can depend on the problem type to solve (Garrido et al., 1999). Any of the these options can be easily introduced in the described reasoning processes, since the function *Consistency Test* can determine which successor contexts ($W_s$) become inconsistent at each new constraint (lc'$_{ij}$) in a context ($W_k$):

$W_s \in$ Successor-Contexts($W_k$) / $\exists elc_{ij,p} \in$ get-upward ($n_i$, $n_j$, $W_k$), $W_s \in \{label_{ij,p}\}$ $\land$

$\neg \exists elc_{ij,r} \in$ (get-upward ($n_i$, $n_j$, $W_k$) $\oplus_{lc}$ lc'$_{ij}$), $W_s \in \{label_{ij,r}\}$.

On the other hand, when: *(i)* the successor contexts ($W_{k1}$, $W_{k2}$, ..., $W_{kp}$) of a context $W_k$ are a complete partition of it, and *(ii)* all constraints in ($W_{k1}$, $W_{k2}$, ..., $W_{kp}$) have been asserted, then constraints in $W_k$ can be restricted according to the final existing constraints in ($W_{k1}$, $W_{k2}$, ..., $W_{kp}$). To do this, the context $W_k$ should be constrained by the temporal union of the constraints in all its successor contexts.

### 7.3  A Minimal and Consistent Context-Dependent TCN

**Definition 9.** A context-dependent TCN is minimal (and consistent) if the constraints in each context are consistent (with respect to constraints in this context, in all its predecessor contexts, and all its successor contexts) and minimal (with respect to constraints in this context and in all its predecessor contexts). ◊

**Theorem 12.** At each updating process, the context-dependent reasoning processes obtain a minimal (and consistent) context-dependent TCN if the previous context-dependent TCN is minimal.

*Proof:* If the previous context-dependent TCN is minimal, the *Consistency-Test* function guarantees the consistency of each new context-dependent input constraint:

i)    in its context and in all its parent contexts (get-upward function and *Theorem 5*),

ii)   in all its successor contexts (depending of the two identified cases in Section 7.2).

The closure process of a new constraint in a given context ($W_k$) propagates its effects to this context and to all its successor contexts. Therefore (*Theorem 7*), the process obtains the new minimal constraints in this context ($W_k$) and in all its successor contexts. ◊

Moreover, the obtained *context-dependent* TCN is *globally labeled-consistent*. Thus, we can deduce whether a set of elemental constraints (between different pairs of time points) is consistent (*Theorem 10*). That is, this set of elemental constraints holds in some context. For instance, given the previous constraints $lc_{12}$, $lc_{31}$ and $lc_{32}$ (previous expressions e4, e5 and e6), we can deduce that:

$$(n_1 \{[40\ 40]\}\ n_2) \land (n_3 \{[40\ 40]\}\ n_1) \land (n_3 \{[40\ 40]\}\ n_2)$$

is full consistent since:

$\exists elc_{12,x} \in lc_{12}$, $\exists elc_{31,y} \in lc_{31}$, $\exists elc_{32,z} \in lc_{32}$ / ($\{label_{12,x}\} \cup \{label_{12,x}\} \cup \{label_{12,x}\}$) is not an I-L-Set.

Specifically, these instantiations hold in $\{R_1\ R_0\ W_1\ W_0\}$ and $\{R_1\ R_0\ W_0\}$. Thus, this set of elemental constraints holds in context W1 (and, obviously, in all its predecessor contexts).

Likewise, from a minimal context-dependent TCN, the user can retrieve the constraints that hold in each context or the constraints that simultaneously hold in a set of given contexts. To do this, the *Context-Constraints* function retrieves the constraints that hold between a pair of nodes ($n_i$, $n_j$) in a given context (context$_k$). That is, the result of Get-upwards($n_i$, $n_j$, context$_k$) except those elemental constraints belonging to successor contexts of context$_k$:

***Context-Constraints ($n_i$, $n_j$, context$_k$)***::= Get-upwards ($n_i$, $n_j$, context$_k$) –

$\{lec_{ij \cdot p} \in lc_{ij}$ / $\exists context_q \in$ Succesor-Contexts(context$_k$), $\{context_q\} \cap \{label_{ij \cdot p}\} \neq \varnothing\}$.

For instance, given the context-dependent constraint $lc_{12}$ in Figure 13 *(expression e3)*, the following constraint would hold between ($n_1$, $n_2$) in both contexts $W_1$ and $W_3$:

Context-Constraints($n_1$, $n_2$, $W_1$) $\oplus_{lc}$ Context-Constraint($n_1$, $n_2$, $W_3$) =

$\{[0\ 50]_{\{R3\ R1\ W1\ W0\}}, [200\ 210]_{\{R4\ R2\ W1\ W0\}}\} \oplus_{lc} \{[0\ 25]_{\{R7\ R1\ W3\ W0\}}, [260\ 280]_{\{R8\ R2\ W3\ W0\}}\} =$

$\{[0\ 25]_{\{R7\ R3\ R1\ W3\ W1\ W0\}}\}^6$.

In addition, we can obtain the constraints, which simultaneously hold in a context and in *any* of its successor ones. For instance, in context $W_1$ and in any of its successor contexts ($W_{11}$, $W_{12}$), the following constraint holds:

Context-Constrains($n_1$, $n_2$, $W_1$) $\oplus_{lc}$ [Context-Constraints($n_1$, $n_2$, $W_{11}$) $\cup_{lc}$ Context-Constraints($n_1$, $n_2$, $W_{12}$)]=

$\{[0\ 50]_{\{R3\ R1\ W1\ W0\}}, [200\ 210]_{\{R4\ R2\ W1\ W0\}}\} \oplus_{lc}$

$\{[0\ 25]_{\{R9\ R3\ R1\ W11\ W1\ W0\}}\} \cup_{lc} \{[30\ 50]_{\{R9\ R3\ R1\ W12\ W1\ W0\}}, [200\ 205]_{\{R10\ R2\ R4\ W12\ W1\ W0\}}\} =$

$\{[200\ 205]_{\{W12\ R10\ R4\ R2\ W1\ W0\}}, [0\ 25]_{\{W11\ R0\ R3\ R1\ W1\ W0\}}, [30\ 50]_{\{W12\ R9\ R3\ R1\ W1\ W0\}}\}$.

On the other hand, each alternative context ($W_i$) can be associated to an alternative hypothesis ($H_i$). Each hypothesis $H_i$ gives rise to a set of constraints, which will be asserted in the associated context $W_i$. Thus, the proposed reasoning processes assure minimal constraints in the hierarchy of hypotheses. Moreover, if a hypothesis ($H_i$) becomes unavailable, then the label set $\{W_i\}$ should be added to the set of I-L-Sets. Thus, all constraints in context $W_i$ (and in all its successor contexts) will be removed. That is, all constraints that depend on the unavailable hypothesis $H_i$ will be removed.

## 7.4 Computational Complexity of Temporal Context Management

The management of temporal context does not increase the complexity of the reasoning processes detailed in Section 4. In fact, we can consider that each label associated to a disjunct ($R_i$) in labeled disjunctive constraints is also associated to a context ($W_i$). Thus, the computational cost of each updating process is also bounded by $O(n^2\ l^{2e})$, where *'l'* is the maximum number of input disjuncts between any pair of nodes in all contexts.

The temporal labeled algebra proposed in this paper (Section 3) has been applied on the point-based disjunctive metric constraints (Dechter, Meiri & Pearl, 1991). However, this labeled algebra can also be applied on other temporal constraints. In this case, the operations $\oplus_{lc}$, $\otimes_{lc}$, $\cup_{lc}$ and $\subseteq_{lc}$ should be specified (Section 3) on the basis of the operations $\oplus$, $\otimes$, $\cup_T$ and $\subseteq_T$ of the underlying algebra. In this way, the management of temporal contexts can also be applied to other types of constraints.

**Theorem 13**. The computational complexity of the proposed reasoning process applied to context-dependent non-disjunctive metric constraints is polynomial ($O(n^2\ W^2)$) in the number W of managed contexts.

---

[6] However, note that this is an impossible situation, since $W_1$ and $W_3$ are mutually exclusive contexts. That is, $\{W_3, W_1\}$ is an I-L-Set.

**Proof**: Disjunctions in constraints are only related to the contexts in which input constraints are asserted, if non-disjunctive constraints are managed. That is, constraints between each pair of nodes are in the form:

$$(n_i \{(ec_{ij.0}\{W_0 \ R_0\}), (ec_{ij.1}\{W_1 \ R_0\}), \ ...... \ , (ec_{ij.k}\{W_k \ R_0\})\} \ n_j), \qquad 0 \le k \le W \ / \ W = |\{W_i\}|$$

Thus, the maximum number of disjuncts in constraints is bounded by the maximum number of managed contexts W. Moreover, the maximum length of associated label sets is the maximum depth in the hierarchy of contexts, and the set of I-L-Sets has only 2-length sets (i.e.: pairs of labels associated to each pair of successor contexts of each context). Therefore, the computational cost of operations $\otimes_{lc}$ and $\oplus_{lc}$ is bounded by $O(W^2)$. ◊

The methods proposed in Section 7.1 for management of temporal contexts can also be applied to other temporal reasoning algorithms, instead of the reasoning methods detailed in Section 4. This requires that these other reasoning algorithms be based on the operations of composition and intersection of temporal constraints. Thus,

i) Each elemental constraint should only be associated to the context $(W_i)$ in which it is asserted[7]. Thus, label sets associated to elemental constraints have only *one* contextual label $\{W_i\}$.

ii) The methods for management of temporal contexts described in Section 7.1 should be integrated into the new reasoning algorithms. These algorithms should use the operations $\oplus_{lc}$, $\otimes_{lc}$, *get* and *get-upwards*. The computational cost of operations $\oplus_{lc}$ and $\otimes_{lc}$ related to management of temporal contexts is polynomial ($O(W^2)$) in the number (W) of managed contexts. Therefore, the computational cost of the reasoning algorithms is increased by a factor $W^2$ when temporal contexts are managed.

For instance, when interval-based constraints are managed, the TCA algorithm can be used to obtain a path-consistent context-dependent IA-TCN, with a $O(n^3 \ W^2)$ cost. Similarly, when a context-dependent reasoning is applied to PIDN networks (Pujari & Sattar, 1999), the computational cost of specific reasoning algorithms on PIDN constraints is increased by a factor $W^2$. When the proposed temporal algebra in Section 3 is applied to tractable classes of constraints, the specific reasoning algorithms for management of these classes of constraints can also be applied. The computational cost of these reasoning algorithms (which should be based on combination and intersection operations on constraints) is increased by a polynomial factor $W^2$. For instance, when non-disjunctive metric constraints are managed, the TCA algorithm can be used as the closure algorithm in Section 7.1. This algorithm will obtain a minimal context-dependent TCN with a computational cost $O(n^3 \ W^2)$.

## 8. Conclusions

Several problems remain pending in representation and reasoning problems on temporal constraints. In relation to this, we have dealt with reasoning on complex qualitative and quantitative constraints between time-points and intervals, which can be organized in a hierarchy of alternative temporal contexts. We have described a new-labeled temporal algebra, whose main elements are *labeled disjunctive metric constraints*, label sets associated to elemental constraints, and sets of inconsistent elemental constraints (I-L-Sets). The temporal model presented is able to integrate qualitative and metric constraints on time-points and intervals. In fact, symbolic and metric constraint between intervals can be represented by means of disjunctive metric constraints between time points and a set of I-L-Sets. The model is also able to manage (*non-binary*) logical relations among elemental constraints. The reasoning algorithms on the described model are based on the distributive property for composition over intersection in labeled constraints, and guarantee consistency and obtain a minimal TCN of disjunctive metric point-based constraints. In addition, a special type of global labeled-consistent TCN is also obtained.

---

[7] That is, there are not labels $(R_i)$ associated to disjunctions in disjunctive constraints. Thus, Definition 3 is not applied in the *Put-Label-Context* function. Therefore, the distributive property for $\otimes_k$ over $\oplus_k$ does not hold for disjunctive constraints. However, this is not relevant since other reasoning processes will be applied.

Labeled constraints can be organized in a hierarchy of alternative *temporal contexts*, such that temporal reasoning processes can be performed on these contexts. Reasoning algorithms guarantee consistency in each hierarchy of contexts, maintain a minimal context-dependent TCN, and allow us to determine what constraints hold in each context or in a set of alternative contexts. Thus, we can reason on a hierarchy of context-dependent constraints on intervals, points and unary durations (Figure 17).

These described features are useful functionalities for modeling important problems in the temporal reasoning area. However, they have not been identified in previous models. Therefore, the temporal model presented here represents a flexible framework for reasoning on complex, context-dependent, metric and qualitative constraints on time-points, intervals and unary durations.

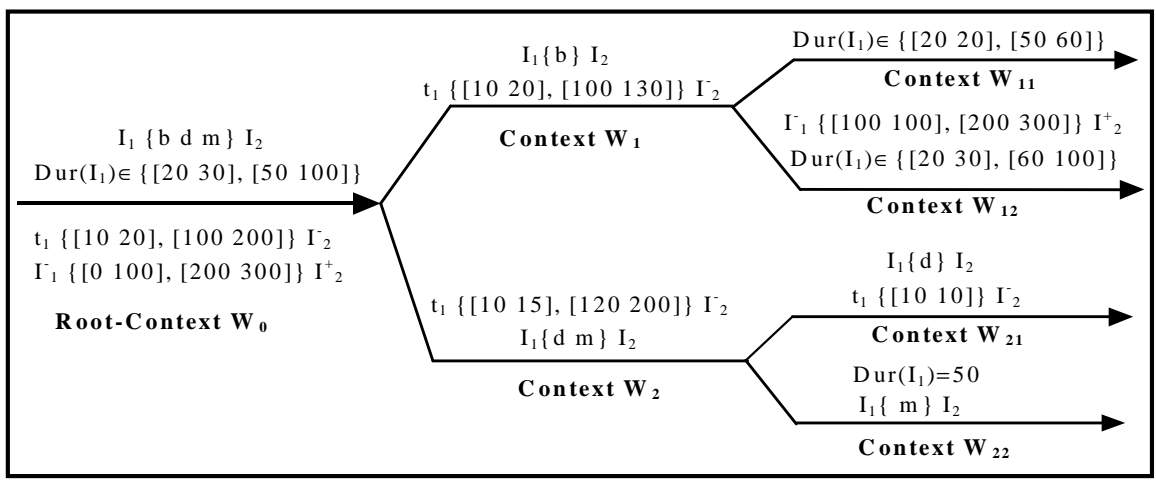

Figure 17: Context-dependent constraints on intervals, time points and unary durations

A path-consistent algorithm can be used as the closure process on labeled TCNs, like the typical TCA algorithm as applied by Allen (1983). This path-consistent algorithm would obtain a minimal context-dependent TCN of disjunctive metric constraints. We have proposed an *incremental* reasoning process. Thus, a minimal (and consistent) context-dependent TCN is assured at each new assertion. This incremental reasoning allows us to detect whether each new input constraint is inconsistent with the previously existing ones. This can be useful when problem constraints are not initially known but are successively deduced from an incremental independent process (Garrido et al., 1999).

A prototype of proposed reasoning algorithms has been implemented in Common-Lisp and is available from the author. These reasoning algorithms are being applied to an integrated architecture of planning and scheduling processes (Garrido et al., 1999). Here, the scheduling process should guarantee the consistency of each alternative partial plan (i.e.: temporal constraints and availability of resources for operations) simultaneously as the planner is generating each partial plan (Srivastava & Kambhampati, 1999). Thus, the following main features are needed:

- Management of disjunctive metric constraints. Particularly, in planning and scheduling problems the number disjuncts in input constraints is generally bounded by $l \leq 2$ (i.e.: non-simultaneous use of resources). However, temporal dependencies between constraints (i.e.: non-binary constraints) can appear. For instance, operation durations can be dependent on the order in which they are scheduled.

- Incremental reasoning. The process should interactively guarantee the consistency of each new input temporal constraint (about resources, plans, ordering, and objects) as each new step is deduced in a partial plan.

- Management of temporal contexts, where each context is associated to an alternative plan (action or state). Reasoning algorithms simultaneously work over different and alternative partial plans.

A globally labeled-consistent (and minimal) TCN allows us to determine consistent alternative choices and to obtain optimal solutions in each plan. Additionally, the proposed model can be a useful framework to apply on problems where these features also appear (Dousson et al., 1993; Garcia & Laborie, 1996; Srivastava & Kambhampati, 1999; etc.).

The computational cost of reasoning algorithms is exponential, due to the inherent complexity of the management of disjunctive constraints. However, the management of temporal contexts does not increase the complexity of the reasoning processes on disjunctive constraints.

Some improvements to decrease the empirical cost of reasoning algorithms have been proposed in this paper. The application of algorithms to handle only an explicit TCN (without making the derived constraints explicit) and empirical evaluations on several test cases are under study. Moreover, other reasoning algorithms can be applied to the temporal algebra presented, as proposed in Section 4. On the other hand, it is interesting to identify subclasses of the labeled temporal algebra where the size of label sets can be bounded, and to identify tractable subclasses of IA on the proposed model. It could also be interesting to identify the expressive power of I-L-Sets (and labeled constraints) on the basis of method described by Jeavons, Cohen and Cooper (1999). Here, each I-L-Set represents a special *derived constraint*, which expresses the inconsistency of a set of *input* elemental constraints; that is, a special type of disjunctive linear constraint (Jonsson & Bäckström, 1996; Stergiou & Koubarakis, 1996).

The proposed-labeled algebra (labeled constraints and the operations on them) can be applied to other temporal models (i.e.: to other classes of temporal constraints, operations, and reasoning algorithms). To do this, the operations of the labeled algebra ($\oplus_{lc}$, $\otimes_{lc}$, $\cup_{lc}$ and $\subseteq_{lc}$) should be defined on the basis of the respective operations ($\oplus$, $\otimes$, $\cup_T$ and $\subseteq_T$) of these models, and the reasoning algorithms should use the operations defined on labeled constraints ($\oplus_{lc}$, $\otimes_{lc}$, $\cup_{lc}$ and $\subseteq_{lc}$). This requires that these reasoning algorithms be based on the composition and intersection operations. Specifically, the application of the proposed model to tractable temporal constraints -as those identified in Section 1 (Jonsson et al., 1999; Drakengren & Jonsson, 1997; Vilain, Kautz and Van Beek, 1986; etc.)- allows for a tractable reasoning process on a hierarchy of temporal constraint contexts.

## Acknowledgements

This work was partially supported by the Generalitat Valenciana (Research Project #GV-1112/93) and by the Spanish Government (Research Project #CYCIT-TAP-98-0345). The author would sincerely like to thank the JAIR reviewers for their helpful comments and suggestions on previous versions of this paper.